\documentclass[Afour,sageh,times]{sagej}

\usepackage{moreverb,url}

\usepackage[colorlinks,bookmarksopen,bookmarksnumbered,citecolor=red,urlcolor=red,hidelinks]{hyperref}
\usepackage{graphicx}
\usepackage{svg}
\usepackage{picinpar}
\usepackage{url}
\usepackage{flushend}
\usepackage[utf8]{inputenc}
\usepackage{colortbl}
\usepackage{soul}
\usepackage{multirow}
\usepackage{pifont}
\usepackage{alltt}
\usepackage{enumerate}
\usepackage{siunitx}
\usepackage{epstopdf}
\usepackage{pbox}
\usepackage{algorithm,algorithmicx}
\usepackage{algpseudocode}
\usepackage{amsfonts} % 或者 \usepackage{dsfont}
\usepackage{dsfont}
\usepackage{threeparttable}
\usepackage{booktabs}
\usepackage[switch]{lineno}
\usepackage{empheq}  % 在导言区添加
\usepackage{bbding}
\usepackage{amsmath,amssymb,mathtools,bm}

\usepackage{amsthm}
\usepackage{thmtools}
\usepackage[normalem]{ulem}  % 提供 \uline，下划线更美观；normalem 保留  的斜体效果

\newcommand*\patchAmsMathEnvironmentForLineno[1]{%
  \expandafter\let\csname old#1\expandafter\endcsname\csname #1\endcsname
  \expandafter\let\csname oldend#1\expandafter\endcsname\csname end#1\endcsname
  \renewenvironment{#1}%
     {\linenomath\csname old#1\endcsname}%
     {\csname oldend#1\endcsname\endlinenomath}}%
\newcommand*\patchBothAmsMathEnvironmentsForLineno[1]{%
  \patchAmsMathEnvironmentForLineno{#1}%
  \patchAmsMathEnvironmentForLineno{#1*}}%
\AtBeginDocument{%
  \patchBothAmsMathEnvironmentsForLineno{equation}%
  \patchBothAmsMathEnvironmentsForLineno{align}%
  \patchBothAmsMathEnvironmentsForLineno{flalign}%
  \patchBothAmsMathEnvironmentsForLineno{alignat}%
  \patchBothAmsMathEnvironmentsForLineno{gather}%
  \patchBothAmsMathEnvironmentsForLineno{multline}%
}
\usepackage[final]{microtype}
\colorlet{reflink}{blue!80!black}
\hypersetup{colorlinks=true, linkcolor=reflink, urlcolor=reflink, citecolor=reflink, hypertexnames=false}

\makeatletter
\providecommand{\theHalgorithm}{}
\renewcommand{\theHalgorithm}{\arabic{algorithm}}
\providecommand{\theHALG@line}{}
\renewcommand{\theHALG@line}{\theHalgorithm.\arabic{ALG@line}}
\makeatother

\usepackage[nameinlink,capitalize,noabbrev]{cleveref} % 在 hyperref 之后加载
\usepackage{tabularx}
\newtheorem{theorem}{Theorem}
\newtheorem{proposition}{Proposition}

\theoremstyle{remark}
\newtheorem{remark}{Remark}

\theoremstyle{definition}

\algrenewcommand\algorithmiccomment[1]{%
  \hfill{\scriptsize\color{gray}\(\triangleright\)~#1}%
}

\newcommand{\dtstar}{\delta\bm{t}^{\star}(\bm{\phi})}

\DeclareMathOperator*{\argmin}{arg\,min}

\newcommand{\rt}{\textcolor{red}}

\definecolor{DarkGreen}{RGB}{0,180,0}

\newcommand\BibTeX{{\rmfamily B\kern-.05em \textsc{i\kern-.025em b}\kern-.08em
T\kern-.1667em\lower.7ex\hbox{E}\kern-.125emX}}

\def\volumeyear{2016}

\begin{document}
% \linenumbers
% \renewcommand{\linenumberfont}{\color{black}}

\runninghead{Hu et al.}

\title{DCReg: Decoupled Characterization for Efficient Degenerate LiDAR Registration}

\author{Xiangcheng Hu\affilnum{1}, Xieyuanli Chen\affilnum{2}, Mingkai Jia\affilnum{1}, Jin Wu\affilnum{3}, Ping Tan\affilnum{1}, Steven L. Waslander\affilnum{4}}

\affiliation{\affilnum{1}Department of Electronic and Computer Engineering, Hong Kong University of Science and Technology, Hong Kong, China\\
\affilnum{2}Department of Mechanical Engineering at National University of Defense Technology, Changsha, Hunan\\
\affilnum{3}School of Intelligent Science and Technology, University of Science and Technology Beijing, Beijing, China\\
\affilnum{4}University of Toronto Institute for Aerospace Studies and the University of Toronto Robotics Institute, Toronto, Canada}

\corrauth{Jin Wu, Full Professor, School of Intelligent Science and Technology, University of Science and Technology Beijing, Beijing, China}

\email{wujin@ustb.edu.cn}

\begin{abstract}

LiDAR point cloud registration is fundamental to robotic perception and navigation. However, in geometrically degenerate or narrow environments (e.g., corridors), registration problems become ill-conditioned:
{certain motion directions are weakly constrained, leading to unstable solutions and degraded accuracy. }
{Existing methods adopt a detect-then-mitigate paradigm but fail to reliably detect, physically interpret, and stabilize such ill-conditioning without corrupting the optimization problem.}
{In this study, we introduce \texttt{DCReg} (\textit{Decoupled Characterization for Ill-conditioned Registration}), which establishes a principled detect-characterize-mitigate paradigm that systematically addresses ill-conditioned registration through three integrated innovations.}
\textbf{First}, \texttt{DCReg} achieves \textbf{reliable ill-conditioning detection} by employing a \texttt{Schur complement} decomposition to the hessian matrix. This {structure-aware technique decouples the $6$-DoF registration problem into $3$-DoF clean rotational and translational subspaces}, eliminating coupling effects that mask degeneracy patterns in conventional {full-Hessian analyses.} 
\textbf{Second}, within these {decoupled subspaces, we develop \textbf{physically interpretable characterization} techniques that resolve \texttt{eigen-basis ambiguities} through principled basis alignment. 
This establish stable and quantitative mappings between mathematical eigenspaces and physical motion directions, providing actionable insights about which specific motions lack constraints and to what extent.} 
\textbf{Third}, leveraging this directional {spectral information}, we design a \textbf{targeted mitigation} strategy {via a structured preconditioner. While guided by MAP regularization principles for spectral shaping, our approach implements eigenvalue clamping exclusively within the preconditioner rather than modifying the origin problem. This preserves the original least-squares objective and its minimizer while enabling efficient and stable optimization} via \textit{Preconditioned Conjugate Gradient} with a single physically interpretable parameter.
Extensive experiments demonstrate \texttt{DCReg} achieves at least 20\%--50\% improvement in long-duration localization accuracy and {typical speedups of $5-30\times$ (up to $116\times$ in certain scenarios) over comparable degeneracy-aware baseline methods} across diverse environments.
Our implementation will be available at \texttt{https://github.com/JokerJohn/DCReg}.

\end{abstract}

\keywords{LiDAR Degeneracy, Point Cloud Registration, Ill-conditioning, LiDAR SLAM}

\maketitle

% Despite its fundamental importance, registration frequently becomes ill-conditioned in challenging environments, corridors, tunnels, and stairs, where certain motion directions lack sufficient geometric constraints. This vulnerability manifests as amplified noise propagation and unstable solutions, ultimately leading to catastrophic failures in autonomous navigation. The ability to handle ill-conditioning is thus not merely a technical optimization, but a prerequisite for deploying reliable autonomous systems in real-world environments.
% we show the difference and relationship in this paragraph
% Despite its fundamental importance, LiDAR registration is notoriously vulnerable in challenging environments like corridors, tunnels, and open fields with repetitive or sparse structures. This vulnerability arises because such geometrically degenerate scenes, where certain motion directions lack sufficient geometric constraints, give rise to an ill-conditioned optimization problem. This numerical ill-conditioning makes the solution highly sensitive to sensor noise and initial estimates, which manifests as amplified error propagation and unstable solutions, ultimately leading to catastrophic failures in autonomous navigation. The ability to diagnose and resolve ill-conditioning is thus not merely a technical optimization but a prerequisite for deploying reliable autonomous systems for real-world applications.

\section{Introduction}
\subsection{Motivation and Challenges}

LiDAR-based perception forms the foundation of modern autonomous systems, from self-driving vehicles to industrial robots. At its core, point cloud registration, enables these systems to build spatial understanding through applications in motion estimation~\cite{wu2022quadratic, xue2025sqpep}, mapping~\cite{ye2019tightly,qin2020lins,shan2020lio,huang2021bundle, hu2024ms},  localization~\cite{jiao2025litevloc, hu2024paloc}, {calibration~\cite{jiao2021robust, wu2019hand, wu2020globallyoptimal}, and task planning~\cite{FusionPlanner}}. 

Despite its fundamental importance, LiDAR registration remains inherently vulnerable to failure in geometrically degenerate environments such as corridors, tunnels, and open fields, where repetitive patterns or sparse features predominate.  {These challenging scenarios inherently lack sufficient constraints along specific motion directions, resulting in {near-rank-deficient} or near-singular information matrices \cite{nashed2021robust} that render the optimization problem ill-conditioned. 
This numerical {or geometrical} ill-conditioning makes the solution highly sensitive to sensor noise and initial estimates, which manifests as amplified error propagation and unstable solutions, ultimately leading to catastrophic failures in autonomous navigation.} 
Consequently, even minor perturbations in sensor measurements or initial estimates can induce catastrophic divergence in the solution, precipitating complete navigation failure. 
The ability to understand and resolve such ill-conditioning is thus not merely a technical optimization but a prerequisite for deploying reliable autonomous systems for real-world applications.
% 对“how”的区别：传统TSVD/投影要么丢弃弱信息、要么全局加正则（污染良好方向，改变原问题的可观部分解）；DCReg 在干净子空间上做等价重构（商空间/伪逆解）与定向谱整形（仅在退化方向 clamp），既不污染良好方向，又保留弱但有用的信息。
% 数学保证：DCReg 的可观部分与伪逆解一致；预条件仅改变迭代几何，不改变可观部分的最优解，且谱压缩有明确上界。
    % \item \textbf{Why it fails}: \rt{Even when ill-conditioning is detected, current methods \cite{hu2024paloc,hinduja2019degeneracy} cannot explain which physical motions are degenerate and characterize to what extent they are degenerate. The mathematical abstractions (eigenvectors) remain ambiguous from the physical reality of robot motion, leaving practitioners without actionable insights for navigation decisions or system design improvements.}
% Current approaches~\cite{tuna2023x, tuna2024informed, Zhang2016On} typically adopt a detect-then-mitigate paradigm, yet they fail to address the fundamental questions: when, why, and how does registration become ill-conditioned? This failure stems from three interconnected challenges:
{Current approaches~\cite{tuna2023x, tuna2024informed, Zhang2016On} adopt a detect-then-mitigate paradigm and have made notable progress, yet three interconnected challenges limit their effectiveness:}
\begin{itemize}
    \item \textbf{{Unreliable Detection}}: Existing detection methods~\cite{hinduja2019degeneracy, hu2024paloc, tuna2023x, Zhang2016On} {perform Hessian Spectral Analysis in optimization, yet inherent scale disparity  and rotation-translation coupling mask critical ill-conditioning patterns. This coupling causes degeneracy to be masked, leading to missed or over-detections in those scenarios where robots are most vulnerable.}

    \item {\textbf{{Insufficient Characterization:}} 
    While recent works~\cite{tuna2023x, zhao2024superloc}  have attempted to understand the underlying reasons for degeneracy, these analyses remain limited in characterizing the critical conditions. 
    They still struggle to answer: which and to what extent specific robot motions are ill-conditioned? 
     The mapping from eigenspace to physical motion axes remains unclear, leaving practitioners without actionable insights for navigation decisions or system design.
    }
% spectral truncation 
    \item \textbf{{Limited Mitigation}}: {Existing regularization-based mitigation strategies~\cite{golub1999tikhonov, tuna2024informed, hansen1990truncated, Zhang2016On}, directly modify the underlying optimization problem rather than targeted improving its conditioning. As a result, well-constrained directions can be distorted or over-regularized while attempting to stabilize ill-conditioned ones, compromising the accuracy and robustness of the original registration problem.}

\end{itemize}

% apply blanket corrections that alter the optimization problem. 
% They either applied uniformed regularization or directly discard valid information, inadvertently corrupting well-constrained directions while attempting to stabilize ill-conditioned ones.

These fundamental limitations motivate us to rethink ill-conditioned registration from first principles. Rather than treating it as an inevitable failure mode requiring external sensors or ad-hoc workarounds, we ask: can we develop a self-contained numerical framework that fundamentally understands and resolves ill-conditioning? 
Specifically, we seek an approach that: (i) reliably detects ill-conditioning by properly accounting for the inherent structure of the registration problem; (ii) provides quantitative, physically interpretable diagnosis of specific degenerate motions; and (iii) stabilizes the optimization without corrupting well-constrained directions.
% Our goal is a real-time solution that maintains mathematical integrity, ultimately enabling robust autonomous operation where current methods fail.

\subsection{Contributions}

This paper introduces a comprehensive framework to address ill-conditioned LiDAR registration problems. 
% \rt{for real-world SLAM applications.} 

% \textbf{Our first contribution} (\cref{sec:accurate_detection}) is a novel decoupled formulation of ill-conditioning detection that revealing the problematic subspaces issues of traditional coupled eigenvalue-based framework. {We analyze eigen spectral after separating translation and rotation subspaces (\cref{subsec:decoupling_theory}), to eliminate unreliable detection caused by the inherent scale disparity between rotation and translation eigenvalues~\cite{hinduja2019degeneracy} and their environmental variations~\cite{Zhang2014Jun, Zhang2016On}, which typically require frequent heuristic parameter tuning. }
% Crucially, our formulation eliminate the coupling effects between translation and rotation via \textit{Schur complement} \cite{zhang2005schur} during optimization, avoiding over or under-detection common in subspace isolation methods~\cite{hu2024paloc}. 
\textbf{Our first contribution} (\cref{sec:accurate_detection}) is a scale-robust and coupling-aware formulation for ill-conditioning detection that addresses fundamental limitations of traditional eigenvalue-based frameworks. 
Due to inherent scale disparities between translation and rotation (\cref{subsec:structural_ill_conditioning}) and environmental variations, we analyze eigen-spectral properties by separating translation and rotation subspaces (\cref{subsec:decoupling_theory}). This scale-consistent approach avoids both over- and under-detection without frequent parameter tuning (\cref{subsec:normalized_eigenvalue_analysis}), unlike {full-Hessian analyses} \cite{hinduja2019degeneracy, Zhang2014Jun, Zhang2016On}. 
 Crucially, our novel approach employs the \textit{Schur complement} to properly account for coupling effects between translation and rotation components (\cref{subsec:schur_conding}), enabling analysis within clean subspaces that exposes hidden degeneracies masked by conventional methods \cite{tuna2023x,tuna2024informed,hu2024paloc}.

While detecting ill-conditioning in {decoupled rotational and translational subspaces is necessary, it is insufficient for guiding practical applications without clearly characterizing how directions in the optimization space relate to physical robot motions.} {This challenge stems from inherent \textit{eigen-basis ambiguities}~\cite{davis1970rotation, golub2013matrix, li2006matrix} in standard \textit{Eigen Decomposition} (EVD) or \textit{Singular Value Decomposition} (SVD)  (\cref{subsec:notation}).
Unlike the canonical physical axes, eigenvectors are mathematical constructs that suffer from: 
(i) \textit{Sign Ambiguity}, 
(ii) \textit{Ordering Ambiguity}, and 
(iii) \textit{Basis Ambiguity}.
% make the eigen-directions numerically unstable and 
These ambiguities prevent a consistent and stable mapping to physical motion directions across optimization iterations, obscuring the physical structure of degeneracy.}
To tackle this issue, \textbf{our second contribution} (\cref{sec:principled_alignment}) is a physically interpretable  ill-conditioning analysis with quantitative characterization. 
We demonstrate how to: 
(i) identify specific ill-conditioned directions within subspaces via inner product analysis (\cref{subsec:inner_product}), (ii) solve for the linear coefficients that combine ill-conditioned directions and physical motion axes (\cref{subsec:max_component}), and (iii) construct stable orthogonal bases aligned with parameter space using \textit{Gram-Schmidt orthogonalization} to ensure interpretability despite eigenvector ambiguities (\cref{subsec:gram_schmidt}).

{\textbf{Our third contribution} (\cref{sec:targeted_mitigation}) is a targeted mitigation strategy that improves numerical stability through structure-aware preconditioning.
Unlike regularization-based methods (\cref{subsec:ill_conditioning_updates}), which alter the normal equations by adding explicit uniformed~\cite{golub1999tikhonov, tuna2024informed}, non-uniformed~\cite{Zhang2016On} penalties, or discarding information~\cite{hansen1990truncated}, our approach preserves the underlying optimal solution (\cref{subsec:pcg_framework}).
Leveraging the decoupled subspaces and DOF-aligned orthogonal bases established in \cref{sec:accurate_detection} and \cref{sec:principled_alignment}, we design a structure-aware and targeted preconditioner via cluster-wise eigenvalue clamping: within each identified degenerate spectral cluster, eigenvalues are shaped through a MAP view while well-conditioned directions remain essentially unaffected (\cref{subsec:tg_pre_design}). 
This design enables efficient and stable optimization via \textit{preconditioned conjugate gradient} (PCG) governed by a single interpretable parameter that controls the effective condition number bound (\cref{subsec:complementary_perspectives}).
}

\textbf{Our fourth contribution} (\cref{sub_sec:ably_hy}) demonstrates the scalability and efficiency of {\textit{DCReg}} in real-world robotics applications. Our detection module can be seamlessly integrated with existing mitigation methods, significantly enhancing performance in degenerate registration scenarios (\cref{sub_sec:hybrid_eva}). Furthermore, benefiting from the convergence properties and the efficient implementation of our methods, {we achieve $5-30\times$ (up to $116\times$) speedup compared to state-of-the-art degeneracy-aware approaches in long-duration experiments (\cref{sub_sec:run_time})}.

\section{Related Work}\label{related}
% Addressing ill-conditioned registration typically involves two steps: detection and mitigation.
{Addressing ill-conditioned registration typically involves two steps: detection and mitigation, 
with characterization often treated implicitly within the detection module.}

\subsection{Ill-Conditioning Detection}
Existing ill-conditioning detection methods can be categorized into spectral analysis and adaptive modeling approaches.

\subsubsection{Spectral Analysis Methods}
Spectral analysis methods rely on analyzing {second-order information}, covariance, or Hessian matrices from the optimization problem. \cite{Zhang2014Jun, Zhang2016On} pioneered the use of minimal eigenvalues for ill-conditioning detection. 
However, this approach requires frequent parameter tuning across different environments, as eigenvalues vary with {scene geometry, initial pose, and sensor measurements. 
More importantly, this absolute eigenvalue-based method struggles with the inherent scale disparity between rotation and translation: rotational eigenvalues typically exceed translational ones by orders of magnitude due to \textit{lever-arm effects}~\cite{pomerleau2015review, bryson2008observability}, making rotational degeneracies particularly elusive to detect.
In addition, \cite{nashed2021robust,hinduja2019degeneracy} proposed using condition numbers (i.e., relative eigenvalues) of the full Hessian matrix, yielding more consistent detection metrics. 
However, by operating on the coupled full Hessian, these methods still fail to disentangle scale disparity and rotation-translation coupling effect, often leading to unreliable detection results.}

{
Several variants attempt to resolve the scale disparity through subspace decomposition. 
Methods proposed in~\cite{rong2016detection,tagliabue2021lion,ebadi2023present} compute relative condition numbers for translation and rotation components separately using diagonal Hessian blocks. 
While this provides more reliable detection without frequent retuning, it overlooks the coupling effects between translation and rotation. 
Similar limitations affect the diagonal-block analysis in~\cite{tuna2023x, chen2024relead}, leading to both false positives and false negatives in degeneracy detection.
Alternative approaches utilize various quantitative metrics of covariance matrices, including determinants, traces, and D-optimality~\cite{carrillo2015monotonicity, liu2021localizability, jiao2021greedy, khattak2020complementary, zhao2024superloc}. 
Despite their theoretical foundation, these uncertainty-based metrics exhibit even greater sensitivity to environmental variations than minimum-eigenvalue approaches, {exacerbating parameter tuning and undermining detection reliability.}
}

% Alternative approaches utilize various quantitative metrics of covariance matrices, including determinants, traces, and D-optimality~\cite{carrillo2015monotonicity, liu2021localizability, jiao2021greedy, khattak2020complementary, zhao2024superloc}. Despite their theoretical foundation, these uncertainty-based metrics exhibit even greater sensitivity to environmental variations than minimum eigenvalue approaches, exacerbating the parameter tuning  challenge.
% To address parameter sensitivity,  \cite{nashed2021robust} proposed using condition numbers of the full Hessian matrix, yielding more consistent detection metrics. Nevertheless, this method struggles with the inherent scale disparity between rotation and translation parameters, since rotational eigenvalues typically exceed translational ones by orders of magnitude, making rotational degeneracies particularly elusive to detect.

\subsubsection{Adaptive Modeling Methods}
Adaptive modeling methods employ probabilistic or data-driven techniques for degeneracy detection. \cite{hatles2024probab, yuan2022efficient} utilize probabilistic modeling of measurement noise to calculate signal-to-noise ratios along Hessian eigenvector directions, determining degeneracy through signal strength probability definitions. While theoretically elegant, this explicit modeling approach suffers from significant performance variations based on parameter settings for multiple noise sources and their correlations in point and normal vectors.
Data-driven approaches~\cite{nobili2018predicting, nubert2022learning} implement implicit modeling for degeneracy detection. Although these methods show promise in specific scenarios, their generalization across diverse environments remains constrained by training dataset limitations. Moreover, they present significant deployment challenges on resource-constrained robotic platforms due to computational requirements.

\subsection{Ill-Conditioning Characterization}

{
Most existing methods analyze degeneracy in spectral Hessian space but provide limited descriptions of its structure in the physical parameter space, making it difficult to design truly targeted mitigation strategies.
Current approaches~\cite{hu2024paloc} often implicitly treat motion parameter axes (e.g., X/Y/Z) as directly corresponding to eigenspace directions, neglecting the fundamental difference between these bases. 
This leads to coarse characterizations such as "$X$-axis degeneracy", which in turn weakens subsequent mitigation strategies.
Other works~\cite{zhao2024superloc, tuna2023x} characterize constraints for individual measurements along physical directions, but still lack a global, quantitative analysis of degeneracy for each motion axis.
These methods do not explicitly address the fact that degeneracy characterization fundamentally depends on handling \textit{eigenbasis ambiguities} within degenerate subspaces.
Due to \textit{ordering ambiguity} from magnitude-based eigenvalue sorting and \textit{basis ambiguity} within degenerate subspaces, the resulting eigenvectors are often arbitrary linear combinations of physical axes. 
This prevents eigenvalues from being reliably mapped to specific motion parameters, particularly in scenarios with coupled multi-axis ill-conditioning.}

% \rt{Current approaches~\cite{hu2024paloc} implicitly assume direct correspondence between motion parameter space (e.g., X/Y/Z axes) and eigenspace, neglecting the fundamental difference in their basis vectors. This misconception leads to imprecise characterizations such as "$X$-axis degeneracy" and subsequently compromises mitigation strategies. 
% Furthermore, standard spectral methods suffer from \textit{eigen-basis misalignment}. 
% Due to the \textit{ordering ambiguity} from magnitude-based sorting and the \textit{basis ambiguity} in degenerate subspaces, the resulting eigenvectors are often arbitrary linear combinations of physical axes. 
% This prevents the eigenvalues from being reliably mapped to specific motion parameters, particularly in scenarios with coupled multi-axis ill-conditioning.}

% A fundamental limitation shared by these spectral methods is their inability to provide quantitative characterization of degeneracy patterns. Current approaches~\cite{hu2024paloc} incorrectly assume direct correspondence between motion parameter space (e.g., X/Y/Z axes) and eigenspace, neglecting the difference in their basis vectors. This misconception leads to imprecise characterizations such as "$X$-axis degeneracy" and subsequently compromises mitigation strategies. \rt{Furthermore, the inherent ambiguity of eigenvectors from EVD/SVD causes eigenvalues to lose their parameter space correspondence when sorted, particularly problematic in scenarios with combined ill-conditioning across multiple motion directions.}

\begin{table*}[htbp]
 \centering
  \caption{Technical Characteristics of Degeneracy-Aware Registration Algorithms}
  \label{tab:method_comparison}
  \scriptsize
  \setlength{\tabcolsep}{10pt}
  \renewcommand{\arraystretch}{1.2} 
  \setlength{\arrayrulewidth}{1.2pt}
  \begin{tabular}{@{}l|cc|cc|c|c|c@{}}
    \toprule
    \multirow{2}{*}{\textbf{Methods}} & \multicolumn{2}{c|}{\textbf{Ill-conditioning Handling}} & \multicolumn{2}{c|}{\textbf{Spectral Analysis}} & \multirow{2}{*}{\textbf{Parametrization}} & \multirow{2}{*}{\textbf{Space}} & \multirow{2}{*}{\textbf{Frame}} \\
    \cmidrule(lr){2-3} \cmidrule(lr){4-5}
    & \textbf{Detection} & \textbf{Mitigation} & \textbf{Scale} & \textbf{Coupling} & & & \\
    \midrule
    % O3D \cite{Zhou2018o3d} & \ding{56} & \ding{56} & \ding{56} & \ding{56} & $SE(3)$ & \ding{56} & $\mathcal{W}$ \\
    % \midrule
    ME-SR \cite{Zhang2016On} & Eigenvalue & Projection & \ding{56} & \Checkmark & $\mathbb{R}^3 \times SO(3)$ & Eigen & $\mathcal{B}$ \\
    ME-TSVD \cite{tuna2024informed}& Eigenvalue & TSVD & \ding{56} & \Checkmark & $\mathbb{R}^3 \times SO(3)$ & Eigen & $\mathcal{B}$ \\
    ME-TReg \cite{tuna2024informed}& Eigenvalue & Regularization & \ding{56} & \Checkmark & $\mathbb{R}^3 \times SO(3)$ & Eigen & $\mathcal{B}$ \\
    FCN-SR \cite{hinduja2019degeneracy}& Cond. & Projection & \ding{56} & \Checkmark & $\mathbb{R}^3 \times SO(3)$ & Eigen & $\mathcal{B}$ \\
    \midrule
    SuperLoc \cite{zhao2024superloc} & Statistics & \ding{56} & \ding{56} & \Checkmark & $\mathbb{R}^3 \times S^3$ & Parameter & $\mathcal{B}$ \\
    X-ICP \cite{tuna2023x}& Sampling & Constrained & \Checkmark & \ding{56} & $\mathbb{R}^3 \times SO(3)$ & Parameter & $\mathcal{W}$ \\
    \midrule
    \rowcolor{blue!8}
    \textbf{DCReg (Ours)} & \textbf{Cond.} & \textbf{PCG} & \textbf{$\checkmark$} & \textbf{$\checkmark$} & $\mathbb{R}^3 \times SO(3)$ & \textbf{Parameter} & $\mathcal{W}$ \\
    \bottomrule
  \end{tabular}
  % \vspace{-0.5pt}
  \begin{flushleft}
  \scriptsize{\textit{Note}: \textbf{Abbreviations}: Cond. = Condition number. Frame denotes the coordinate system for pose optimization. $\mathcal{B}$/\textit{$\mathcal{W}$} = Body/World frame. }
  \end{flushleft}
  \vspace{-3em}
\end{table*}

\subsection{Ill-Conditioning Mitigation}
Existing mitigation strategies that do not require external sensors fall into two main categories.

\subsubsection{Numerical Methods}
 {Numerical methods modify the Hessian or solution space of the optimization problem to address ill-conditioning, can generally be viewed as regularization-based methods.} 
Truncated SVD~\cite{hansen1990truncated} discards components corresponding to weak singular values, effectively stabilizing solutions at the cost of information loss in potentially useful directions. 
Tikhonov regularization~\cite{golub1999tikhonov, tuna2024informed} applies penalty terms uniformly across all directions, improving numerical stability but inadvertently affecting well-constrained dimensions and requiring scenario-specific parameter tuning.
 {This means that all motion directions are penalized in the same way, regardless of whether they are already well constrained. }
\cite{Zhang2014Jun, Zhang2016On} propose solution remapping that applies infinite penalties in ill-conditioned directions, restricting updates to well-constrained subspaces.  {This can be viewed as a targeted, non-uniform regularization scheme.} While providing stability, this approach ignores parameter interdependencies during optimization, potentially leading to suboptimal solutions or convergence to local minima.
 {
Another line of work adopts a two-stage optimization strategy~\cite{shan2018lego} in which a subset of variables is solved first. 
For example,~\cite{tuna2023x} preprocesses point correspondences by selecting measurements that better constrain degenerate dimensions. Although offering targeted improvements, this method similarly requires extensive environment-specific tuning, may over-suppress certain motion directions, and can be computationally expensive.}

\subsubsection{Robust Registration Approaches}
Robust registration approaches modify residual formulations through robust kernels (e.g., M-estimator, Geman-McClure, Cauchy)~\cite{vizzo2023kiss, yang2020teaser} or  {learning-based estimation~\cite{qin2023geotransformer, chen2022sc2}} to reduce outlier influence. 
While effective for outlier handling, these methods do not specifically address the numerical challenges posed by ill-conditioning. 
Extended methods~\cite{hatles2024probab} employ probabilistic adaptive weighting for correspondences, though the use of robust kernels often compromises their probabilistic interpretation.

\subsection{Research Gap}
\cref{tab:method_comparison} compares the characteristics of existing methods. 
Current approaches exhibit three fundamental limitations that motivate our work. 
 {First, detection methods struggle with the scale disparity and coupling effects between rotation and translation parameters, leading to unreliable detection.
Second, the \textit{eigen-basis misalignment} prevents accurate characterization of degeneracy patterns in the physical motion space. The lack of explicit alignment between abstract eigenvectors and physical parameter axes renders the detected degeneracies difficult to interpret and utilize for navigation decisions.
Third, existing mitigation methods typically stabilize ill-conditioning by regularizing. 
While this can improve numerical stability, it inherently induces a trade-off between stability and faithfulness to the original least-squares problem.
Most critically, current detect–then–mitigate pipelines lack an explicit characterization module that bridges spectral detection and mitigation design. 
Our work addresses these challenges through an integrated detect–characterize–mitigate framework combining \textit{Schur complement} decomposition for structure-aware subspace analysis (\cref{sec:accurate_detection}), principal direction alignment for physical characterization (\cref{sec:principled_alignment}), and a targeted stabilization solver based on structure-aware spectral preconditioning (\cref{sec:targeted_mitigation}).}

\section{Problem Formulation and Preliminaries}\label{problem_formulation}

\subsection{Notations}
\label{subsec:notation}

We denote the rigid transformation as $\bm{T} = \{\bm{R}, \bm{t}\} \in SE(3)$, comprising rotation $\bm{R} \in SO(3)$ and translation $\bm{t} \in \mathbb{R}^3$. The rotation matrix $\bm{R}$ belongs to the special orthogonal group $SO(3):= \left \{ \bm{R} \in \mathbb{R}^{3 \times 3} \,\middle|\, \bm{R}^\top \bm{R} = \bm{I}, \det(\bm{R}) = 1 \right \}$. 
For incremental updates, we employ the minimal representation $\bm{\xi} = [\bm{\phi}^\top, \delta\bm{t}^\top]^\top \in \mathbb{R}^6$, where $\bm{\phi} \in \mathbb{R}^3$ represents rotation in axis–angle form and $\delta\bm{t} \in \mathbb{R}^3$ denotes the translation increment. 
This vector resides in the \textit{parameter space}, i.e., the tangent space $T_{\bm{x}}SE(3)$ at the current linearization point $\bm{x}$. 
 {Crucially, this tangent space is spanned by a \textit{canonical physical basis} $\mathcal{E} = \{\bm{e}_1, \dots, \bm{e}_6\}$, aligned with the motion axes (e.g., $x, y, z$, roll, pitch, yaw) in the chosen parameter frame.}
The mapping between $\bm{\phi}$ and rotation matrices utilizes the exponential map $\exp: \mathfrak{so}(3) \rightarrow SO(3)$, where $\mathfrak{so}(3)$ is the Lie algebra of skew-symmetric matrices. For small rotations, we approximate $\bm{R} \approx \bm{I} + [\bm{\phi}]_\times$, with $[\cdot]_\times$ denoting the skew-symmetric matrix operator.

 {
Given a source point set $\bm{P} = \{\bm{p}_i\}_{i=1}^N$ and a target point set $\bm{Q} = \{\bm{q}_i\}_{i=1}^M$ with associated normals $\{\bm{n}_i\}_{i=1}^M$, we formulate registration as a nonlinear least-squares problem minimizing $\frac{1}{2}\|\bm{r}(\bm{\xi})\|^2$, where $\bm{r}(\bm{\xi}) \in \mathbb{R}^m$ stacks the point-to-plane residuals. 
Linearization yields the Jacobian $\bm{J} = \partial\bm{r}/\partial\bm{\xi} \in \mathbb{R}^{m \times 6}$, the gradient $\bm{g} = \bm{J}^\top\bm{r} \in \mathbb{R}^6$, and the Gauss--Newton Hessian approximation $\bm{H} = \bm{J}^\top\bm{J} \in \mathbb{R}^{6 \times 6}$.
}

%  {
% Given source point set $\mathcal{P} = \{\bm{p}_i\}_{i=1}^N$ and target point set $\mathcal{Q} = \{\bm{q}_i, \bm{n}_i\}_{i=1}^M$, we formulate the registration as a nonlinear least-squares problem minimizing $\frac{1}{2}\|\bm{r}(\bm{\xi})\|^2$, where $\bm{r}(\bm{\xi}) \in \mathbb{R}^m$ stacks the point-to-plane residuals. 
% Linearization yields the Jacobian $\bm{J} = \partial\bm{r}/\partial\bm{\xi} \in \mathbb{R}^{m \times 6}$, the gradient $\bm{g} = \bm{J}^\top\bm{r}$, and the Hessian approximation $\bm{H} = \bm{J}^\top\bm{J} \in \mathbb{R}^{6 \times 6}$.
% }

 {The spectral decomposition $\bm{H} = \bm{V}\bm{\Lambda}\bm{V}^\top$ yields the orthonormal eigenbasis $\mathcal{V} = \{\bm{v}_1, \ldots, \bm{v}_6\}$ and eigenvalues $\bm{\Lambda} = \text{diag}(\lambda_1, \ldots, \lambda_6)$. 
While $\mathcal{V}$ spans the same tangent space as the canonical basis $\mathcal{E}$, it is determined solely by the principal curvatures of the cost function. 
Consequently, $\mathcal{V}$ typically does not align with $\mathcal{E}$ due to parameter coupling. 
Moreover, in degenerate subspaces where eigenvalues cluster ($\lambda_i \approx \lambda_j$), the eigenvectors become {arbitrary} due to rotational invariance~\cite{davis1970rotation}. 
Thus, individual eigenvectors in such subspaces lack stable physical meaning; only the subspace as a whole allows for robust interpretation.}
 {
The \textit{observable subspace} corresponds to $\mathrm{range}(\bm{H})$, while its orthogonal complement constitutes the \textit{null space}. 
We quantify numerical stability using the condition number $\kappa(\bm{H}) = \lambda_{\max} / \lambda_{\min}$, where $\lambda_{\min}$ denotes the smallest non-zero eigenvalue over the observable range of $\bm{H}$.
}

 {It is important to distinguish between theoretical rank deficiency and practical ill-conditioning.
In real-world LiDAR registration, absolute singularities ($\lambda_i = 0$) are rare due to sensor noise and unstructured geometry. 
Instead, degeneracy manifests as numerical ill-conditioning, characterized by non-zero but sufficiently small eigenvalues that cause severe noise amplification. 
Therefore, our work focuses on these weakly constrained, rather than the strictly unobservable scenarios.}

\subsection{Nonlinear Point-to-Plane Registration}
\label{subsec:registration}

Point cloud registration aims to estimate the rigid transformation $\bm{T}$ that optimally aligns a source point set $\mathcal{P}$ to a target set $\mathcal{Q}$. The point-to-plane error metric, widely adopted in LiDAR registration for its superior convergence properties~\cite{xu2022fast, he2023point}, minimizes:
\begin{equation}
E(\bm{R}, \bm{t}) = \sum_{i=1}^{N} \left[\bm{n}_i^\top(\bm{R}\bm{p}_i + \bm{t} - \bm{q}_i)\right]^2,
\label{eq:point_to_plane_objective}
\end{equation}
where $\bm{q}_i$ and $\bm{n}_i$ are the corresponding target point and normal within the set of size $N$, respectively.

To solve this nonlinear optimization problem, we employ iterative linearization. At each iteration, we approximate the transformation update using the first-order Taylor expansion of rotation. The transformed source point becomes:
\begin{equation}
\bm{R}\bm{p}_i + \bm{t} \approx (\bm{I} + [\bm{\phi}]_\times)\bm{p}_i + \bm{t} = \bm{p}_i - [\bm{p}_i]_\times\bm{\phi} + \bm{t},
\label{eq:transform_approx}
\end{equation}
where we utilize the property $[\bm{\phi}]_\times\bm{p} = -[\bm{p}]_\times\bm{\phi}$.

The linearized residual for each correspondence and its associated Jacobian are:
\begin{subequations}
\begin{align}
\bm{r}_i &= \bm{n}_i^\top \left( -[\bm{p}_i]_\times\bm{\phi} + \delta \bm{t} + \bm{e}_i \right), \label{eq:residual}\\
\bm{J}_i &= \bm{n}_i^\top \begin{bmatrix} -[\bm{p}_i]_\times & \bm{I}_3 \end{bmatrix} \in \mathbb{R}^{1 \times 6}, \label{eq:jacobian}
\end{align}    
\end{subequations}
where $\bm{e}_i = \bm{p}_i - \bm{q}_i$ is the current point-to-point error.

Stacking all residuals and Jacobians, we form the linear system $\bm{r} = \bm{J}\bm{\xi} + \bm{r}_0$, where $\bm{\xi} = [\bm{\phi}^\top, \delta\bm{t}^\top]^\top$ is the parameter increment. The optimal increment is obtained by solving the normal equations:
\begin{equation}
\bm{H}\bm{\xi}^* = -\bm{g},
\label{eq:normal_equations}
\end{equation}
where $\bm{H} = \bm{J}^\top\bm{J}$ and $\bm{g} = \bm{J}^\top\bm{r}_0$.

Once $\bm{\xi}^* = [\bm{\phi}^{*\top}, \delta\bm{t}^{*\top}]^\top$ is computed, the transformation is updated as:
\begin{subequations}
\begin{align}
\bm{R} &\leftarrow \exp([\bm{\phi}^*]_\times)\bm{R}, \label{eq:rotation_update}\\
\bm{t} &\leftarrow \bm{t} + \delta\bm{t}^*. \label{eq:translation_update}
\end{align}
\end{subequations}
This process iterates until convergence, typically indicated by $\|\bm{\xi}^*\| < \epsilon$ or negligible reduction in the cost function.

\begin{figure}
    \centering
    \includegraphics[width=0.45\textwidth]{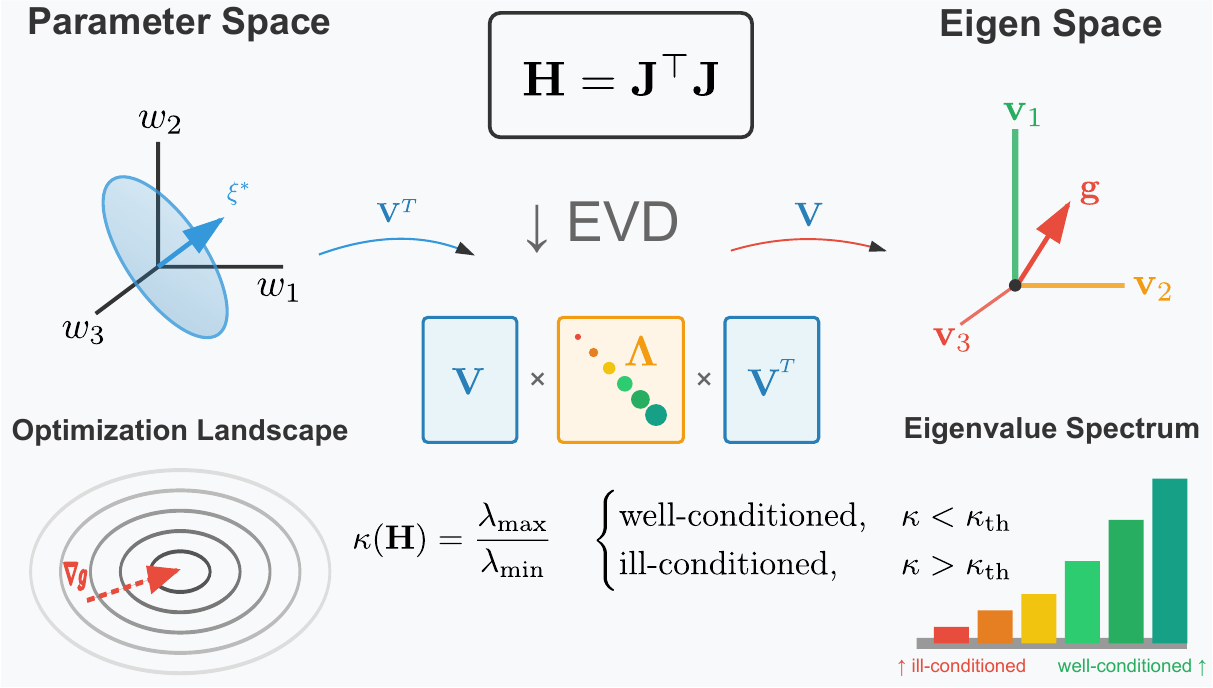}
% \caption{Physical interpretation of Hessian eigendecomposition in ICP. Transformation between parameter space and eigenspace reveals degenerate directions and their corresponding space projections, enabling principled degeneracy characterization (\cref{subsec:structural_ill_conditioning}).
% }
\caption{ {Physical interpretation of Hessian eigendecomposition in ICP for a 3-DoF decoupled subspace (e.g., rotational block)}. The transformation between the canonical parameter basis and eigenspace reveals degenerate directions and their projections in the parameter space, enabling principled degeneracy characterization (\cref{subsec:structural_ill_conditioning}).}

\label{fig:evd_explain}
    \vspace{-1.5em}
\end{figure}

\subsection{Spectral Analysis of Ill-Conditioning}
\label{subsec:structural_ill_conditioning}

Ill-conditioning in point cloud registration manifests when the Hessian matrix approaches singularity, causing numerical instability without strict rank deficiency~\cite{nashed2021robust}. This phenomenon arises from two fundamental issues: parameter scale disparity and rotation-translation coupling.

The Hessian matrix from the linearized point-to-plane formulation exhibits a block structure:
\begin{equation}
\bm{H} = \sum_{i=1}^{N} \bm{J}_i^\top \bm{J}_i = 
\begin{bmatrix} 
\bm{H}_{RR} & \bm{H}_{Rt} \\
\bm{H}_{Rt}^\top & \bm{H}_{tt} 
\end{bmatrix},
\label{eq:hessian_blocks}
\end{equation}
with sub-blocks defined as:
\begin{subequations}
\label{eq:hessian_subblocks}
\vspace{-0.5em}
\begin{align}
\bm{H}_{RR} &= \sum_{i=1}^{N} [\bm{p}_i]_\times^\top\bm{n}_i\bm{n}_i^\top[\bm{p}_i]_\times, \label{eq:HRR_v2} \\
\bm{H}_{Rt} &= -\sum_{i=1}^{N} [\bm{p}_i]_\times^\top\bm{n}_i\bm{n}_i^\top, \label{eq:HRt_v2} \\
\bm{H}_{tt} &= \sum_{i=1}^{N} \bm{n}_i\bm{n}_i^\top. \label{eq:Htt_v2}
\end{align}
\end{subequations}

\textbf{Scale Disparity:} The rotation block $\bm{H}_{RR}$ and translation block $\bm{H}_{tt}$ exhibit fundamentally different scaling properties, known as the  {\textit{lever arm effect}~\cite{pomerleau2015review, bryson2008observability} }. Analysis of $\bm{H}_{RR}$ reveals:
\begin{equation}
\|\bm{H}_{RR}\| \propto \sum_{i=1}^{N} \|\bm{p}_i\|^2\sin^2(\angle(\bm{p}_i, \bm{n}_i)),
\label{eq:rotation_scaling}
\end{equation}
indicating that rotational constraints scale quadratically with point distances from the origin. Conversely, $\bm{H}_{tt}$ depends solely on the normal distribution, independent of point distances. This creates significant scale disparity for spatially extensive point clouds.

\textbf{Rotation-Translation Coupling:} The off-diagonal blocks $\bm{H}_{Rt}$ encode cross-coupling between rotational and translational parameters~\cite{tuna2024informed}. This coupling, also scaling with point distances, causes error propagation between parameter sets and complicates the optimization landscape.

To analyze the mathematical properties of ill-conditioning, we examine the Hessian matrix through SVD, revealing the underlying geometric structure of the parameter space.
The SVD of the Hessian $\bm{H}$ is given by:
\begin{equation}
\bm{H} = \bm{U}\bm{\Sigma}\bm{V}^\top.
\label{eq:svd_decomposition}
\end{equation}
where $\bm{U}$ and $\bm{V}$ are orthogonal matrices containing the left and right singular vectors, and $\bm{\Sigma}$ is a diagonal matrix of singular values $\sigma_i$. Since $\bm{H}$ is symmetric positive semi-definite (SPD), this simplifies to an eigen decomposition:
\begin{equation}
\bm{H} = \bm{V}\bm{\Lambda}\bm{V}^\top.
\label{eq:eigen_decomposition}
\end{equation}
where $\bm{V}$ contains the eigenvectors $\bm{v}_i$ and $\bm{\Lambda}$ is a diagonal matrix of eigenvalues $\lambda_i$ (with $\lambda_i = \sigma_i^2$).
% The geometric interpretation of this decomposition reveals the fundamental structure of the optimization landscape (\Cref{fig:evd_explain})
 {As illustrated in \Cref{fig:evd_explain}, we visualize a decoupled 3D eigenspace, where the eigenvectors form a rotated orthonormal basis in the parameter space and the eigenvalues control the axis lengths. 
Near-zero eigenvalues produce elongated, flat directions, indicating structurally degenerate motions.
Although the figure shows a 3D example for clarity, the same interpretation applies to each $3\times 3$ \textit{Schur complement} block of the full 6-DoF system introduced in \Cref{subsec:decoupling_theory}.}
\begin{itemize}
\item Each eigenvector $\bm{v}_i$ represents an orthogonal principal direction in the parameter space of $\bm{\xi}$.
\item The corresponding eigenvalue $\lambda_i$ quantifies the curvature of the objective function along that direction.
% \item The transformation $\bm{V}^T\bm{\xi}$ maps the parameter vector from the standard basis to the eigenspace, revealing its components along each principal direction.
% \item $\bm{V}\bm{\eta}$ (where $\bm{\eta}$ is a vector in the eigenspace) maps from the eigenspace back to the parameter space.
\item The $\bm{V}^\top\bm{\xi}$ maps the parameter vector from the standard basis to the eigen space, decomposing it along the principal directions.
\end{itemize}

The gradient $\bm{g} = \bm{J}^\top\bm{r}$ resides in the observable subspace $\mathrm{range}(\bm{J}^\top) = \mathrm{range}(\bm{H})$, orthogonal to the null space. In the eigenspace, the gradient decomposes as:
\begin{equation}
% \vspace{-1em}
\bm{g} = \sum_{i=1}^{6} g_i\bm{v}_i, \quad \text{where} \quad g_i = \bm{v}_i^\top\bm{g},
\label{eq:gradient_decomposition}
\end{equation}
with $g_i = 0$ for eigenvectors in the null space ($\lambda_i = 0$).

This spectral framework forms the theoretical foundation for our approach to handling ill-conditioning, encompassing ill-conditioning detection (\cref{sec:accurate_detection}), quantitative characterization (\cref{sec:principled_alignment}), and targeted mitigation (\cref{sec:targeted_mitigation}). By explicitly mapping between eigenspace and parameter space, we systematically address scale disparity and coupling effects without modifying the origin problems.

\section{Ill-Conditioning Detection through Parameter Decoupling}
\label{sec:accurate_detection}

In this section, we introduce our novel parameter decoupling method that reliably detects ill-conditioning. Our key insight is that by decoupling and elimination, we can independently analyze the constraint of rotation and translation, thereby avoiding interference from scale disparity and coupling effects, resulting in clean information for each subproblem.

\subsection{Understanding Ill-conditioning through Condition Number}
\label{subsec:condition_define}

\begin{figure}
    \centering
    \includegraphics[width=0.45\textwidth]{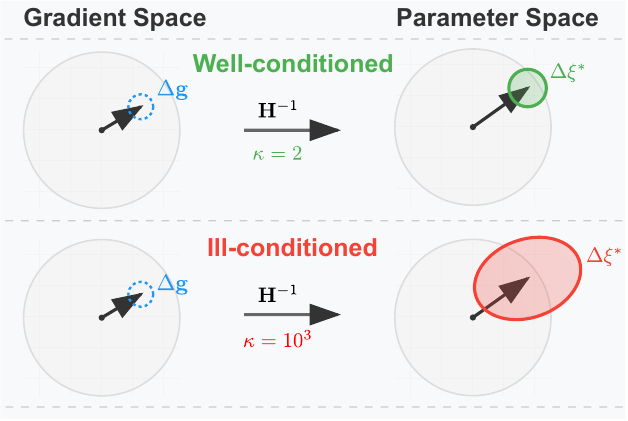}
\caption{The error amplification effect in ill-conditioning. Visualization of how condition number determines the upper bound on relative parameter error given relative perturbations in the optimization space (\cref{subsec:condition_define}).} 
\label{fig:error_amplified}
    \vspace{-1.5em}
\end{figure}

To understand the physical meaning of ill-conditioning~\cite{tuna2024informed, golub2013matrix}, we begin with the update step in optimization (\cref{eq:normal_equations}):
\begin{equation}
\bm{\xi}^* = -\bm{H}^{-1}\bm{g} = -\bm{V}\bm{\Lambda}^{-1}\bm{V}^\top\bm{g} = -\sum_{i=1}^{6} \frac{\bm{v}_i^\top\bm{g}}{\lambda_i}\bm{v}_i,
\label{eq:update_eigen}
\end{equation}
Now consider a small perturbation $\Delta(-\bm{g})$ in the right-hand side due to sensor noise or measurement errors. This causes a corresponding error in the solution $\Delta\bm{\xi}^*$:
\begin{equation}
\|\Delta\bm{\xi}^*\| \leq \|\bm{H}^{-1}\| \cdot \|\Delta(-\bm{g})\| = \frac{1}{\lambda_{\min}} \cdot \|\Delta(-\bm{g})\|,
\end{equation}
where $\lambda_{\min}$ is the smallest eigenvalue of $\bm{H}$. Similarly, for the original system:
\begin{equation}
\|\bm{\xi}^*\| \geq \frac{1}{\|\bm{H}\|} \cdot \|-\bm{g}\| = \frac{1}{\lambda_{\max}} \cdot \|-\bm{g}\|,
\end{equation}
where $\lambda_{\max}$ is the largest eigenvalue of $\bm{H}$. Combining these inequalities:
\begin{equation}
\frac{\|\Delta\bm{\xi}^*\|}{\|\bm{\xi}^*\|} \leq \frac{\lambda_{\max}}{\lambda_{\min}} \cdot \frac{\|\Delta(-\bm{g})\|}{\|-\bm{g}\|},
\label{eq:error_bound}
\end{equation}
This ratio of eigenvalues appears naturally in our error analysis, defining the condition number: $\kappa(\bm{H}) = \lambda_{\max} /\lambda_{\min}$.
% \begin{equation}
% \kappa(\bm{H}) = \frac{\lambda_{\max}}{\lambda_{\min}}
% \label{eq:condition_number}
% \end{equation}
The error bound now becomes:
\begin{equation}\label{eq:cond_def}
\frac{\|\Delta\bm{\xi}^*\|}{\|\bm{\xi}^*\|} \leq \kappa(\bm{H}) \cdot \frac{\|\Delta(-\bm{g})\|}{\|-\bm{g}\|}.
\end{equation}
This inequality reveals the fundamental importance of condition number: it acts as an error amplifier. When $\kappa(\bm{H})$ is large, even small relative perturbations in the input can lead to large relative errors in the solution (\Cref{fig:error_amplified}). 

From \cref{eq:update_eigen}, we can see why this happens, directions with small eigenvalues receive updates scaled by $1/\lambda_i$, making them sensitive to noise.
Therefore, the condition number provides a quantitative measure for detecting ill-conditioning. 
A large condition number indicates the system is ill-conditioned, with some directions much more weakly constrained than others, making the solution unstable and unreliable. With this understanding, we present our parameter decoupling approach for reliable conditioning analysis discussed below.
 {Note that the above bound is meaningful only when $\lambda_{\min}(H) > 0$, i.e., when the problem is locally
observable in the directions of interest. 
% If $\lambda_{\min}(H) = 0$, the corresponding eigen-directions lie in the null space of $H$ and are fundamentally unobservable from the given measurements: no numerical stabilization can reconstruct motion along such directions without additional information.
Our mitigation strategy is therefore designed for these weakly constrained scenarios.}

\subsection{ {Subspace Decoupling via \textit{Schur Conditioning}}}
\label{subsec:decoupling_theory}

Traditional ill-conditioning detection approaches~\cite{tuna2023x, hu2024paloc} have either analyzed the full Hessian matrix or examined its diagonal blocks in isolation.  {However, isolated analysis proves insufficient. 
Analyzing $\kappa(\bm{H}_{RR})$ merely reflects how well the point cloud constrains rotation under fixed translation, while $\kappa(\bm{H}_{tt})$ measures translation constraints under fixed rotation.}
This approach overestimates the effective curvature and the actual constraint strength, also underestimates the magnitude of ill-conditioning, leading to false negatives for degeneracy caused by rotation-translation coupling. 
Even when both $\kappa(\bm{H}_{RR})$ and $\kappa(\bm{H}_{tt})$ appear well-conditioned, the overall system can exhibit severe ill-conditioning due to strong rotation-translation coupling. 
Therefore, we must analyze rotation and translation constraints separately while properly accounting for their coupling effects in subspace. The \textit{Schur complement} provides the ideal mathematical framework for this decoupling.
For a general block matrix:
\begin{equation}
\bm{M} = \begin{bmatrix} \bm{A} & \bm{B} \\ \bm{C} & \bm{D} \end{bmatrix}, \quad
\bm{S}_A = \bm{A} - \bm{B}\bm{D}^{-1}\bm{C}.
\end{equation}
where $\bm{D}$ is assumed invertible, the \textit{Schur complement} with respect to block $\bm{D}$ is defined as $\bm{S}_A$. This operation eliminates variables associated with $\bm{D}$ while preserving their influence on variables in block $\bm{A}$. 
 {For a detailed explanation of \textit{Schur complements} and their role in block Gaussian elimination, we refer the reader to standard matrix analysis texts such as \cite{horn2013matrix}.}
%  {Applying this principle to the Hessian matrix yields the rotation and translation \textit{Schur complements}, according to conventions in mechanical engineering, these complements are defined as rotational and translational stiffness matrix: }
% \begin{subequations}
% \vspace{-1em}
% \begin{align}
% \bm{S}_R &= \bm{H}_{RR} - \bm{M}_R, \label{eq:schur_R} \\
% \bm{S}_t &= \bm{H}_{tt} - \bm{M}_t \label{eq:schur_t}, \\
% \bm{M}_R &= \bm{H}_{Rt}\bm{H}_{tt}^{-1}\bm{H}_{tR}, \label{eq:schur_mr}\\
% \bm{M}_t &= \bm{H}_{tR}\bm{H}_{RR}^{-1}\bm{H}_{Rt}. \label{eq:schur_mt}
% \end{align}
% \vspace{-1.5em}
% \end{subequations}
%  {Applying this principle to the Hessian matrix yields the rotation and translation \textit{Schur complements}. Following conventions from mechanical engineering, we define the \textit{rotational}/\textit{translational stiffness matrices} as:}
% \begin{subequations}
% % \vspace{-0.5em}
% \begin{align}
% \bm{S}_R \triangleq \bm{H}_{RR} - \bm{M}_R, \label{eq:schur_R} \\
% \bm{S}_t \triangleq \bm{H}_{tt} - \bm{M}_t, \label{eq:schur_t}
% \end{align}
% % \vspace{-0.5em}
%  {where the marginal terms are given by:}
% % \vspace{-0.5em}
% \begin{align}
% \bm{M}_R &= \bm{H}_{Rt}\bm{H}_{tt}^{-1}\bm{H}_{tR}, \label{eq:schur_mr}\\
% \bm{M}_t &= \bm{H}_{tR}\bm{H}_{RR}^{-1}\bm{H}_{Rt}. \label{eq:schur_mt}
% \end{align}
% % \vspace{-1.5em}
% \end{subequations}

 {To achieve degree-of-freedom (DoF) decoupling, we apply this principle to the Hessian matrix. This yields the {rotational} and translational \textit{ Schur complements}, defined as:}
\begin{subequations}
\begin{align}
\bm{S}_R \triangleq \bm{H}_{RR} - \bm{M}_R, \label{eq:schur_R} \\
\bm{S}_t \triangleq \bm{H}_{tt} - \bm{M}_t, \label{eq:schur_t}
\end{align}
 {where the marginalization terms are:}
\begin{align}
\bm{M}_R &= \bm{H}_{Rt}\bm{H}_{tt}^{-1}\bm{H}_{tR}, \label{eq:schur_mr}\\
\bm{M}_t &= \bm{H}_{tR}\bm{H}_{RR}^{-1}\bm{H}_{Rt}. \label{eq:schur_mt}
\end{align}
\end{subequations}

\begin{figure}
    \centering
    \includegraphics[width=0.45\textwidth]{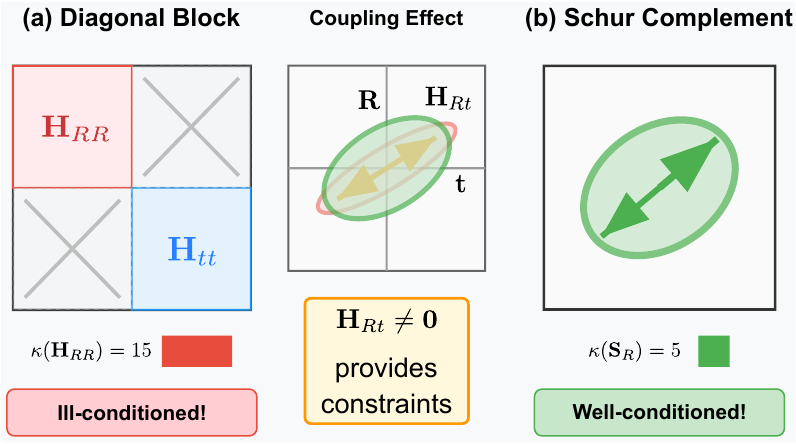}
\caption{Visualization of rotation-translation coupling effect on ill-conditioning detection. (a) Traditional diagonal subspace decoupling ignores cross-coupling terms, which may hide existing ill-conditioning. (b) \textit{Schur complement} decoupling properly isolates subspaces by accounting for coupling effects, enabling reliable degeneracy detection in the clean parameter subspace (\cref{subsec:decoupling_theory}).}
\label{fig:scale_schur_disparity}
    \vspace{-1.5em}
\end{figure}

 {These \textit{Schur complements} isolate the curvature of each DoF by optimally accounting for inter-block coupling, as formalized in ~\Cref{thm:effective_curvature}.}
The correction terms $\bm{M}_R$ and $\bm{M}_t$ quantify the bidirectional influence between rotation and translation, potentially revealing ill-conditioning that remains hidden in isolated analysis when strong coupling exists (\Cref{fig:scale_schur_disparity}).
The condition numbers of these \textit{Schur complement}s, reveal the effective curvature and true sensitivity after optimally accommodating the complementary motion components (\cref{thm:effective_curvature}). In the Jacobian space, this is equivalent to orthogonally removing components explicable by the complementary subspace (\cref{prop:projection}), while maintaining strict invariance to uniform scaling of the eliminated block (\cref{prop:scale_invariance}). Consequently, our decoupled methods can provide insights that conventional analysis would miss.

\subsection{ {Relative Condition Numbers} in Decoupled Subspaces}
\label{subsec:normalized_eigenvalue_analysis}

The decoupled subspaces provide a foundation for reliable ill-conditioning detection. Beyond merely identifying which parameter subspace exhibits ill-conditioning, we can pinpoint the specific physical directions most vulnerable to numerical instability.
We begin by extracting the eigen structure of the \textit{Schur complement} matrices:
\begin{equation}
\bm{S}_R = \bm{V}_R \bm{\Lambda}_R \bm{V}_R^\top, \quad \bm{S}_t = \bm{V}_t \bm{\Lambda}_t \bm{V}_t^\top,
\label{eq:eigen_decomp}
\end{equation}
where $\bm{\Lambda}_R = \text{diag}(\lambda_{R,1}, \lambda_{R,2}, \lambda_{R,3})$ and $\bm{\Lambda}_t = \text{diag}(\lambda_{t,1}, \lambda_{t,2}, \lambda_{t,3})$ contain the eigenvalues in ascending order. The columns of $\bm{V}_R$ and $\bm{V}_t$ represent the principal directions of constraint in the rotation and translation parameter spaces.

Then we compute  {relative condition numbers} to identify direction-specific vulnerabilities:
\begin{equation}
\kappa_{R,i} = \frac{\lambda_{R,3}}{\lambda_{R,i}}, \quad \kappa_{t,i} = \frac{\lambda_{t,3}}{\lambda_{t,i}},
\label{eq:normalized_eigenvalues}
\end{equation}
where $\kappa_{R,i}$ and $\kappa_{t,i}$ are the  {relative condition numbers} for rotation and translation, respectively. This formulation resembles direction-specific condition numbers, with the largest eigenvalue divided by each individual eigenvalue. Larger values of $\kappa_{R,i}$ or $\kappa_{t,i}$ indicate weaker constraints in the corresponding directions.

We then identify ill-constrained directions using a threshold on these  {relative condition numbers}:
\begin{equation}
\kappa_{R,i} > \kappa_{\text{th}} \text{ or } \kappa_{t,i} > \kappa_{\text{th}}.
\label{eq:ill_cond_criterion}
\end{equation}
where $\kappa_{\text{th}}$ is a threshold typically set between \SI{10}{} and \SI{50}{}. The corresponding eigenvectors indicate the specific physical directions requiring regularization.
% This approach connects directly to the error amplification analysis presented earlier. From Eq.~\ref{eq:update_eigen}, updates in direction $\bm{v}_i$ are scaled by $1/\lambda_i$, making directions with large normalized $\kappa$ values particularly susceptible to disproportionate updates that can destabilize optimization.

By analyzing these direction-specific condition numbers, we gain precise insight into which specific motion components are poorly constrained, enabling targeted stabilization strategies (\cref{alg:cond_detetcion}).

\begin{algorithm}[t]
\caption{Direction-Specific Ill-Conditioning Detection}
\label{alg:cond_detetcion}
\begin{algorithmic}[1] \\
\textbf{Input:} Hessian $\bm{H}$ with blocks $\bm{H}_{RR}$, $\bm{H}_{tt}$, $\bm{H}_{Rt}$, $\bm{H}_{tR}$; threshold $\kappa_{\text{th}}$ \\
\textbf{Output:} Ill-conditioned directions $\{\bm{V}^{\text{ill}}_R, \bm{V}^{\text{ill}}_t\}$
\State $\bm{S}_R \leftarrow \bm{H}_{RR} - \bm{H}_{Rt}\bm{H}_{tt}^{-1}\bm{H}_{tR}$ \Comment{\cref{eq:schur_R}}
\State $\bm{S}_t \leftarrow \bm{H}_{tt} - \bm{H}_{tR}\bm{H}_{RR}^{-1}\bm{H}_{Rt}$ \Comment{\cref{eq:schur_t}}
\State $\{\lambda_{R,i}, \bm{V}_{R,i}\}_{i=1}^3 \leftarrow \text{EVD}(\bm{S}_R)$ \text{(asc.)}
\State $\{\lambda_{t,i}, \bm{V}_{t,i}\}_{i=1}^3 \leftarrow \text{EVD}(\bm{S}_t)$ \text{(asc.)}
\State $\bm{V}^{\text{ill}}_R \leftarrow \emptyset$, $\bm{V}^{\text{ill}}_t \leftarrow \emptyset$
\For{$i = 1$ to $3$}
    \State $\kappa_{R,i} \leftarrow \lambda_{R,3}/\lambda_{R,i}$
    \State $\kappa_{t,i} \leftarrow \lambda_{t,3}/\lambda_{t,i}$
    \If{$\kappa_{R,i} > \kappa_{\text{th}}$} $\bm{V}^{\text{ill}}_R \leftarrow \bm{V}^{\text{ill}}_R \cup \{\bm{V}_{R,i}\}$ \EndIf
    \If{$\kappa_{t,i} > \kappa_{\text{th}}$} $\bm{V}^{\text{ill}}_t \leftarrow \bm{V}^{\text{ill}}_t \cup \{\bm{V}_{t,i}\}$ \EndIf
\EndFor
\State \Return $\{\bm{V}^{\text{ill}}_R, \bm{V}^{\text{ill}}_t\}$
\end{algorithmic}
\end{algorithm}

\subsection{Theoretical  {Guarantee}}
\label{subsec:schur_conding}

% We establish the theoretical foundation for \textit{Schur complement} conditioning based on the following assumptions.
% Unless otherwise specified, we assume $\bm{H}_{tt}\succ 0$ and $\bm{H}_{RR}\succ 0$. $\bm{S}_R$ and $\bm{S}_t$ are well-defined (Generalizations to the Moore–Penrose inverse).

We establish the theoretical foundation for \textit{Schur complement} conditioning under the following default assumptions.
Unless otherwise specified, we assume $\bm{H}_{tt}\succ 0$ and $\bm{H}_{RR}\succ 0$, so that $\bm{S}_R$ and $\bm{S}_t$ are well-defined (Generalizations using the Moore-Penrose inverse).
% All matrix norms $\|\cdot\|$ denote the spectral (2-)norm.

\begin{theorem}[\textbf{Elimination-Curvature Equivalence}]
\label{thm:effective_curvature}

For the Hessian $\bm{H}$, the \textit{Schur complements} $\bm{S}_R$ and $\bm{S}_t$ satisfy:
\begin{enumerate}[(i)]
\item The reduced quadratic after eliminating $\delta\bm{t}$ has Hessian $\bm{S}_R$:
\begin{equation}
Q(\bm{\phi}) = \tfrac{1}{2}\bm{\phi}^\top\bm{S}_R\bm{\phi} - \tilde{\bm{g}}_R^\top\bm{\phi} + \text{const}
\end{equation}
\item Under perturbation $\Delta\tilde{\bm{g}}_R$ in the reduced gradient:
\begin{equation}
\frac{\|\Delta\bm{\phi}^*\|_2}{\|\bm{\phi}^*\|_2} \leq \kappa(\bm{S}_R) \cdot \frac{\|\Delta\tilde{\bm{g}}_R\|_2}{\|\tilde{\bm{g}}_R\|_2}
\end{equation}
\end{enumerate}
Analogous results hold for $\bm{S}_t$ when eliminating $\bm{\phi}$.
\end{theorem}
\textit{Proof}: See \cref{appendix:effective_curvature}.

This theorem demonstrates that rotation and translation sensitivities are governed by $\bm{S}_R$ and $\bm{S}_t$, with their condition numbers $\kappa(\bm{S}_R)$ and $\kappa(\bm{S}_t)$ bounding the relative error amplification. 
That is, $\bm{S}_R$ is precisely the Hessian of the rotation subproblem after optimally accommodating translation; hence spectral analysis on the rotation subproblem is equivalent (the sensitivity of $\bm{\phi}$) to analyzing the full problem with $\delta\bm{t}$ eliminated.
This property enables \textit{Schur complements} to expose hidden degeneracy by optimally accounting for inter-block coupling degeneracy that remain invisible when analyzing diagonal blocks in isolation.

\begin{proposition}[\textbf{Projection Representation}]
\label{prop:projection}
Consider the Jacobian matrix $\bm{J} = [\bm{J}_R \mid \bm{J}_t] \in \mathbb{R}^{m \times 6}$, where $\bm{J}_R \in \mathbb{R}^{m \times 3}$ and $\bm{J}_t \in \mathbb{R}^{m \times 3}$ represent the rotational and translational components, respectively. Let $\bm{P}_t$ denote the orthogonal projection operator onto $\mathrm{range}(\bm{J}_t)$ with Moore-Penrose pseudoinverse, given by
\begin{equation}
\bm{P}_t \triangleq \bm{J}_t\bm{J}_t^{+}.
\end{equation}
 {Then, the rotational Schur complements can be expressed as}
\begin{equation}
\bm{S}_R = \bm{J}_R^\top(\bm{I}_m - \bm{P}_t)\bm{J}_R,
\end{equation}
where $\bm{I}_m \in \mathbb{R}^{m \times m}$ is the identity matrix, and $(\bm{I}_m - \bm{P}_t)$ projects onto the orthogonal complement of $\mathrm{range}(\bm{J}_t)$.

For the weighted case with positive definite information matrix $\bm{W} \succ 0$, the corresponding $\bm{W}$-orthogonal projector is given by $\bm{P}_t^{(\bm{W})} \triangleq \bm{J}_t(\bm{J}_t^\top\bm{W}\bm{J}_t)^{+}\bm{J}_t^\top\bm{W}$, and  {the weighted rotational Schur complements becomes}
\begin{equation}
\bm{S}_R = \bm{J}_R^\top\bm{W}(\bm{I}_m - \bm{P}_t^{(\bm{W})})\bm{J}_R,
\end{equation}
where the projections are performed with respect to the $\bm{W}$-inner product $\langle \bm{u}, \bm{v} \rangle_{\bm{W}} = \bm{u}^\top\bm{W}\bm{v}$.
\end{proposition}

\textit{Proof}: See \cref{appendix:projection}.

This projection removes components of $\mathrm{range}(\bm{J}_R)$ that can be explained by $\bm{J}_t$, retaining only the rotation information that cannot be compensated by translation. Consequently, the eigenvalues of $\bm{S}_R$ in the projected subspace accurately reflect the rotational information of the corresponding subproblem.

\begin{proposition}[\textbf{Scale- and Orthogonal-Basis Invariance}]
\label{prop:scale_invariance}

The \textit{Schur complements} exhibit scale and orthogonal-basis invariance:
\begin{enumerate}[(i)]
    \item (\textbf{Scale invariance}) For any $s>0$, if $\delta\bm{t}'=s\,\delta\bm{t}$ so that $\bm{H}_{tt}\mapsto s^2\bm{H}_{tt}$ and $\bm{H}_{Rt}\mapsto s\bm{H}_{Rt}$, then $\bm{S}_R'=\bm{S}_R$.
    
    \item  (\textbf{Orthogonal-basis invariance}) If the retained block undergoes an orthogonal change of basis $\bm{J}_R\mapsto \bm{J}_R\bm{Q}$ with $\bm{Q}\in\mathrm{SO}(3)$, then $\bm{S}_R\mapsto \bm{Q}^\top\bm{S}_R\bm{Q}$, preserving the spectrum and thus $\kappa(\bm{S}_R)$.
\end{enumerate}
 Analogous results hold for $\bm{S}_t$ and $\kappa(\bm{S}_t)$.
\end{proposition}

% \noindent
% \textit{Remark.} Proposition~\ref{prop:scale_basis_invariance_rigorous} clarifies that $\kappa(\bm{S}_R)$ is robust to (a) unit scaling of eliminated variables and (b) orthogonal re-parameterizations of the retained block. More general (non-orthogonal) re-parameterizations need not preserve the condition number.

% \begin{proposition}[\textbf{Scale- and Parameterization-Invariance}]
% \label{prop:scale_invariance}
% The \textit{Schur complements} exhibit scale  and parameterization invariance:
% For any $s > 0$, if $\delta\bm{t}' = s\delta\bm{t}$, then $\bm{S}_R' = \bm{S}_R$. Analogous results hold for $\bm{S}_t$.
% \end{proposition}
\textit{Proof}: See \cref{appendix:scale_invariance}.

This property demonstrates that \textit{Schur complements} naturally eliminate sensitivity to unit or scale changes in the eliminated parameters, directly addressing the scale disparity between rotation (radians) and translation (meters). 
If the translation unit is rescaled as $\bm{t} \leftarrow s\,\bm{t}$ with $s>0$, the Hessian blocks transform proportionally as $\bm{H}_{tt}\sim s^{2}$ and $\bm{H}_{Rt}\sim s$.
These factors cancel in $\bm{M}_{R}$, hence $\bm{S}_R $ is invariant to this scaling.
More general (non-orthogonal) re-parameterizations do not necessarily preserve the condition number.
%  {This provides the basis for our scale-invariant detection criterion, but the diagonal block $\bm{H}_{RR}$ and $\bm{H}_{tt}$ lacks this robustness.}
% This underpins a scale-invariant detection on the clean subspaces. 
% While the diagonal blocks $\bm{H}_{RR}$ and $\bm{H}_{tt}$ are themselves unaffected (up to uniform scaling) by unit changes, they lack coupling-awareness: analyzing them in isolation can substantially overestimate the effective curvature under strong rotation–translation coupling, hence missing hidden degeneracy that $\bm{S}_R$ and $\bm{S}_t$ expose.

% invariant under orthogonal re-basis of the retained block

% This property shows that $\kappa(\bm{S}_R)$ is robust to (i) unit scaling of the eliminated block and (ii) orthogonal re-basis of the retained block (e.g., frame rotations that induce $\bm{J}_R\mapsto \bm{J}_R\bm{Q}$ with $\bm{Q}\in\mathrm{SO}(3)$).
% More general (non-orthogonal) re-parameterizations do not necessarily preserve the condition number.

\begin{theorem}[\textbf{Spectral Bounds for Schur Conditioning}]
\label{thm:condition_bounds}
Let $\bm{M}_R$ and $\bm{M}_t$ denote the coupling matrices, $\bm{M}_R \succeq \bm{0}$ and $\bm{M}_t \succeq \bm{0}$, the spectral properties satisfy:
\begin{enumerate}[(i)]
\item \textbf{Loewner ordering}: $\bm{S}_R \preceq \bm{H}_{RR}$ and $\bm{S}_t \preceq \bm{H}_{tt}$, 
% \begin{align*}
% \small
% \lambda_{\min}(\bm{S}_R) \le \lambda_{\min}(\bm{H}_{RR}),\quad
% \lambda_{\max}(\bm{S}_R) \le \lambda_{\max}(\bm{H}_{RR}).
% \end{align*}

\item \textbf{Eigenvalue bounds:} For $i \in \{1,2,3\}$,
\begin{align}
    \lambda_i(\bm{H}_{RR}) - \lambda_{\max}(\bm{M}_R) &\leq \lambda_i(\bm{S}_R) \\
    &\leq \lambda_i(\bm{H}_{RR}).
\end{align}
% Analogous results hold for $\lambda_i(\bm{S}_t)$.

\item \textbf{Condition number bounds:} Under the assumption $\lambda_{\min}(\bm{H}_{RR}) > \lambda_{\max}(\bm{M}_R)$,
\begin{equation}
\kappa(\bm{S}_R) \leq \frac{\lambda_{\max}(\bm{H}_{RR})}{\lambda_{\min}(\bm{H}_{RR}) - \lambda_{\max}(\bm{M}_R)}
\end{equation}
% with an analogous bound for $\kappa(\bm{S}_t)$ under $\lambda_{\min}(\bm{H}_{tt}) > \lambda_{\max}(\bm{M}_t)$.

% \item There exists $\bm{v} \in \mathbb{R}^3$:
% \begin{equation}
% \bm{v}^T \bm{M}_R \bm{v} \approx \bm{v}^T \bm{H}_{RR} \bm{v},
% \end{equation}
% since  $\lambda_{\min}(\bm{S}_R) \ll \lambda_{\min}(\bm{H}_{RR})$.
\item \textbf{Near-cancellation:}
Let $\bm{v}_*\in\mathbb{S}^2$ attain the minimum Rayleigh quotient of $\bm{S}_R$. Then
% \begin{equation}
\begin{align}
        \bm{v}_*^\top\bm{M}_R\bm{v}_* &=\bm{v}_*^\top\bm{H}_{RR}\bm{v}_*-\lambda_{\min}(\bm{S}_R)\  \\
        &\ge\ \lambda_{\min}(\bm{H}_{RR})-\lambda_{\min}(\bm{S}_R).
\end{align}

% \end{equation}
Hence, if $\lambda_{\min}(\bm{S}_R)\ll \lambda_{\min}(\bm{H}_{RR})$, the coupling term nearly cancels the rotational curvature along $\bm{v}_*$:
\begin{align}
        \frac{\bm{v}_*^\top\bm{M}_R\bm{v}_*}{\bm{v}_*^\top\bm{H}_{RR}\bm{v}_*}\ &\ge\
1-\frac{\lambda_{\min}(\bm{S}_R)}{\bm{v}_*^\top\bm{H}_{RR}\bm{v}_*}\ \\
&\ge\  1-\frac{\lambda_{\min}(\bm{S}_R)}{\lambda_{\min}(\bm{H}_{RR})}.
\end{align}
% Analogous results hold for $\lambda_{\min}(\bm{S}_t)$.
Analogous results hold for $\bm{S}_t$, $\kappa(\bm{S}_t)$ and $\lambda_i(\bm{S}_t)$.

\end{enumerate}
\end{theorem}
\textit{Proof}: See \cref{appendix:condition_bounds}.

% \begin{theorem}[Necessary Condition for Observability Degeneration]
% \label{thm:observability_degeneration}
% If $\lambda_{\min}(S_R) \ll \lambda_{\min}(H_{RR})$, then for $v \in \mathbb{R}^3$:
% \begin{equation}
% v^T H_{Rt} H_{tt}^{-1} H_{tR} v \approx v^T H_{RR} v,
% \end{equation}
% where the real observability information is masked by the cross-correlation terms. This constitutes the degeneration mechanism being ``covered'' on $H_{RR}$, observable only in $S_R$.

% \textit{Remark:} The condition is a direct consequence of $S_R = H_{RR} - H_{Rt} H_{tt}^{-1} H_{tR}$ and can be incorporated into theoretical frameworks with rigorous proof.
% \end{theorem}

% \noindent
% \textit{Implications.} (i)–(ii) show that Schur complements can reveal {hidden degeneracy} induced by coupling:
% even when the diagonal block appears well-conditioned, subtracting the PSD coupling $\bm{M}_R$ may drastically reduce the smallest eigenvalue, inflating $\kappa(\bm{S}_R)$. (iii) provides a finite upper bound for the SPD regime; if the assumption fails, $\bm{S}_R$ may become singular and $\kappa(\bm{S}_R)=+\infty$, correctly signaling degeneracy. 
% (iv) gives a quantitative form of the “masking” mechanism without using informal “$\approx$” notation.

\begin{remark}
    From (i) and (ii), $\bm{S}_R$ and $\bm{S}_t$ are not guaranteed to be better-conditioned than their diagonal block $\bm{H}_{RR}$ and $\bm{H}_{tt}$. 
% Since the cross terms $\bm{M}_{R}$ and $\bm{M}_{t}$ are positive semi-definite (PSD).
As the Loewner ordering \cite{Loewner1934} implies that:
\begin{align*}
\lambda_{\min}(\bm{S}_R) \le \lambda_{\min}(\bm{H}_{RR}),\quad
\lambda_{\max}(\bm{S}_R) \le \lambda_{\max}(\bm{H}_{RR}).
\end{align*}
Consequently, $\kappa(\bm{S}_R)$ may be smaller than $\kappa(\bm{H}_{RR})$ when coupling is weak, or substantially larger when coupling is strong. 
% The eigenvalue inequalities follow from the Rayleigh-Ritz theorem~\cite{horn2013matrix} and Weyl's perturbation theorem~\cite{Weyl1912}. 
So the \textit{Schur complements} can reveal {hidden degeneracy} induced by coupling:
even when the diagonal block appears well-conditioned, subtracting the coupling $\bm{M}_R$ may reduce the smallest eigenvalue, inflating $\kappa(\bm{S}_R)$. 
%  {Finally, these theorems establish that $\kappa(\bm{S}_R)$ and $\kappa(\bm{S}_t)$ are scale-robust and coupling-aware.}

\end{remark}

\begin{remark}
(iii) provides a finite upper bound for the SPD regime; 
% if the assumption fails, $\bm{S}_R$ may become singular and $\kappa(\bm{S}_R)=+\infty$, correctly signaling degeneracy. 
% The bound in (iii) quantifies the sensitivity amplification due to rotation-translation coupling.
As $\lambda_{\max}(\bm{M}_R) \to \lambda_{\min}(\bm{H}_{RR})$, we have $\kappa(\bm{S}_R) \to \infty$, revealing degeneracy masked by diagonal-block analysis.
\end{remark}

\begin{remark}
    (iv) shows that the real observability information can be masked by the cross terms ($\bm{M}_R$ and $\bm{M}_t$). This constitutes the degeneration mechanism being "covered" on $\bm{H}_{RR}$, observable only in $\bm{S}_R$. Finally, it gives a quantitative form of the "masking" mechanism. 
\end{remark}

\begin{figure*}
    \centering
    \includegraphics[width=0.95\textwidth]{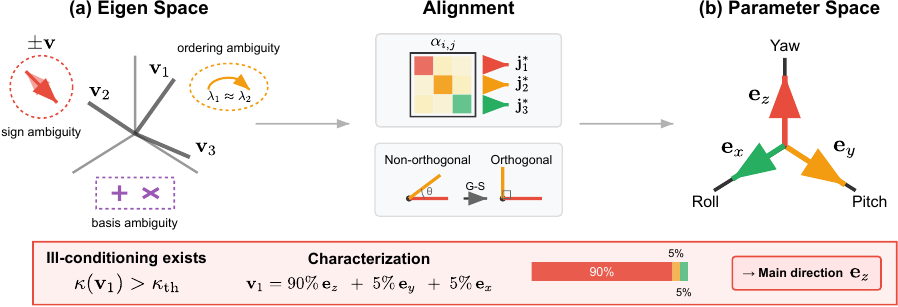}
\caption{Pipeline for quantitative ill-conditioning characterization. The three-stage process addresses eigenvector ambiguities in optimization subspaces: (1) Inner product matching resolves sign ambiguity and determines linear combinations in parameter space (\cref{subsec:inner_product}); (2) Maximum component analysis resolves ordering ambiguity to identify principal axis alignment (\cref{subsec:max_component}); (3) Gram-Schmidt orthogonalization resolves basis ambiguity, producing a stable orthonormal basis aligned with the parameter space for subsequent degeneracy mitigation (\cref{subsec:gram_schmidt}).}
\label{fig:evd_mapping_decoupy}
    \vspace{-1.5em}
\end{figure*}

\section{Resolving Eigenvector Ambiguity with Principled Alignment}
\label{sec:principled_alignment}
Building upon the decoupled analysis in \cref{subsec:decoupling_theory}, we now address the challenge of mapping abstract eigenvectors to physical motion parameters. While our \textit{Schur complement} approach successfully eliminate the coupling effect in subspaces, the eigenvectors obtained from these decoupled matrices require further interpretation to guide practical regularization strategies.

Let $\mathcal{E}_R = \{\bm{v}_{R,1}, \bm{v}_{R,2}, \bm{v}_{R,3}\}$ and $\mathcal{E}_t = \{\bm{v}_{t,1}, \bm{v}_{t,2}, \bm{v}_{t,3}\}$ represent the eigenvector sets from our \textit{Schur complement} analysis, sorted by ascending eigenvalues. In practical robotics scenarios, these eigenvectors exhibit three forms of ambiguity that complicate physical interpretation:
\begin{itemize}
    \item \textit{Sign ambiguity}: The arbitrary sign of eigenvectors ($\bm{v}$ or $-\bm{v}$) creates inconsistent physical interpretations. For instance, a robot's rotational uncertainty around the vertical axis could be represented by either $\bm{v} = [0,0,1]^\top$ or $\bm{v} = [0,0,-1]^\top$, though they represent the same physical constraint.
    
    \item \textit{Ordering ambiguity}: In corridor-like environments, where motion x and roll are similarly constrained, corresponding eigenvalues become nearly identical or the spectral gap is small, making their natural ordering unstable across iterations or frames.
    
    \item \textit{Basis ambiguity}: In planar scenarios, where the robot may be constrained to move in the horizontal plane, multiple linearly independent vectors within this plane could equally represent the degenerate subspace, leading to arbitrary basis choices by the eigen decomposition.
\end{itemize}

In this section, we present a principled alignment framework comprising three steps (\Cref{fig:evd_mapping_decoupy} and \cref{alg:characterization}): inner-product matching (\cref{subsec:inner_product}), maximum component analysis (\cref{subsec:max_component}), and \textit{Gram-Schmidt orthogonalization} (\cref{subsec:gram_schmidt}) to address these ambiguities.

\subsection{Inner-product Matching}
\label{subsec:inner_product}

We first establish that any eigenvector can be expressed as a linear combination of standard parameter axes $\{\bm{e}_x, \bm{e}_y, \bm{e}_z\}$ corresponding to the canonical basis in $\mathbb{R}^3$:
\begin{equation}
\bm{v} = v_1\bm{e}_x + v_2\bm{e}_y + v_3\bm{e}_z,
\label{eq:linear_combination}
\end{equation}
where $v_i$ are the projection coefficients that determine the contribution of each canonical direction.

Our alignment framework begins by addressing sign ambiguity through inner-product matching. For each eigenvector $\bm{v}_{R,i} \in \mathcal{E}_R$ and $\bm{v}_{t,i} \in \mathcal{E}_t$, we compute alignment coefficients with the canonical basis $\{\bm{e}_x, \bm{e}_y, \bm{e}_z\}$:
\begin{subequations}
\small
\begin{equation}
\alpha_{R,i,j} = ||\bm{v}_{R,i} \cdot \bm{e}_j|| = ||v_j|| = \cos \theta_{ij}, \quad \forall j \in \{x, y, z\},
\label{eq:rotation_inner_product}
\end{equation}
\begin{equation}
\alpha_{t,i,j} = ||\bm{v}_{t,i} \cdot \bm{e}_j|| = ||v_j|| = \cos \theta_{ij}, \quad \forall j \in \{x, y, z\}.
\label{eq:translation_inner_product}
\end{equation}
\end{subequations}
where $\theta_{ij}$ represents the angle between eigenvector $\bm{v}_i$ and canonical axis $\bm{e}_j$. 

This geometric interpretation provides intuitive understanding: larger values of $||v_j||$ indicate smaller angles, meaning the eigenvector is more aligned with that axis. The absolute value ensures invariance to arbitrary sign flips in eigen decomposition. This yields normalized alignment coefficients $\alpha_{R,i,j}, \alpha_{t,i,j} \in [0,1]$ that quantify projection magnitude onto each physical axis. These coefficients form the foundation for subsequent alignment steps by establishing initial relationship between eigenvectors and physical parameter space.

\begin{algorithm}[t]
% \algshrink
\caption{Characterization of Ill-conditioning}
\label{alg:characterization}
\begin{algorithmic}[1]
\State \textbf{Input:} $\mathcal{E}_R=\{\bm{v}_{R,i}\}_{i=1}^3$, $\mathcal{E}_t=\{\bm{v}_{t,i}\}_{i=1}^3$
\State \textbf{Output:} $\mathcal{P}_R$, $\mathcal{P}_t$
\State $\mathcal{P}_R \leftarrow \emptyset$, $\mathcal{P}_t \leftarrow \emptyset$
\State \% Inner-product matching \Comment{\crefrange{eq:rotation_inner_product}{eq:translation_inner_product}}
\For{$i=1$ to $3$}
  \For{$j\in\{x,y,z\}$}
    \State $\alpha_{R,i,j}\leftarrow |\lvert\bm{v}_{R,i}\cdot\bm{e}_j\rvert|$, \quad
           $\alpha_{t,i,j}\leftarrow|\lvert \bm{v}_{t,i}\cdot\bm{e}_j\rvert|$
  \EndFor
  \State \% Maximum component analysis \Comment{\crefrange{eq:rotation_max_component}{eq:contribution_percentage}}
  \State $j_{R,i}^* \leftarrow \arg\max_{j}\alpha_{R,i,j}$, \quad
         $j_{t,i}^* \leftarrow \arg\max_{j}\alpha_{t,i,j}$
  \State \% Alignment strength \Comment{\crefrange{eq:rotation_alignment_strength}{eq:translation_alignment_strength}}
  \State $\gamma_{R,i}\leftarrow\max_{j}\alpha_{R,i,j}$, \quad
         $\gamma_{t,i}\leftarrow\max_{j}\alpha_{t,i,j}$
\EndFor
\State \% Gram-Schmidt orthogonalization \Comment{\cref{eq:gram_schmidt_process}}
% \For{$i=1$ to $3$}
%   \State $\bm{v}'_{R,i}\leftarrow \bm{v}_{R,i}-\sum_{k=1}^{i-1}(\bm{v}_{R,i}\cdot\bm{p}_{R,k})\bm{p}_{R,k}$,
%          \quad $\bm{p}_{R,i}\leftarrow \bm{v}'_{R,i}/\|\bm{v}'_{R,i}\|$
%   \State $\bm{v}'_{t,i}\leftarrow \bm{v}_{t,i}-\sum_{k=1}^{i-1}(\bm{v}_{t,i}\cdot\bm{p}_{t,k})\bm{p}_{t,k}$,
%          \quad $\bm{p}_{t,i}\leftarrow \bm{v}'_{t,i}/\|\bm{v}'_{t,i}\|$
%   \State $\mathcal{P}_R\leftarrow \mathcal{P}_R\cup\{\bm{p}_{R,i}\}$, \quad
%          $\mathcal{P}_t\leftarrow \mathcal{P}_t\cup\{\bm{p}_{t,i}\}$
% \EndFor
\For{$i = 1$ to $3$} 
    \State $\bm{v}_{R,i}' \leftarrow \bm{v}_{R,i} - \sum_{k=1}^{i-1}(\bm{v}_{R,i} \cdot \bm{p}_{R,k})\bm{p}_{R,k}$
    \State $\bm{p}_{R,i} \leftarrow \bm{v}_{R,i}'/\|\bm{v}_{R,i}'\|$
    \State $\bm{v}_{t,i}' \leftarrow \bm{v}_{t,i} - \sum_{k=1}^{i-1}(\bm{v}_{t,i} \cdot \bm{p}_{t,k})\bm{p}_{t,k}$
    \State $\bm{p}_{t,i} \leftarrow \bm{v}_{t,i}'/\|\bm{v}_{t,i}'\|$
    \State $\mathcal{P}_R \leftarrow \mathcal{P}_R \cup \{\bm{p}_{R,i}\}$, $\mathcal{P}_t \leftarrow \mathcal{P}_t \cup \{\bm{p}_{t,i}\}$
\EndFor
\State \Return $\mathcal{P}_R$, $\mathcal{P}_t$, $\{j_{R,i}^*\}$, $\{j_{t,i}^*\}$, $\{\gamma_{R,i}\}$, $\{\gamma_{t,i}\}$
\end{algorithmic}
% \vspace{-1em}
\end{algorithm}

\subsection{Maximum Component Analysis}
\label{subsec:max_component}

The second step addresses ordering ambiguity through maximum component analysis. For each eigenvector, we identify its dominant physical axis alignment:
\begin{equation}
j_{R,i}^* = \underset{j \in \{x,y,z\}}{\arg\max}\, \alpha_{R,i,j}, \quad  j_{t,i}^* = \underset{j \in \{x,y,z\}}{\arg\max}\, \alpha_{t,i,j},
\label{eq:rotation_max_component}
\end{equation}
To provide a more intuitive relative measure, we compute the contribution percentage of each component:
\begin{equation}
c_j = \frac{||v_j||}{\sum_{i=1}^{3} ||v_i||}, \quad j \in \{x,y,z\},
\label{eq:contribution_percentage}
\end{equation}
which reflects the relative magnitude of components. For instance, if $c_z \approx 90\%$, the eigenvector primarily contributes along $\bm{e}_z$ (corresponding to yaw or vertical translation).

This operation establishes a consistent mapping criterion between eigenvectors and physical axes. The ill-conditioned direction characterized by $\bm{v}_{R,1}$ (associated with the smallest eigenvalue $\lambda_{R,1}$) is thus identified with physical rotation axis $j_{R,1}^*$. The strength of alignment is quantified by:
% \begin{subequations}
% \begin{equation}
% \gamma_{R,i} = \alpha_{R,i,j_{R,i}^*} = \max_j \alpha_{R,i,j},
% \label{eq:rotation_alignment_strength}
% \end{equation}
% \begin{equation}
% \gamma_{t,i} = \alpha_{t,i,j_{t,i}^*} = \max_j \alpha_{t,i,j}.
% \label{eq:translation_alignment_strength}
% \end{equation}    
% \end{subequations}
\begin{subequations}\label{eq:alignment_strength}
\begin{align}
    \gamma_{R,i} &= \alpha_{R,i,j_{R,i}^*} = \max_j \alpha_{R,i,j}, \label{eq:rotation_alignment_strength} \\
\gamma_{t,i} &= \alpha_{t,i,j_{t,i}^*} = \max_j \alpha_{t,i,j}. \label{eq:translation_alignment_strength}
\end{align}  
\end{subequations}
When $\gamma_{R,i}$ or $\gamma_{t,i}$ approaches $1$, the eigenvector closely aligns with a single physical axis, providing clear physical interpretation. When significantly below $1$, the ill-conditioning affects a complex combination of physical axes, requiring further orthogonalization to resolve the basis ambiguity.
This mapping allows us to understand which physical degrees of freedom each eigenvector primarily represents, providing a foundation for targeted regularization that is more precise than using raw eigenvectors which may represent arbitrary linear combinations of physical axes.

\subsection{Gram-Schmidt Orthogonalization}
\label{subsec:gram_schmidt}

The third step addresses basis ambiguity, particularly when multiple eigenvalues are similar or with small spectral gap. When \textit{eigenvalue clustering} occurs (i.e., $\lambda_{R,i} \approx \lambda_{R,j}$), the corresponding eigenvectors become numerically unstable and their orientations can vary significantly between iterations. This instability arises because while the invariant subspace $\mathcal{S}_R = \text{span}\{\bm{v}_{R,i}, \bm{v}_{R,j}\}$ remains well-defined, individual eigenvectors within it are not uniquely determined  \cite{DavisKahan1970, stewart1990matrix, Kato1995}.
According to the \textit{Davis-Kahan theorem}, the angular deviation of computed eigenvectors satisfies:
\begin{equation}
% \vspace{-0.5em}
\|\sin\Theta(\bm{U}, \tilde{\bm{U}})\|_2 \leq \frac{\|\bm{\Delta}\|_2}{\delta}
\label{eq:davis_kahan}
\end{equation}
where $\delta = \min_{i \neq j}||\lambda_i - \lambda_j||$ denotes the minimum spectral gap between distinct eigenvalue clusters, $\bm{\Delta}$ represents the perturbation matrix, and $\Theta(\bm{U}, \tilde{\bm{U}})$ measures the canonical angles between the true and computed eigenspaces. 

To establish a stable and consistent basis for subsequent processing, we construct canonical orthogonal bases $\mathcal{P}_R = \{\bm{p}_{R,1}, \bm{p}_{R,2}, \bm{p}_{R,3}\}$ and $\mathcal{P}_t = \{\bm{p}_{t,1}, \bm{p}_{t,2}, \bm{p}_{t,3}\}$ aligned with physical axes through Gram-Schmidt orthogonalization. The process prioritizes the eigenvector most strongly aligned with a physical axis:
\begin{subequations}
\label{eq:gram_schmidt_process}
\vspace{-1em}
\begin{align}
    \bm{v}_{R,i}' &= \bm{v}_{R,i} - \sum_{k=1}^{i-1}(\bm{v}_{R,i} \cdot \bm{p}_{R,k})\bm{p}_{R,k} \label{eq:gram_schmidt_rotation} \\
    \bm{p}_{R,i} &= \|\bm{v}_{R,i}'\|^{-1}\bm{v}_{R,i}' \label{eq:normalized_basis_rotation} \\
    \bm{v}_{t,i}' &= \bm{v}_{t,i} - \sum_{k=1}^{i-1}(\bm{v}_{t,i} \cdot \bm{p}_{t,k})\bm{p}_{t,k} \label{eq:gram_schmidt_translation} \\
    \bm{p}_{t,i} &= \|\bm{v}_{t,i}'\|^{-1}\bm{v}_{t,i}' \label{eq:normalized_basis_translation}
\end{align}
\end{subequations}

For rotation subspaces with eigenvalue multiplicity $m > 1$ (indicating $m$-dimensional degenerate subspaces), our orthogonalization ensures the resulting basis $\{\bm{p}_{R,1}, \bm{p}_{R,2}, ..., \bm{p}_{R,m}\}$ maintains maximum alignment with physical axes while preserving the invariant subspace spanned by the original eigenvectors. This canonical basis construction provides numerical stability and interpretability for the eigenvalue clamping strategy detailed in the following section.

% \subsection{Gram-Schmidt Orthogonalization}
% \label{subsec:gram_schmidt}

% The third step addresses basis ambiguity, particularly when multiple eigenvalues are similar. We construct orthogonal basis $\mathcal{P}_R = \{\bm{p}_{R,1}, \bm{p}_{R,2}, \bm{p}_{R,3}\}$ and $\mathcal{P}_t = \{\bm{p}_{t,1}, \bm{p}_{t,2}, \bm{p}_{t,3}\}$ aligned with physical axes through Gram-Schmidt orthogonalization. The process starts with the eigenvector most strongly aligned with a physical axis:
% \begin{subequations}
% \label{eq:gram_schmidt_process}
% \begin{align}
%     \bm{v}_{R,i}' &= \bm{v}_{R,i} - \sum_{k=1}^{i-1}(\bm{v}_{R,i} \cdot \bm{p}_{R,k})\bm{p}_{R,k} \label{eq:gram_schmidt_rotation} \\
% \bm{p}_{R,i} &= \|\bm{v}_{R,i}'\|^{-1}\bm{v}_{R,i}' \label{eq:normalized_basis_rotation} \\
% \bm{v}_{t,i}' &= \bm{v}_{t,i} - \sum_{k=1}^{i-1}(\bm{v}_{t,i} \cdot \bm{p}_{t,k})\bm{p}_{t,k} \label{eq:gram_schmidt_translation} \\
% \bm{p}_{t,i} &= \|\bm{v}_{t,i}'\|^{-1}\bm{v}_{t,i}' \label{eq:normalized_basis_translation}
% \end{align}
% \end{subequations}
% For rotation subspaces with eigenvalue multiplicity $m > 1$ (indicating $m$-dimensional degenerate subspaces), our orthogonalization ensures the resulting basis $\{\bm{p}_{R,1}, \bm{p}_{R,2}, ..., \bm{p}_{R,m}\}$ maintains maximum alignment with physical axes while spanning the same subspace as the original eigenvectors.

\section{ {Targeted Ill-Conditioning Mitigation via Structure-Aware Preconditioning}}
\label{sec:targeted_mitigation}

 {Building on the Schur-based decoupling in \cref{sec:accurate_detection} and the physically aligned eigenbases in \cref{sec:principled_alignment}, we develop a targeted ill-conditioning mitigation strategy based on a structure-aware preconditioner that acts in the decoupled subspaces with eigenvalue clamping (\cref{subsec:tg_pre_design}). This stabilizes optimization under degeneracy while preserve the least-squares solution on the observable subspace.}

\subsection{Spectral View of Ill-conditioning and Classical Regularization}
\label{subsec:ill_conditioning_updates}

\begin{figure}
    \centering
    \includegraphics[width=0.45\textwidth]{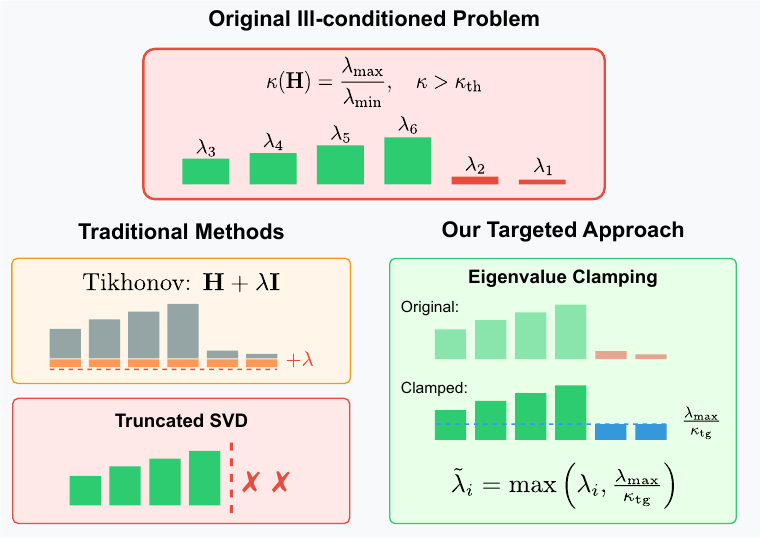}
% \caption{Algorithmic principles for mitigating ill-conditioned optimization. Visualization of how different methods modify the solution space to mitigate numerical degeneracy in point cloud registration.}
\caption{ {\textbf{Algorithmic principles for spectral shaping.}
Visualization of how different methods shape the eigenvalue spectrum to mitigate degeneracy: traditional regularization directly modifies the Hessian, whereas our method uses targeted eigenvalue clamping only in a structure-aware preconditioner.}}
\label{fig:pcg}
    \vspace{-1.5em}
\end{figure}

 {From the spectral form of the pose update in \cref{eq:update_eigen}, the solution of the normal equations \cref{eq:normal_equations} can be written component-wise as
\begin{equation}
    \label{eq:spectral_update_again}
    \Delta\xi_j^* = -\sum_{i} (\bm{v}_i^\top \bm{g}/\lambda_i)\,[\bm{v}_i]_j.
\end{equation}
Here $-(\bm{v}_i^\top \bm{g})/\lambda_i$ is the update amplitude along eigen-direction $\bm{v}_i$: the numerator $\bm{v}_i^\top \bm{g}$ is the directional gradient, while $1/\lambda_i$ acts as an inverse curvature term, and each pose component $\Delta\xi_j^*$ is obtained by projecting this expansion back to the canonical basis.
Ill-conditioning arises when some $\lambda_i$ become very small: the inverse gain $1/\lambda_i$ amplifies linearization errors and measurement noise along the corresponding eigen-directions, making the update highly sensitive and potentially unstable (\cref{subsec:condition_define}). Even if the cost is only weakly informative along $\bm{v}_i$, the term $\bm{v}_i^\top \bm{g}$ rarely vanishes exactly, so small $\lambda_i$ can still induce large and unreliable steps. As illustrated in \cref{fig:pcg}, classical remedies can be interpreted directly at the level of \cref{eq:spectral_update_again}:
\begin{itemize}
    \item \textbf{TReg.}
    Solving $(\bm{H} + \lambda\bm{I})\Delta\bm{\xi} = -\bm{g}$ yields the same eigenvectors $\bm{v}_i$, but shifts all eigenvalues uniformly $\lambda_i \mapsto \lambda_i + \lambda$~\cite{golub1999tikhonov, tuna2024informed}.
    In \cref{eq:spectral_update_again}, the coefficients become
    $(\bm{v}_i^\top \bm{g})/(\lambda_i + \lambda)$: only the denominators are modified, uniformly shrinking all update amplitudes.
    This improves conditioning but penalizes every direction, including already well-constrained ones, and is equivalent to adding isotropic virtual measurements, i.e., solving a different, explicitly regularized least-squares problem.
    \item \textbf{TSVD.}
    TSVD~\cite{hansen1990truncated, tuna2024informed} discards directions whose singular values fall below a threshold.
    Spectrally, this is equivalent to setting $(\bm{v}_i^\top \bm{g})/\lambda_i \mapsto 0$ whenever $\lambda_i < \tau$, i.e. completely removing the contribution of weak directions from the sum in \cref{eq:spectral_update_again}.
    This stabilizes the solution but irreversibly loses potentially useful information along those directions.
    \item \textbf{SR.}
    Methods such as ME-SR and FCN-SR~\cite{Zhang2014Jun, Zhang2016On}
    detect ill-conditioned directions and explicitly project the update onto a well-constrained subspace.
    From a spectral viewpoint, this can be seen as imposing an infinite penalty along selected directions: $\lambda_i \mapsto \lambda_i + \gamma_i$ with $\gamma_i \to \infty$, forcing $(\bm{v}_i^\top \bm{g})/\lambda_i \to 0$ along these axes.
    This is a targeted but extremely strong regularization and effectively redefines the feasible update set.
    Moreover, because the eigenvectors $\bm{v}_i$ are in general linear combinations of physical axes, canceling motion along a degenerate eigen-direction suppresses a coupled combination of rotational and translational components, rather than a single physical DoF, and thus inevitably affects other pose dimensions~\cite{tuna2023x}.
\end{itemize}
}

 {
All three strategies act directly on the spectrum of the full Hessian and either
(i) uniformly damp all directions (TReg),
(ii) drop weak directions completely (TSVD), or
(iii) impose infinite penalties on selected modes (SR).
They operate in the coupled 6-DoF space and ignore the geometric structure revealed by the rotational/translational Schur complements $\bm{S}_R, \bm{S}_t$ and the aligned eigenbases introduced in \cref{sec:principled_alignment}.
As a result, they reweight mixed motion directions rather than stabilizing individual decoupled axes, and, by directly modifying the normal equations, they generally alter the underlying least-squares objective and its observable subspace solution.
Therefore, our goal is to bound the inverse gain along genuinely weak directions within the decoupled Schur subspaces via a structure-aware preconditioner, so that degeneracy is mitigated while the minimizer in the observable subspace remains identical to that of the original problem.
}

\begin{algorithm}[t]
\caption{Preconditioned Conjugate Gradient}
\label{alg:pcg}
\begin{algorithmic}[1] \\
\textbf{Input:}
$\bm{H}$, $\bm{g}$, $\{\lambda_{R,i}, \bm{V}_{R,i}\}$, $\{\lambda_{t,i}, \bm{V}_{t,i}\}$, $\kappa_{\text{tg}}$  \\
\textbf{Output:} Optimized parameter update $\Delta\boldsymbol{\xi}$ 
\State \% Construct preconditioner \Comment{\crefrange{eq:preconditioner_block_final}{eq:eigenvalue_clamping}}
\State $\tilde{\lambda}_{i} \leftarrow \max(\lambda_{i}, \lambda_{\text{max}}/\kappa_{\text{tg}})$ \quad $\forall i \in \{1,2,3\}$
\State  $\bm{P} \leftarrow \text{blkdiag}(\bm{V}_R\tilde{\bm{\Lambda}}_R^{-1}\bm{V}_R^{\top}, \bm{V}_t\tilde{\bm{\Lambda}}_t^{-1}\bm{V}_t^{\top})$ 
\State \%  Solving PCG
\State $\bm{r}_0 \leftarrow -\bm{g}$, $\bm{z}_0 \leftarrow \bm{P}\bm{r}_0$, $\bm{p}_0 \leftarrow \bm{z}_0$, $\Delta\boldsymbol{\xi}_0 \leftarrow \bm{0}$
\For{$k = 0$ to convergence}
    \State $\alpha_k \leftarrow \bm{r}_k^{\top}\bm{z}_k / \bm{p}_k^{\top}\bm{H}\bm{p}_k$ \Comment{Step size}
    \State $\Delta\boldsymbol{\xi}_{k+1} \leftarrow \Delta\boldsymbol{\xi}_k + \alpha_k\bm{p}_k$ \Comment{Update parameters}
    \State $\bm{r}_{k+1} \leftarrow \bm{r}_k - \alpha_k\bm{H}\bm{p}_k$ \Comment{Update residual}
    \State $\bm{z}_{k+1} \leftarrow \bm{P}\bm{r}_{k+1}$ \Comment{Targeted preconditioning}
    \State $\beta_k \leftarrow \bm{r}_{k+1}^{\top}\bm{z}_{k+1} / \bm{r}_k^{\top}\bm{z}_k$ \Comment{Conjugacy coefficient}
    \State $\bm{p}_{k+1} \leftarrow \bm{z}_{k+1} + \beta_k\bm{p}_k$ \Comment{New search direction}
\EndFor
\State \Return $\Delta\boldsymbol{\xi}_{k+1}$
\end{algorithmic}
\end{algorithm}

\subsection{Preconditioned Conjugate Gradient}
\label{subsec:pcg_framework}

 {We apply left preconditioning to the normal equations \cref{eq:normal_equations}.
Let $\bm{P} \succ 0$ be a \textit{symmetric positive definite} (SPD) preconditioner for
$\bm{H}$.
The preconditioned system is
\begin{equation}
    \underbrace{(\bm{P}\bm{H})}_{\text{well-conditioned}}
    \Delta\boldsymbol{\xi} = \bm{P}(-\bm{g}),
    \label{eq:pcg_system}
\end{equation}
and, since $\bm{P}$ is invertible, \cref{eq:pcg_system} is algebraically
equivalent to the original system: both have the same unique solution
$\Delta\bm{\xi}^* = -\bm{H}^{-1}\bm{g}$.
Preconditioning therefore does not change the minimizer; it only reshapes the
spectrum of the effective operator $\bm{P}\bm{H}$ seen by the linear solver.
With $\bm{H} \succ 0$ and $\bm{P} \succ 0$, the symmetrically preconditioned
operator $\bm{P}^{1/2}\bm{H}\bm{P}^{1/2}$ remains SPD, so PCG applied to
\cref{eq:pcg_system} is well defined and its convergence is governed by the
spectrum of $\bm{P}\bm{H}$ (which shares its eigenvalues with
$\bm{P}^{1/2}\bm{H}\bm{P}^{1/2}$).
In addition, \cref{thm:solution_equivalence} guarantees that eliminating
translation via the \textit{Schur complement} preserves the least-squares minimizer
in the rotational subspace, so our preconditioner can safely operate on these
decoupled blocks.}

 {
Preconditioned Krylov methods such as PCG~\cite{golub2013matrix,shewchuk1994introduction}
build search directions in the Krylov subspace of $\bm{P}\bm{H}$, and their convergence rate depends on the condition number and eigenvalue clustering of $\bm{P}\bm{H}$, rather than on $\kappa(\bm{H})$.
A well-designed $\bm{P}$ compresses the spectral range and clusters the eigenvalues of $\bm{P}\bm{H}$, which improves numerical stability and reduces the iterations counts.
PCG performs the iterations
\begin{equation}
    \bm{z}_k = \bm{P}\bm{r}_k,
    \qquad
    \bm{r}_k = -\bm{g} - \bm{H}\Delta\bm{\xi}_k,
    \label{eq:pcg_step_short}
\end{equation}
where the preconditioning step $\bm{z}_k$ approximates the
Newton direction $\bm{H}^{-1}\bm{r}_k$ when $\bm{P}$ is chosen to approximate
$\bm{H}^{-1}$.
This rescaling of the search space prevents extremely weakly constrained
directions from dominating the optimization, while well-constrained
directions retain their natural behavior. \cref{alg:pcg} demonstrates the whole process.}

 {
It is important to emphasize that the stability and convergence
improvements in \textit{DCReg} stem from the spectral shaping implemented by
$\bm{P}$, not from PCG alone. 
Unlike explicit regularization, which permanently alters the normal equations,
preconditioning confines spectral modifications to the iterative geometry:
extreme eigenvalue ratios are absorbed into $\bm{P}$, while the underlying
least-squares objective and its minimizer remain unchanged.
In principle, the same preconditioner could be used together with any
symmetric positive-definite solver; we choose PCG for its scalability and
memory efficiency in large-scale point cloud registration.
}

 {
The crucial question is therefore how to design $\bm{P}$ in a way that is
(i) solution-preserving, (ii) targeted to weak directions, and
(iii) aware of the Hessian structure.
Following problem-structured preconditioners in certifiable estimation
\cite{rosen2022accelerating}, we design $\bm{P}$ in the decoupled Schur
subspaces and use eigenvalue clamping to enforce a desired condition-number
bound, thereby achieving targeted spectral shaping without modifying the
original problem.
}

\subsection{Structure-Aware Preconditioner in Decoupled Subspaces}
\label{subsec:tg_pre_design}

% We then exploit the Schur-based decoupling and the physically aligned eigenbases
% (\cref{sec:accurate_detection,sec:principled_alignment}) to design a
% targeted and structure-aware preconditioner:
We exploit the Schur-based decoupling and the physically aligned eigenbases
(\cref{sec:accurate_detection,sec:principled_alignment}) to design a
targeted, structure-aware block-diagonal preconditioner:
% We first construct a block-diagonal preconditioner:
\begin{equation}
    \bm{P}
    \triangleq
    \begin{bmatrix}
        \bm{P}_R & \bm{0} \\
        \bm{0}   & \bm{P}_t
    \end{bmatrix},
    \label{eq:preconditioner_block_final}
\end{equation}
with each block defined in its Schur eigenbasis as
\begin{equation}
    \bm{P}_R=\bm{V}_R \tilde{\bm{\Lambda}}_R^{-1} \bm{V}_R^\top,
    \qquad
    \bm{P}_t=\bm{V}_t \tilde{\bm{\Lambda}}_t^{-1} \bm{V}_t^\top,
    \label{eq:preconditioner_blocks_final}
\end{equation}
where $\bm{V}_R,\bm{V}_t$ are the aligned eigenbases of $\bm{S}_R,\bm{S}_t$
and $\tilde{\bm{\Lambda}}_R,\tilde{\bm{\Lambda}}_t$ denote the clamped spectra.
 {In practice, the action of $\bm{P}$ on a vector is implemented by
projecting into the eigenbases $\bm{V}_R,\bm{V}_t$, scaling by
$\tilde{\bm{\Lambda}}_R^{-1},\tilde{\bm{\Lambda}}_t^{-1}$, and projecting back,
without explicitly forming dense matrices.}

 {The critical element is to adjust eigenvalues in the preconditioner to
mitigate ill-conditioning while preserving all information in the original
normal equations.}
Unlike uniform regularization that penalizes all directions equally, we
introduce a $\kappa_{\text{tg}}$-based eigenvalue clamping strategy. For each
Schur block, let $\lambda_{\max}$ denote its largest eigenvalue and
$\{\lambda_i\}$ its eigenvalues.
We define
\begin{equation}
\tilde{\lambda}_{i} = f(\lambda_i) =
\begin{cases}
\lambda_{i} & \text{if $\lambda_{i} > \lambda_{\text{max}}/\kappa_{\text{tg}}$}, \\
\lambda_{\text{max}}/\kappa_{\text{tg}} & \text{otherwise},
\end{cases}
\label{eq:eigenvalue_clamping}
\end{equation}
and set $\tilde{\bm{\Lambda}}_R = \mathrm{diag}(\tilde{\lambda}_{R,i})$,
$\tilde{\bm{\Lambda}}_t = \mathrm{diag}(\tilde{\lambda}_{t,i})$ by applying
\cref{eq:eigenvalue_clamping} to the rotational and translational eigenvalues
with their respective $\lambda_{R,\max}$ and $\lambda_{t,\max}$.
 {
Equivalently, let
$\bm{\Delta}_R = \mathrm{diag}(\tilde{\lambda}_{R,i} - \lambda_{R,i})$ and
$\bm{\Gamma}_R = \bm{V}_R \bm{\Delta}_R \bm{V}_R^\top$ (and similarly
$\bm{\Gamma}_t$ for $\bm{S}_t$), so that
$\bm{S}_R + \bm{\Gamma}_R = \bm{V}_R \tilde{\bm{\Lambda}}_R \bm{V}_R^\top$.
}
 {
This design has three important properties:
\begin{itemize}
    \item \textbf{Structure-aware.}
    The preconditioner operates on $\bm{S}_R$ and $\bm{S}_t$, which already
    encode rotational and translational curvature after eliminating
    cross-coupling (\cref{sec:accurate_detection}). 
    The clamping is applied in the aligned eigenbases
    (\cref{sec:principled_alignment}) in a cluster-wise constant fashion,
    ensuring that, despite arbitrary rotations within degenerate subspaces,
    the spectral modification remains consistent and basis-invariant
    (\cref{subsec:gram_schmidt}).
    \item \textbf{Targeted spectral shaping.}
    Only eigenvalues below
    $\lambda_{\max}/\kappa_{\mathrm{tg}}$ are lifted; well-constrained
    directions remain unchanged.
    As shown in \cref{thm:map_interpretation}, this corresponds to
    the positive semidefinite lift that enforces
    $\kappa(\bm{S}_R + \bm{\Gamma}_R) \le \kappa_{\mathrm{tg}}$
    (analogously for $\bm{S}_t$), providing an explicit link between
    parameter $\kappa_{\mathrm{tg}}$ and the spectral envelope.
    \item \textbf{Solution-preserving.}
    The clamped spectra $\tilde{\bm{\Lambda}}_R, \tilde{\bm{\Lambda}}_t$
    are used only in the preconditioner \cref{eq:preconditioner_block_final,eq:preconditioner_blocks_final}.
    We do {not} replace $\bm{H}$ by $\bm{H}+\bm{\Gamma}$ nor $\bm{S}_R$ by $\bm{S}_R+\bm{\Gamma}_R$ in the normal equations.
    By \cref{thm:solution_equivalence,thm:map_interpretation},
    the least-squares minimizer over the observable subspace is preserved,
    while the spectrum of the preconditioned operator $\bm{P}\bm{H}$ is explicitly controlled by $\kappa_{\mathrm{tg}}$.
\end{itemize}
}

\subsection{ {Theoretical Guarantee}}
\label{subsec:complementary_perspectives}

We briefly summarize the theoretical properties of the proposed
preconditioner; detailed proofs are provided in
\cref{appendix:solution_equivalence} and
\cref{appendix:map_interpretation}.

\begin{figure*}
    \centering
    \includegraphics[width=0.9\textwidth]{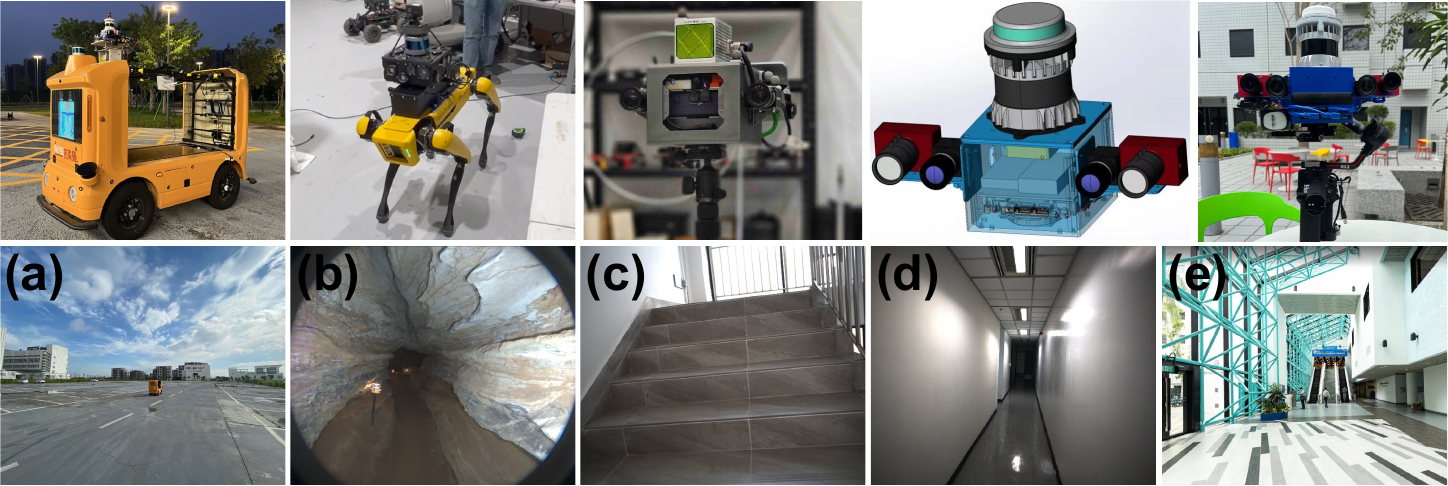}
\caption{Dataset scenarios and robotic platforms. The experimental environments include: (a) open parking lot~\cite{wei2024fusionportablev2}, (b) narrow cave \cite{zhao2024superloc, zhao2024subt}, (c) confined stairway~\cite{chen2024heterogeneous}, (d) narrow corridor\cite{jiao2022fusionportable}, and (e) indoor hallway~\cite{jiao2022fusionportable}.}
\label{fig:sensor_setup}
    \vspace{-1.5em}
\end{figure*}

\begin{theorem}[\textbf{Equivalent Reconstruction on Decoupled Subspace}]
\label{thm:solution_equivalence}
Let $\bm{S}_R = \bm{V}_R\bm{\Lambda}_R\bm{V}_R^\top$ be the eigendecomposition of the rotational Schur complement with $\bm{\Lambda}_R = \text{diag}(\lambda_{R,1}, \lambda_{R,2}, \lambda_{R,3})$, $0 \leq \lambda_{R,1} \leq \lambda_{R,2} \leq \lambda_{R,3}$. For the reduced quadratic:
\begin{equation}
Q(\bm{\phi}) = \frac{1}{2}\bm{\phi}^\top\bm{S}_R\bm{\phi} - \tilde{\bm{g}}_R^\top\bm{\phi} + \mathrm{const}
\end{equation}
where $\tilde{\bm{g}}_R = \bm{g}_R - \bm{H}_{Rt}\bm{H}_{tt}^{-1}\bm{g}_t$, the following holds:
\begin{enumerate}[(i)]
\item $\tilde{\bm{g}}_R \in \mathrm{range}(\bm{S}_R)$
\item $\bm{\phi}^* = \bm{S}_R^{+}\tilde{\bm{g}}_R$, where $\bm{S}_R^{+} = \bm{V}_R\bm{\Lambda}_R^{+}\bm{V}_R^\top$
\item $\forall i: \lambda_{R,i} > 0 \Rightarrow [\bm{V}_R^\top\bm{\phi}^*]_i = \lim_{\epsilon \to 0^+}[\bm{V}_R^\top\bm{\phi}_{\epsilon}]_i$
\end{enumerate}
where $\bm{\phi}_{\epsilon}$ solves $(\bm{S}_R + \epsilon\bm{I})\bm{\phi}_{\epsilon} = \tilde{\bm{g}}_R$.
Analogous results hold for the translation subproblem with $\bm{S}_t$.
\end{theorem}
\textit{Proof}: See \cref{appendix:solution_equivalence}.

 {\cref{thm:solution_equivalence} formalizes the elimination--curvature equivalence in the rotational Schur subspace: the least-squares solution is well defined via the pseudoinverse and stable under vanishing isotropic regularization.
This provides the basis for designing preconditioners that act on $\bm{S}_R$ and $\bm{S}_t$ without changing the observable-subspace solution.}

% \lambda_{\text{max}}/\kappa_{\text{tg}}

\begin{theorem}[ {\textbf{MAP Interpretation of Eigenvalue Clamping}}]
\label{thm:map_interpretation}
Given a target condition number $\kappa_{\text{tg}} > 1$,  {define the clamped eigenvalues:}
\begin{equation}
\tilde{\lambda}_{R,i} = \max\left(\lambda_{R,i}, \frac{\lambda_{R,\max}}{\kappa_{\text{tg}}}\right), \quad i \in \{1,2,3\}.
\end{equation}
Let $\bm{\Gamma}_R = \bm{V}_R\text{diag}(\tilde{\lambda}_{R,i} - \lambda_{R,i})\bm{V}_R^\top$. 
 {Consider the regularized problem:}
\begin{equation}
\argmin{\bm{\phi}} \frac{1}{2}\bm{\phi}^\top(\bm{S}_R + \bm{\Gamma}_R)\bm{\phi} - \tilde{\bm{g}}_R^\top\bm{\phi}.
\end{equation}
Then:
\begin{enumerate}[(i)]
\item This problem is equivalent to MAP estimation with prior $\bm{\phi} \sim \mathcal{N}(\bm{0}, \bm{\Gamma}_R^{-1})$.
\item $\bm{S}_R + \bm{\Gamma}_R = \bm{V}_R\tilde{\bm{\Lambda}}_R\bm{V}_R^\top$ where $\tilde{\bm{\Lambda}}_R = \text{diag}(\tilde{\lambda}_{R,i})$.
\item $\kappa(\bm{S}_R + \bm{\Gamma}_R) \leq \kappa_{\text{tg}}$.
\end{enumerate}
Analogous results hold for the translation block with $\bm{S}_t$.
\end{theorem}
\textit{Proof}: See \cref{appendix:map_interpretation}.

 {This MAP perspective thus serves to justify and parameterize the clamping rule (through the single parameter $\kappa_{\mathrm{tg}}$), while the actual algorithm remains strictly solution-preserving with respect to the original least-squares objective.}
 {
\begin{remark}
\cref{thm:solution_equivalence} and \cref{thm:map_interpretation} show that the proposed structure-aware preconditioner with eigenvalue clamping :
(i) preserves the least-squares solution in the observable subspace;
(ii) bounds condition number in each decoupled subspace, leading to a well-controlled spectrum for the preconditioned operator;
and (iii) achieves {targeted} spectral shaping confined to weak directions in the physically interpretable decoupled subspaces, in contrast to global uniform regularization.
\end{remark}
}

\begin{table}[!t]
  \centering
  \caption{Degenerate Scenarios in Real-world Datasets}
  \label{tab:degenerate_scenarios}
  \footnotesize
  \setlength{\tabcolsep}{5pt}
  \renewcommand{\arraystretch}{1.2} 
  \setlength{\arrayrulewidth}{1pt}
  \begin{tabular}{@{}l|cc|c|c|c@{}}
    \toprule
    \multirow{2}{*}{\textbf{Scene}} & \multicolumn{2}{c|}{\textbf{Degenerate}} & \multirow{2}{*}{\textbf{Duration}} & \multirow{2}{*}{\textbf{Sensor}} & \multirow{2}{*}{\textbf{Dataset}} \\
    \cmidrule(lr){2-3}
    & \textbf{Trans.} & \textbf{Rot.} & \textbf{(\SI{}{s})} & & \\
    \midrule
    % \multicolumn{5}{@{}l}{ \textbf{Simulated}}                    \\
    % \midrule[0.03cm]
    % Cylinder      & \checkmark\checkmark\checkmark & \checkmark & 120 & Simulated & -- \\
    % \midrule
    \multicolumn{5}{@{}l}{ \textbf{Handheld Indoor}}                    \\
    \midrule[0.03cm]
    Corridor      & \checkmark\checkmark & \checkmark & 572 & OS128 & FP \\
    Building      &  &  & 572 & OS128 & FP \\
    Stairs   & \checkmark & \checkmark & 331 & OS64 & GEODE \\
    \midrule
    % \multicolumn{5}{@{}l}{ \textbf{Self-balancing Scooter}}                    \\
    \multicolumn{5}{@{}l}{ \textbf{Open-field Outdoor}}                    \\
    \midrule[0.03cm]
    % Underground   & \checkmark\checkmark & \checkmark & 274 & OS128 & FPV2 \\
    % Grass   & \checkmark\checkmark & \checkmark & 301 & OS128 & FPV2 \\
    Parking Lot   & \checkmark\checkmark & \checkmark & 502 & Pandar32 & MS \\
    \midrule
    \multicolumn{5}{@{}l}{ \textbf{Handheld}}                    \\
    \midrule[0.03cm]
    Cave01   & \checkmark & \checkmark & 838 & VLP16 & SubT-MRS \\
    Cave02   & \checkmark & \checkmark & 986 & VLP16 & SubT-MRS \\
    Cave04   & \checkmark & \checkmark & 959 & VLP16 & SubT-MRS \\
    % \midrule
    % \multicolumn{5}{@{}l}{ \textbf{UGV}}                    \\
    % \midrule[0.03cm]
    % Grass   & \checkmark\checkmark & \checkmark & 210 & OS128 & FPV2 \\
    \bottomrule
  \end{tabular}
  \vspace{0.5pt}
  \begin{flushleft}
  \scriptsize{\textit{Note}: \checkmark~= degenerate dimension. Number of checkmarks indicates severity (e.g., \checkmark\checkmark\checkmark~= all 3 translation axes degenerate).}
  \end{flushleft}
    \vspace{-3em}
\end{table}

\begin{figure}[!t]
    \centering
    \includegraphics[width=0.45\textwidth]{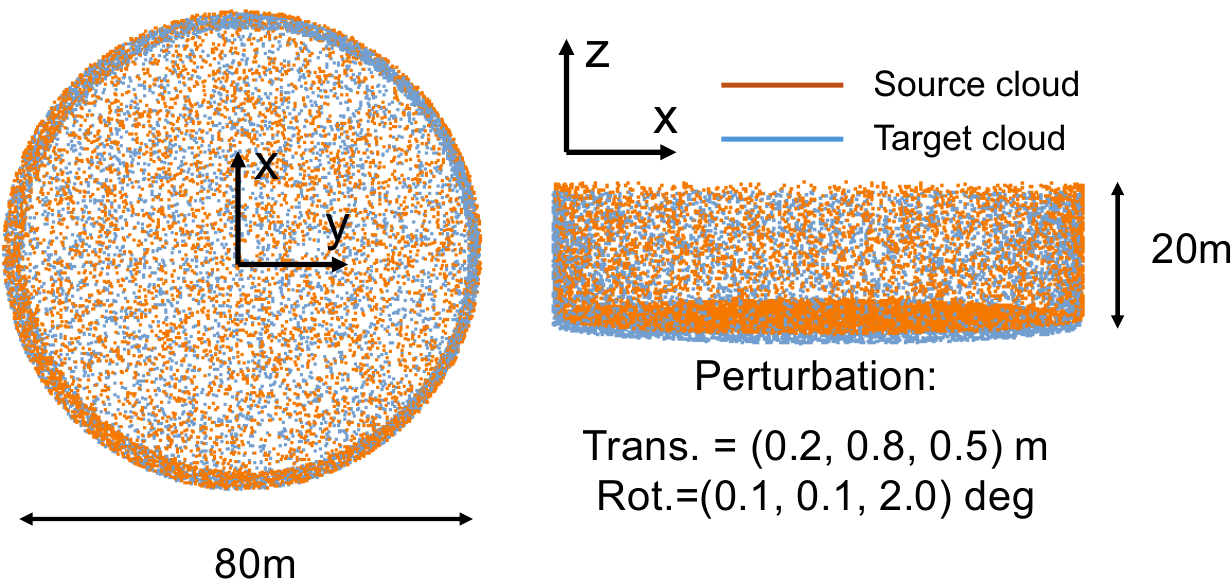}
\caption{Symmetric cylinder scene for simulated experiments. \textbf{(Left)} Top-down view and \textbf{(Right)} side view of the cylindrical point cloud containing \SI{7600}{} points. Self-registration is performed with an small perturbation consisting of a fixed rotation around the $Z$-axis and a large translation along $Z$, creating a scenario with multi-DOF degeneracy.}
\label{fig:cylinder_scene}
\vspace{-1.5em}
\end{figure}

\section{Experiments}\label{sec:exp}

\subsection{Experimental Setup}

\subsubsection{Datasets and Experimental Scenarios}\label{sub_sec:dataset_scenarios}

We evaluate \texttt{DCReg} on both simulated and real-world datasets encompassing three distinct categories for baseline comparison: (1) insufficient geometric constraints in confined environments (stairs, caves), (2) compound degeneracy (corridors, parking lot), and (3) well-constrained scenarios (building).

\textbf{Simulated Dataset:} Following the protocol established in~\cite{tuna2024informed}, we employ a symmetric cylinder scenario (\Cref{fig:cylinder_scene}) with controlled perturbations to analyze convergence behavior under certain degeneracy conditions. The perturbed point cloud is registered with the original to obtain ground truth transformations with absolute accuracy.

\textbf{Real-world Datasets:} Our evaluation leverages public benchmarks including FusionPortable (FP)~\cite{jiao2022fusionportable, wei2024fusionportablev2}, GEODE~\cite{chen2024heterogeneous}, and SubT-MRS~\cite{zhao2024superloc, zhao2024subt}, supplemented by self-collected sequences (MS) captured using a Pandar XT-$32$ LiDAR with RTK-INS ground truth. To induce more controlled degeneracy conditions for comprehensive analysis, we artificially limit the LiDAR sensing range in some sequences (e.g., parking lot and corridor). Dataset characteristics are summarized in \cref{tab:degenerate_scenarios} and \Cref{fig:sensor_setup}.

\subsubsection{Baseline and Evaluation Metrics}
We compare against state-of-the-art methods addressing ill-conditioned registration, categorized by their approach:
\begin{itemize}
\item \textbf{Detection methods:} ME (Minimum Eigenvalue)~\cite{Zhang2016On}, FCN (Full Condition Number)~\cite{hinduja2019degeneracy}, CN (Relative Condition Number)~\cite{hu2024paloc}.
\item \textbf{Mitigation strategies:} SR (Solution Remapping)~\cite{Zhang2016On}, TReg (Tikhonov Regularization)~\cite{tuna2024informed}, TSVD (Truncated SVD)~\cite{tuna2024informed}
\item \textbf{Complete solutions:} ME-SR~\cite{Zhang2016On}, FCN-SR~\cite{hinduja2019degeneracy},  {ME-TSVD~\cite{tuna2024informed}}, ME-TReg~\cite{tuna2024informed},  X-ICP~\cite{tuna2023x}, O3D ~\cite{Zhou2018o3d}.
\end{itemize}
 {We focus on methods performing \textit{active degeneracy handling} that explicitly detect and mitigate ill-conditioning. Global registration methods like~\cite{yang2020teaser, qin2023geotransformer, chen2022sc2} are excluded as they address coarse alignment rather than degeneracy-aware local registration; supplementary experiment results are also provided\footnote{\url{https://github.com/JokerJohn/DCReg/blob/main/paper/supp.pdf}}.}

\begin{table*}[htbp]
\centering
\caption{Performance Comparison of Registration Algorithms on Simulated Ill-Conditioned Cylinder}
\label{tab:registration_performance_professional}
\footnotesize
\setlength{\tabcolsep}{10pt}
\renewcommand{\arraystretch}{1.2}
\setlength{\arrayrulewidth}{1pt}
\begin{tabular}{@{}l|ccc|cccc|cc@{}}
\toprule
\multirow{2}{*}{\textbf{Methods}} & \multicolumn{3}{c|}{\textbf{Error Metrics}} & \multicolumn{4}{c|}{\textbf{ICP Metrics}} & \multicolumn{2}{c}{\textbf{Degeneracy Mask}} \\
\cmidrule(lr){2-4} \cmidrule(lr){5-8} \cmidrule(l){9-10}
& \textbf{Trans.} & \textbf{Rot.} & \textbf{CD} & \textbf{RMSE} & \textbf{Iter.} & \textbf{Fit.} & \textbf{Time} & \textbf{Trans.} & \textbf{Rot.} \\
& (\SI{}{cm}) $\downarrow$ & (\SI{}{deg}) $\downarrow$ & (\SI{}{cm}) $\downarrow$ & (\SI{}{cm}) $\downarrow$ & (\#) $\downarrow$ & (\%) $\uparrow$ & (\SI{}{ms}) $\downarrow$ & (\#) & (\#) \\
\midrule
O3D & {\textit{--}} & {\textit{--}} & {\textit{--}} & {\textit{--}} & 30 & {\textit{--}} & 46.49 & \textit{--} & \textit{--} \\
% SC2 & {96.07} & \cellcolor{blue!10}{0.07} & {109.31} & 14.06 & {\textit{--}} & 9.15 & 401.13 & 30 & \textit{--} \\
% SC2-FH & {95.97} & {2.00} & \cellcolor{blue!30}{2.53} & \cellcolor{blue!30}{3.19} & 30 & 11.54 & 394.78 & \textit{--} & \textit{--} \\
% GeoTrans & {95.91} & {{1.99}} & {18.73} & \cellcolor{blue!10}{3.19} & {\textit{--}} & \cellcolor{blue!30}{11.56} & 379.81 & \textit{--} & \textit{--} \\
\midrule
ME-SR & 96.19 & 3.66 & 92.30 & 19.73 & 30 & 2.25 & 19.40 & \texttt{111} & \texttt{000} \\
ME-TSVD & 2.91 & 0.97 & 34.81 & 14.57 & 30 & 24.85 & 19.81 & \texttt{111} & \texttt{000} \\
ME-TReg & 22.18 & \cellcolor{blue!10}{0.11} & \cellcolor{blue!10}{21.96} & \cellcolor{blue!10}4.18 & 30 & \cellcolor{blue!10}{34.96} & 20.05 & \texttt{111} & \texttt{000} \\
FCN-SR & 96.24 & 3.76 & 93.12 & 19.58 & 12 & 2.43 & 29.74 & \texttt{111} & \texttt{100} \\
\midrule
SuperLoc & 23.82 & 2.81 & 71.37 & 43.94 & {\textit{--}} & 2.29 & \cellcolor{blue!10}{16.98} & \texttt{001} & \texttt{000} \\
X-ICP & \cellcolor{blue!30}{0.42} & 2.86 & 55.88 & 18.27 & \cellcolor{blue!30}{9} & 10.08 & 43.06 & \texttt{001} & \texttt{000} \\
\midrule
\textbf{Ours} & \cellcolor{blue!10}{2.71} & \cellcolor{blue!30}{0.05} & \cellcolor{blue!30}{3.29} & \cellcolor{blue!30}{3.16} & \cellcolor{blue!10}{10} & \cellcolor{blue!30}{100} & \cellcolor{blue!30}{7.79} & \texttt{001} & \texttt{000} \\
\bottomrule
\end{tabular}
\begin{flushleft}
 \footnotesize{\textit{Note}: Trans. = Translation Error, Rot. = Rotation Error, RMSE/Iter./Fit. = {Point-to-Plane ICP Residuals/Iteration Count/Fitness Score}. Degeneracy Mask: Binary indicators where \texttt{1}/\texttt{0} represent for degenerate/well-constrained direction. Best/second-best results: \colorbox{blue!30}{blue}/\colorbox{blue!10}{light blue}. Unavailable data shown in {\textit{--}}.}
\end{flushleft}
    \vspace{-1.5em}
\end{table*}

We employ Absolute Trajectory Error (ATE)~\cite{grupp2017evo} to evaluate localization accuracy. 
For mapping quality assessment~\cite{hu2025humapeval}, we adopt two complementary metrics: Registration Accuracy (AC) and Chamfer Distance (CD). 
AC quantifies geometric precision by computing the root-mean-square distance between corresponding inlier points after dense alignment with the ground-truth map. Inliers are defined as point pairs whose Euclidean distance is below \SI{0.2}{m}. 
The ratio of inlier points to the source cloud points defines the fitness score (Fit.), also referred to as overlap or completeness. 
For comparable fitness scores, lower AC values indicate higher geometric accuracy.
CD measures the bidirectional average nearest-neighbor distance between the estimated and ground-truth maps, i.e., averaging the nearest-neighbor distances from estimated to ground truth and from ground truth to estimated. Smaller CD values reflect better overall map quality in terms of both accuracy and completeness.
 Furthermore, we report the Degeneracy Ratio (DR), which quantifies the percentage of frames identified as degenerate. DR serves as an indicator of detection sensitivity rather than an error metric.
Additionally, in simulated experiments (\cref{sec:simulated_sg_expri}), we also report the point-to-plane ICP performance metrics in \cref{tab:registration_performance_professional}.
% This is to explicitly demonstrate a core characteristic of geometric degeneracy: algorithms may yield deceptively perfect matching metrics while actually converging to incorrect local minima.

\subsubsection{Implementation Details}

 {Most of the baseline methods are implemented within a unified point-to-plane ICP framework based on established open-source codes.\footnote{\url{https://github.com/laboshinl/loam_velodyne}} 
Unless otherwise specified, we strictly follow the default linear solver used in these implementations, and the normal equations are solved by Eigen's \texttt{ColPivHouseholderQR} function. 
Since different algorithms employ varying optimization backends, XICP\footnote{\url{https://github.com/leggedrobotics/perfectlyconstrained}} and superLoc\footnote{\url{https://github.com/JokerJohn/SuperOdom-M}} utilize Ceres for nonlinear optimization, while Open3D\footnote{\url{https://github.com/isl-org/Open3D}} leverages its native implementation with OpenMP parallelization. 
Our \texttt{DCReg} implementation, built upon~\cite{hu2024paloc}, employs Eigen with Intel TBB for efficient parallel computation.}
% Most of the baseline methods are implemented within a unified point-to-plane ICP framework based on established open-source codes$\footnote{\url{https://github.com/laboshinl/loam_velodyne}}$ with OpenMP parallelization.
%  {Unless otherwise specified, we strictly follow the default linear solver used in these open-source implementations, and the normal equations are solved by Eigen's \texttt{ColPivHouseholderQR} function.} 
% Since different algorithms employ varying optimization backends, XICP$\footnote{\url{https://github.com/leggedrobotics/perfectlyconstrained}}$ and SuperLoc$\footnote{\url{https://github.com/JokerJohn/SuperOdom-M}}$ utilize Ceres, while Open3D$\footnote{\url{https://github.com/isl-org/Open3D}}$ leverages its native implementation with OpenMP parallelization . 
% Our \texttt{DCReg} implementation, built upon~\cite{hu2024paloc}, employs Eigen with Intel TBB for efficient parallel computation. 
\cref{tab:method_comparison} summarizes the technical specifications of each approach.
 {For system-level evaluation, we integrate each registration method into a \SI{5}{\second} sliding-window fixed-lag smoother~\cite{hu2024paloc, dellaert2012factor} coupled with LiDAR-inertial odometry~\cite{xu2022fast}. Degeneracy detection and mitigation are applied at every scan-to-map registration process to comprehensively assess their impact on long-duration localization and mapping performance. 
We maintain consistent configuration parameters across all methods, including correspondence search radius (\SI{1.0}{m}) and normal estimation ($5$ points).
All algorithms adopt consistent hyperparameters according to the work~\cite{tuna2023x, tuna2024informed}: a condition number threshold of $\kappa_{th} = 10$, minimum eigenvalue threshold of $120$, and convergence criteria of $10^{-3}$\,m for translation and $10^{-5}$\,rad for rotation. For \texttt{DCReg}, the target condition number is set to $\kappa_{\text{tg}} = 10$. The max iteration for ICP is set to $30$ for simulated experiments and $10$ for real-world experiment. The regularization weight for Treg is fixed at $\lambda = 100$. For the XICP, we set the threshold for information sufficiency to $300$ and for information deficiency to $150$. The maximum and minimum alignment angle limits are specified as $45^\circ$ and $60^\circ$, respectively.
All experiments are conducted on an Intel i$7$-$12700$K CPU with $96$\,GB RAM.}

% \begin{figure}
%     \centering
%     \includegraphics[width=0.48\textwidth]{figures/performance_matrix_journal-crop.pdf}
% \caption{Multi-metric error analysis across different ICP methods in degenerate scenarios (\cref{sec:simulated_sg_expri}). The heatmap displays various error metrics, with color intensity from white (low) to blue (high) indicating error magnitude. In degenerate configurations, multiple local minima exist where lower translation/rotation errors and ICP residuals do not necessarily indicate better registration quality. Our method (bottom row) achieves both correct convergence and superior accuracy across all other methods.}
% \label{fig:error_matrics}
%     \vspace{-0.5em}
% \end{figure}

\subsection{Controlled Simulation Analysis}\label{sec:simulated_sg_expri}
% We first validate our theoretical contributions using the cylinder scenario with a small perturbation (Figure~\ref{fig:cylinder_scene}).
We introduce a sufficiently small initial pose perturbation to induce severe degeneracy in specific translational and rotational dimensions in \Cref{fig:cylinder_scene}.

% \textbf{Experimental Setup:}
% We designed controlled simulation experiments using the scenario illustrated in Figure.~\ref{fig:cylinder_scene} A perfectly symmetric cylinder was generated with radius and height within typical LiDAR scanning range, and point density registration single-frame LiDAR acquisitions. By registering the point cloud against itself with prescribed initial perturbations, we systematically created multi-DOF degenerate scenarios. Following ~\cite{tuna2024informed}, we introduced a substantial initial misalignment: \SI{50}{\degree} rotation around the z-axis and translation of (\SI{0.2}{}, \SI{0.2}{}, \SI{1.25}{}) \SI{}{m}. This configuration enables controlled manipulation of the Hessian matrix's eigenvalues and condition numbers, facilitating rigorous analysis of convergence behavior, accuracy, and robustness across different methods during the iterative registration process.

\begin{figure}
    \centering
    \includegraphics[width=0.48\textwidth]{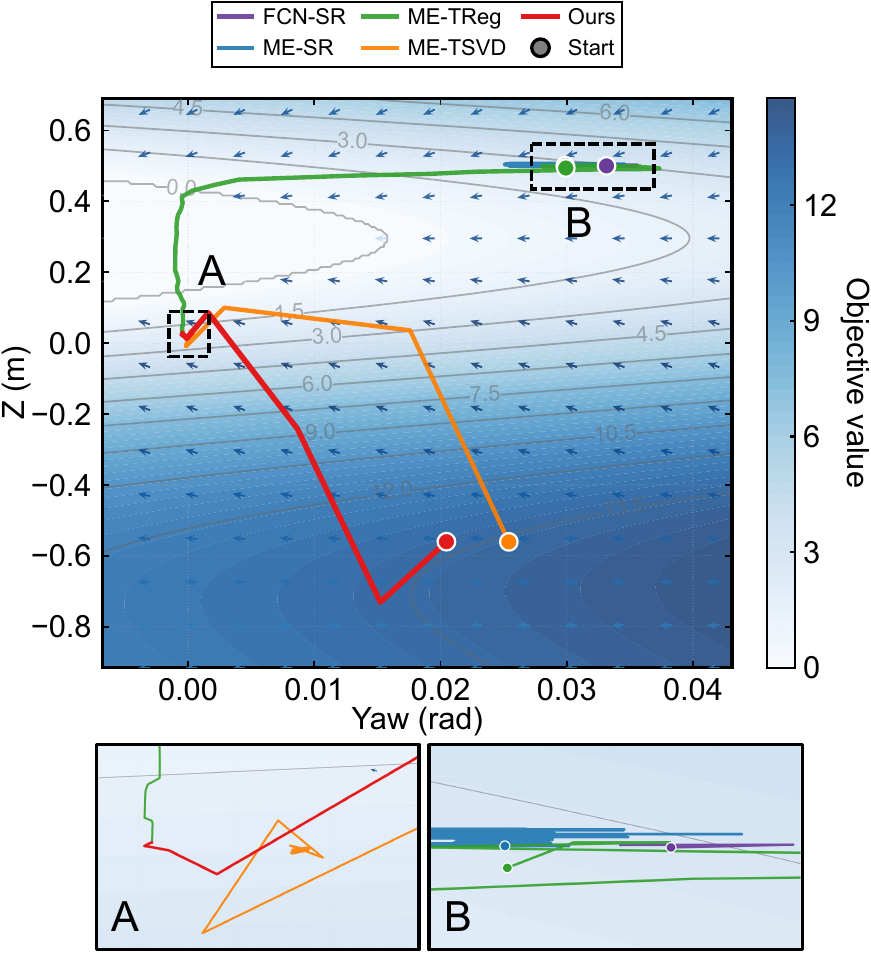}
\caption{Optimization landscape with gradient field visualization in the cylinder scenario after \SI{5000}{} iterations. Arrows indicate negative gradient directions, with optimization trajectories overlaid. Our method demonstrates the most rapid convergence.
% Baseline methods exhibit oscillatory or local minima despite strict convergence criteria. In contrast, our method demonstrates direct convergence to the global minimum with stable terminal behavior, validating its convergence guarantees (\cref{subsec:convergence_analysis}).
}
\label{fig:optimization_paths}
\vspace{-1.5em}
\end{figure}

\subsubsection{Convergence Analysis and Optimization Behavior}\label{subsec:convergence_analysis}

% Table\cref{tab:registration_performance_professional} presents comprehensive quantitative results. In degenerate scenarios, point cloud registration often exhibits multiple local minima, making single metrics (e.g., ICP residual, translation/rotation errors) insufficient to characterize success. While our method does not achieve the lowest values in every individual metric, translation error, rotation error, ICP residual, iteration count, correspondence ratio, or computation time, it demonstrates superior overall performance.
% Notably, O3D achieves minimal translation error and maximal correspondence ratio, yet suffers from \SI{1.86}{\degree} rotation error after 30 iterations, indicating convergence to a local minimum. ME-SR and ME-TReg fail to converge entirely, while FCN-SR and ME-TSVD converge with substantially larger pose errors. Critically, our method achieves the best point-to-point RMSE and CD, more reliable indicators of registration quality while maintaining superior pose accuracy and computational efficiency.

\begin{figure*}
    \centering
    \includegraphics[width=0.95\textwidth]{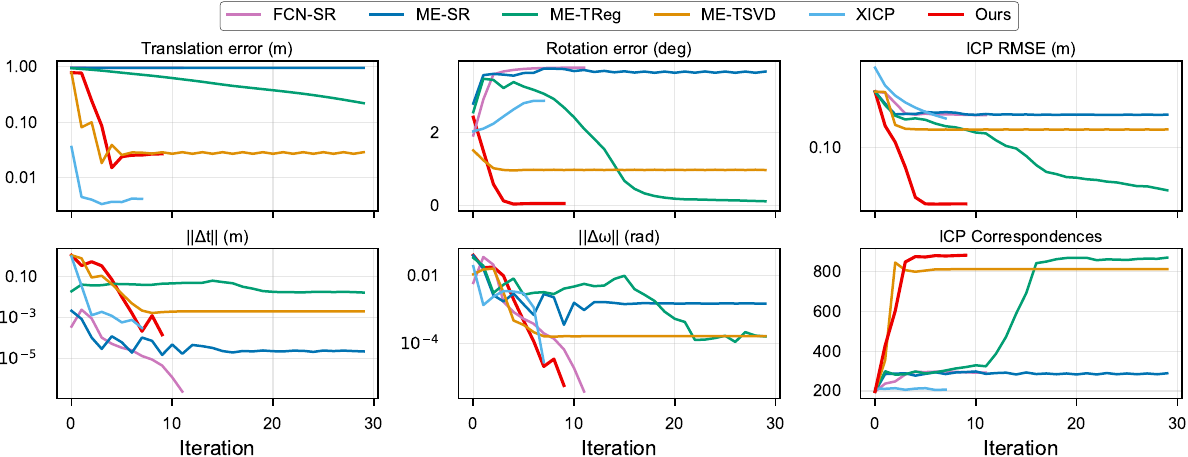}
\caption{ {Iterative convergence process under cylindrical degeneracy. Plots show translation/rotation errors, update magnitudes, ICP residuals, and correspondence counts over iterations. Our method (\rt{red}) exhibits faster convergence, smaller pose errors, and more stable correspondence growth than all baselines, validating its robustness in degenerate configurations.}
}
\label{fig:iterative_perfor_overall}
    \vspace{-1.5em}
\end{figure*}

\cref{tab:registration_performance_professional} presents comprehensive quantitative results for the cylinder scenario. 
% The table includes both ICP performance and error metrics. 
% Point cloud registration frequently exhibits multiple local minima due to geometric ambiguities or initial pose, rendering single metrics (e.g., ICP residual, translation/rotation errors) insufficient for characterizing registration quality.
The standard O3D lacks degeneracy awareness entirely. ME-SR, ME-TSVD, and ME-TReg fail to converge, while FCN-SR converges to a local minimum. SuperLoc exhibits substantial translation and rotation errors, indicating registration failure. Although XICP achieves the smallest translation error, its large rotation error and high Chamfer distance confirm misalignment.
In contrast, our method achieves optimal performance across rotation error, Chamfer distance, ICP residual, and fitness metrics, requiring only the second-fewest iterations for successful convergence. Furthermore, our algorithm demonstrates $2-10$ times efficient compared to baseline methods. These results validate that degeneracy detection coupled with targeted mitigation strategies effectively influences convergence and accuracy throughout the iterative process.

Regarding degeneracy detection, ME-based methods focus exclusively on absolute information while neglecting translation-rotation scale disparity, leading to over-detection in translational part. FCN-based approaches overlook coupling effects, resulting in simultaneous over-detection in both translation and rotation. More critically, the degeneracy mitigation strategy directly influences optimization trajectories during iterations. SR handle degeneracy too conservatively, TSVD discards information from multiple dimensions entirely, and TReg processes all dimensions uniformly, inadvertently introducing noise into well-conditioned dimensions. Our method uniquely achieves correct convergence.

To validate the convergence properties of our proposed method, we establish stringent convergence criteria ($1 \times 10^{-8}$~\si{m} for translation, $1 \times 10^{-10}$~\si{rad} for rotation) to ensure sufficient iterations for comprehensive analysis of the optimization landscape. \Cref{fig:optimization_paths} illustrates the optimization trajectories for $z$ and yaw dimensions. Only our method and TReg converge to the same global minimum, with our approach requiring $16$ iterations compared to TReg's approximately $160$ iterations, demonstrating superior convergence efficiency.

\begin{table*}[htbp]
\centering
\caption{Comprehensive Performance Evaluation on Real-world Scenarios}
\label{tab:map_accuracy_comparison}
\footnotesize
\setlength{\tabcolsep}{3pt}
\renewcommand{\arraystretch}{1.1}
\begin{tabular}{@{}l||c|cc|c||c|cc|c||c|cc|c@{}}
\toprule
& \multicolumn{4}{c||}{\textbf{Stairs}} & \multicolumn{4}{c||}{\textbf{Corridor}} & \multicolumn{4}{c}{\textbf{Building}} \\
& \multicolumn{4}{c||}{\scriptsize{3-5k pts/frame}} & \multicolumn{4}{c||}{\scriptsize{1-2k pts/frame}} & \multicolumn{4}{c}{\scriptsize{ {25-30k pts/frame}}} \\
& \multicolumn{4}{c||}{\scriptsize{128M pts/map}} & \multicolumn{4}{c||}{\scriptsize{67M pts/map}} & \multicolumn{4}{c}{\scriptsize{ {95M pts/map}}} \\
\cmidrule{2-5} \cmidrule{6-9} \cmidrule{10-13}
\textbf{Method} & DR & \makebox[0.9cm][c]{ATE $\downarrow$} & \makebox[0.9cm][c]{AC $\downarrow$} & \makebox[1cm][c]{Time $\downarrow$} & DR & \makebox[0.9cm][c]{ATE $\downarrow$} & \makebox[0.9cm][c]{AC $\downarrow$} & \makebox[1cm][c]{Time $\downarrow$} & DR & \makebox[0.9cm][c]{ATE $\downarrow$} & \makebox[0.9cm][c]{AC $\downarrow$} & \makebox[1cm][c]{Time $\downarrow$} \\
& \scriptsize{(\%)} & \scriptsize{(\SI{}{cm})} & \scriptsize{(\SI{}{cm})} & \scriptsize{(\SI{}{ms})} & \scriptsize{(\%)} & \scriptsize{(\SI{}{cm})} & \scriptsize{(\SI{}{cm})} & \scriptsize{(\SI{}{ms})} & \scriptsize{(\%)} & \scriptsize{(\SI{}{cm})} & \scriptsize{(\SI{}{cm})} & \scriptsize{(\SI{}{ms})} \\
\midrule
Odom & {--} & 25.18 & {--} & \cellcolor{blue!10}16.79 & {--} & 26.34 & {--} & 49.10 & {--} &  {16.53} & {--} &  {101.83} \\
\midrule
O3D & {--} & 80.51 & 7.23 & 889.69 &  {--} & 1418.37 & 7.09 & 15.02 & {--} &  {9.42} &  {4.89} &  {385.64} \\
% GeoTrans & {--} & 3320.69 & {--}  & 1911.32 & {--} & 4841.94 & {--} & 626.87 & {--} & 2907.36 & {--} & 2032.53 \\
\midrule
ME-SR & 18.91 & 160.75 & 7.90 & 129.19 & 39.27 & 173.82 & 4.55 & 13.31 &  {0.00} & \cellcolor{blue!30} {8.58} &  {4.62} & \cellcolor{blue!10} {65.91} \\
ME-TSVD & 9.26 & \cellcolor{blue!10}6.44 & 6.09 & 56.16 & 34.84 & 94.32 & 4.36 & 11.60 &  {0.00} &  {8.64} &  {4.62} &  {66.94} \\
ME-TReg & 9.20 & 7.16 & \cellcolor{blue!10}6.06 & 59.41 & 33.21 & \cellcolor{blue!10}26.24 & \cellcolor{blue!10}3.62 & \cellcolor{blue!10}8.22 &  {0.00} &   \cellcolor{blue!10} {8.63} & \cellcolor{blue!10} {4.61} &  {66.94} \\
FCN-SR & 94.77 & 280.07 & 7.80 & 207.13 & 85.28 & 231.40 & 6.64 & 38.17 &  {99.23} &  {299.46} &  {{--}} &  {786.45} \\
% \midrule
% SuperLoc$^{4}$ & 0 & 0 & 0 & 0 & 0 & 0 & 0 & 0 & 0 & 0 & 0 & 0 \\
% X-ICP$^{2}$ & 0 & 0 & 0 & 0 & 0 & 0 & 0 & 0 & 0 & 0 & 0 & 0 \\
\midrule
\textbf{Ours} & 50.71 & \cellcolor{blue!30}\textbf{3.96} & \cellcolor{blue!30}\textbf{5.55} & \cellcolor{blue!30}\textbf{6.47} & 58.04 & \cellcolor{blue!30}\textbf{7.44} & \cellcolor{blue!30}\textbf{3.45} & \cellcolor{blue!30}\textbf{1.24} &  {0.00} &  {8.65} & \cellcolor{blue!30} {4.60} & \cellcolor{blue!30} {6.78} \\
\bottomrule
\end{tabular}
\vspace{0.2pt}
\begin{flushleft}
\scriptsize{\textit{Note}: Best and second-best results are highlighted in \colorbox{blue!30}{blue} and \colorbox{blue!10}{light blue}. \textbf{DR}: degeneracy ratio.}
\end{flushleft}
\vspace{-3.0em}
\end{table*}

\subsubsection{Condition Number Evolution}

\begin{figure}
    \centering
    \includegraphics[width=0.48\textwidth]{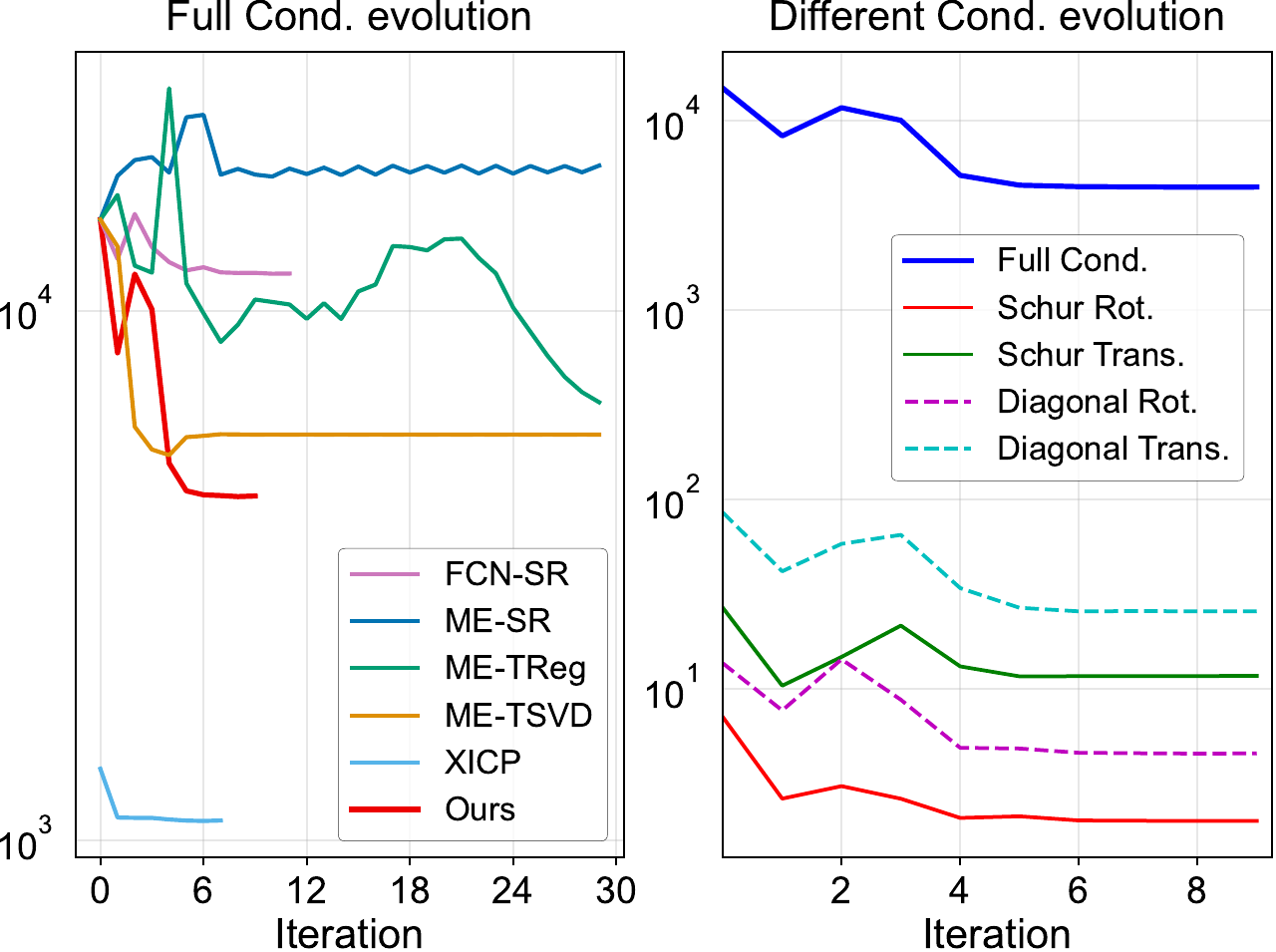}
\caption{ {Condition number evolution over iteration in the cylinder degeneracy scenario. \textbf{Left}: Full Hessian condition numbers for all methods. \textbf{Right}: Schur and diagonal condition numbers for our approach.} 
}
\label{fig:simu_iterative_cond_evo}
    \vspace{-1.5em}
\end{figure}

As established in \cref{subsec:condition_define}, the condition number directly reflects convergence speed. 
\Cref{fig:iterative_perfor_overall} tracks the ICP behavior over iteration of different algorithms. Both translation and rotation errors decrease rapidly before stabilizing at accurate levels. Correspondingly, the update magnitudes in both rotation and translation diminish sharply, demonstrating how our optimization naturally balances the inherent scale disparity between rotational and translational parameters during the iterative process.
\Cref{fig:simu_iterative_cond_evo} compares the condition number evolution across different algorithms during iterations, alongside the relationship between subspace and full Hessian condition numbers for \textit{DCReg}. The results demonstrate that: 1) \textit{DCReg} achieves rapid condition number reduction compared to other methods, enabling faster convergence; 2) The Schur complement and full Hessian condition numbers exhibit nearly identical trends, confirming that controlling subspace conditioning directly influences the full Hessian space and optimization direction. The translational Schur complement condition number rapidly decreases to our target value $\kappa_{tg} = 10$ after degeneracy mitigation, validating the effectiveness of our mitigation strategy.

\subsection{Real-world Performance Evaluation}

We present comprehensive evaluation results comparing \textit{DCReg} against baseline across diverse real-world scenarios.
% The assessment encompasses trajectory accuracy reflecting motion precision at individual timestamps, mapping accuracy demonstrating scene reconstruction fidelity, degeneracy occurrence rates indicating scenario complexity, and per-frame matching times revealing the cumulative performance impact of degeneracy handling over extended operation periods.

\subsubsection{Long-duration Localization Performance}

\textbf{Localization}: 
\cref{tab:trajectory_accuracy_comparison} presents results from challenging  {cave scenario} characterized by extractable features but constrained passages using $16$-beam LiDAR. These scenarios are particularly susceptible to constraint imbalance degeneracy when initial poses deviate or registration not converge. 
The cave02 involve aggressive motion patterns where most baseline methods fail entirely, yet \textit{DCReg} achieves optimal localization accuracy.
The parking lot scenario provides sufficient information but we limit the LiDAR scanning range to \SI{10}{m}, yet ME fail to detect degeneracy, highlighting fundamental limitations in eigenvalue-based detection. This approach systematically overlooks rotational degeneracies due to scale disparity, as rotational eigenvalues typically maintain higher magnitudes than translational counterparts (\cref{subsec:structural_ill_conditioning}).

% =======
\begin{table}[!t]
\centering
\caption{Comprehensive Performance Evaluation on Real-world Degenerate Scenarios}
\label{tab:trajectory_accuracy_comparison}
\footnotesize
\setlength{\tabcolsep}{3.5pt}
\renewcommand{\arraystretch}{1.1}
\begin{tabular}{@{}l||c|c|c||c|c|c@{}}
\toprule
& \multicolumn{3}{c||}{\textbf{Cave02}} & \multicolumn{3}{c}{\textbf{Parking Lot}} \\
& \multicolumn{3}{c||}{\scriptsize{3-5k pts/frame}} & \multicolumn{3}{c}{\scriptsize{4-8k pts/frame}} \\
& \multicolumn{3}{c||}{\scriptsize{53M pts/map}} & \multicolumn{3}{c}{\scriptsize{218M pts/map}} \\
\cmidrule{2-4} \cmidrule{5-7}
\textbf{Method} & DR & \makebox[0.9cm][c]{ATE $\downarrow$} & \makebox[1cm][c]{Time $\downarrow$} & DR & \makebox[0.9cm][c]{ATE $\downarrow$} & \makebox[1cm][c]{Time $\downarrow$} \\
& \scriptsize{(\%)} & \scriptsize{(\SI{}{cm})} & \scriptsize{(\SI{}{ms})} & \scriptsize{(\%)} & \scriptsize{(\SI{}{cm})} & \scriptsize{(\SI{}{ms})} \\
\midrule
Odom & {--} & 129.78 & 22.39 & {--} & 32.98 & 22.31 \\
\midrule
O3D & {--} & 16.88 & 328.53 & {--} & 36.89 & 345.10 \\
% GeoTrans & {--} &  3523.82 & 1942.73 & {--} & 2421.66 & 1684.25\\
\midrule
ME-SR & 6.47 & 565.61 & 146.80 & 89.59 & 62.11 & 16.01 \\
ME-TSVD & 0.83 & 497.05 & 116.54 & 87.69 & \cellcolor{blue!10}30.46 & \cellcolor{blue!10}13.96 \\
ME-TReg & 19.03 & 209.53 & \cellcolor{blue!10}115.87 & 87.24 & 47.07 & 11.40 \\
FCN-SR & 58.89 & 491.29 & 232.51 & 98.15 & 64.50 & 17.59 \\
% \midrule
% SuperLoc$^{4}$ & 0 & 0 & 0 & 0 & 0 & 0 \\
% X-ICP$^{2}$ & 0 & 0 & 0 & 0 & 51.17 & 18.21 \\
\midrule
\textbf{Ours} & 32.90 & \cellcolor{blue!30}\textbf{4.57} & \cellcolor{blue!30}\textbf{3.71} & 86.74 & \cellcolor{blue!30}\textbf{29.56} & \cellcolor{blue!30}\textbf{2.11} \\
\bottomrule
\end{tabular}
\vspace{0.5pt}
\begin{flushleft}
\scriptsize{\textit{Note}: Best/second-best results: \colorbox{blue!30}{blue}/\colorbox{blue!10}{light blue}. \textbf{DR}: degeneracy ratio.}
\end{flushleft}
\vspace{-3.0em}
\end{table}

\textbf{Mapping}: 
\cref{tab:map_accuracy_comparison} extends the evaluation to include both trajectory and mapping accuracy across additional scenarios. In the constrained stair environment, FCN's over-detection strategy leads to matching failure, while ME's insufficient detection yields noticeable accuracy improvements under TSVD and TReg frameworks compared to odometry-only approaches, though still inferior to \textit{DCReg} performance. The corridor scenario reveals TReg's inadequate regularization strength, providing minimal improvement over odometry baselines, while 	\textit{DCReg} maintains optimal accuracy. In building environments, ME methods fail to detect mild degeneracies that 	\textit{DCReg} successfully identifies and mitigates, resulting in measurable accuracy gains.
Across all evaluated datasets, 	\textit{DCReg} demonstrates substantial computational advantages, achieving  {typical $5-30\times$ (up to $116\times$ in certain scenarios)} speedup compared to baseline approaches. This performance enhancement stems from more reliable convergence characteristics and reduced iteration requirements through targeted degeneracy mitigation.  {
The building sequence demonstrates that in well-constrained scenario without degeneracy, most algorithms exhibit  similar performance. However, FCN-SR still suffers from over-detection due to scale disparity and coupling effects.
}

\begin{figure}
    \centering
    \includegraphics[width=0.45\textwidth]{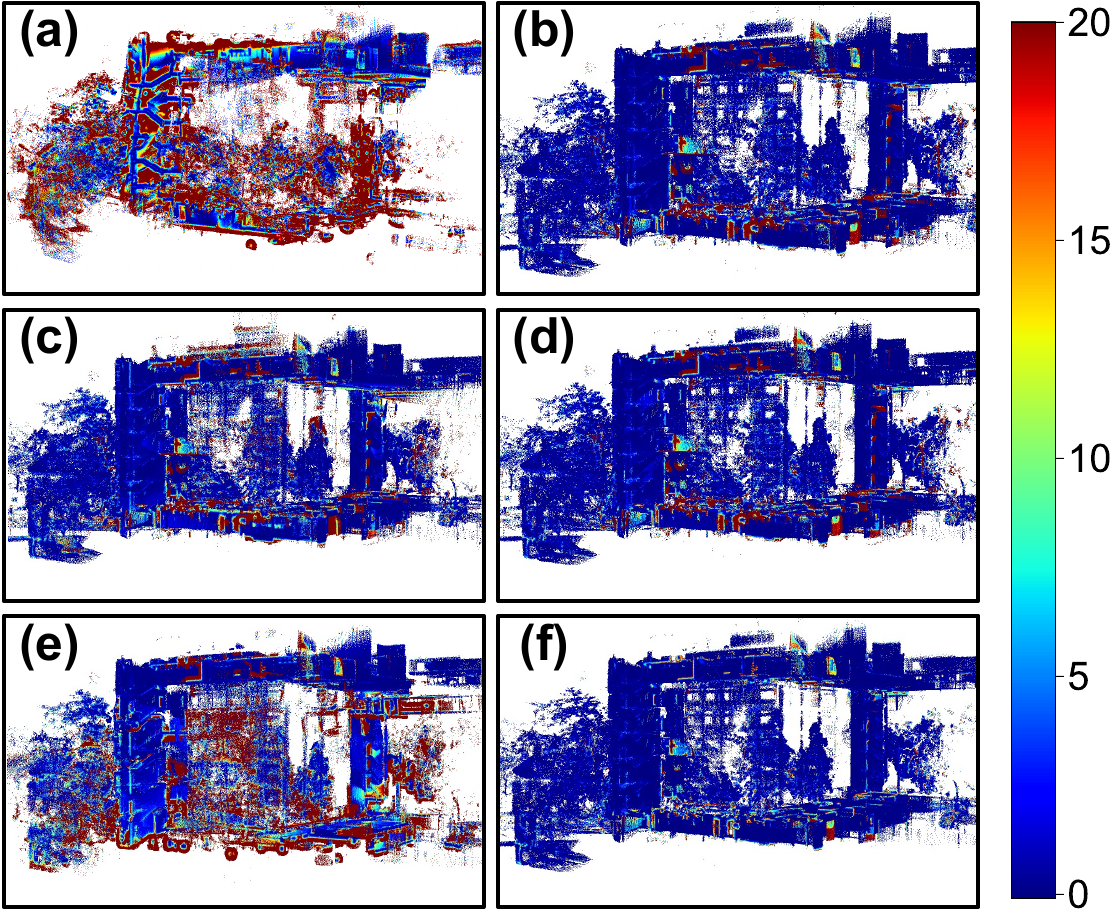}
\caption{Map error comparison across different methods in the stairs scenario. \textbf{(a)}:FCN-SR, \textbf{(b)}:ME-TReg, \textbf{(c)}:O3D, \textbf{(d)}:ME-TSVD, \textbf{(e)}:ME-SR, \textbf{(f)}:\textbf{Ours}. The error maps use a blue-to-red scale indicating increasing error magnitude. \textit{DCReg} exhibits predominantly blue regions compared to others, demonstrating superior mapping accuracy.} 
\label{fig:map_eval}
\vspace{-0.5em}
\end{figure}

\Cref{fig:map_eval} visualizes mapping accuracy comparison in the stair scenario using error color map, where blue-to-red gradients indicate increasing mapping errors. \textit{DCReg} exhibits significantly larger blue regions compared to alternative methods, corroborating the quantitative results presented in the tables and demonstrating superior spatial accuracy preservation throughout the localization process.

\subsubsection{Computational Efficiency Analysis}\label{sub_sec:run_time}

 {Degeneracy handling algorithms significantly impact computational performance through two primary mechanisms. First, unreliable degeneracy handling in single-frame registration can compromise initial estimates for subsequent iterations, leading to increased iteration counts and potential convergence failures. Second, in sequential SLAM applications, poor single-frame registration results propagate to affect both accuracy and efficiency of subsequent frame processing.}
Our computational analysis encompasses four complementary perspectives: system-level average registration times across all frames (\cref{tab:map_accuracy_comparison} and \cref{tab:trajectory_accuracy_comparison}), scalability analysis with varying point cloud densities (\cref{tab:dcreg_efficiency}), component-wise timing breakdown (\cref{tab:dcreg_time_breakdown}), and comparative analysis of degenerate versus non-degenerate frame processing (\cref{tab:trajectory_accuracy_comparison}).

% typical $6--30\times$ (up to $150\times$ in certain scenarios) over comparable degeneracy-aware 
% baseline methods.

\begin{table}[!t]
\centering
\caption{Computational Efficiency w.r.t. Point Cloud Density}
\label{tab:dcreg_efficiency}
\footnotesize
\setlength{\tabcolsep}{5pt}
\renewcommand{\arraystretch}{1.1}
\begin{tabular}{@{}c||c|c|c@{}}
\toprule
& \multicolumn{3}{c}{\textbf{Parking Lot}} \\
& \multicolumn{3}{c}{\scriptsize{Map: 218M pts}} \\
\cmidrule{2-4}
\textbf{LiDAR Range} & \textbf{Pts.} & \textbf{DR} & \textbf{Time $\downarrow$} \\
\scriptsize{(\SI{}{m})} & \scriptsize{(\SI{}{k})} & \scriptsize{(\%)} & \scriptsize{(\SI{}{ms})} \\
\midrule
10 & 1-1.5 & 86.74 & 2.11 \\
20 & 2-2.5 & 65.0 & 3.40 \\
30 & 3-7.5 & 50.29 & 4.35 \\
40 & 4-8.5 & 37.00 & 4.96 \\
50 & 5-9 & 33.81 & 5.13 \\
60 & 5-10 & 31.08 & 5.99 \\
\bottomrule
\end{tabular}
\vspace{0.5pt}
\begin{flushleft}
\scriptsize{\textit{Note}: Pts. shows point count range for per-frame. DR: degeneracy ratio.}
\end{flushleft}
\vspace{-2.0em}
\end{table}

System evaluations presented in \cref{tab:map_accuracy_comparison} and \cref{tab:trajectory_accuracy_comparison} demonstrate that \textit{DCReg} achieves best computational efficiency across all datasets, realizing  {typical $5-30\times$ (up to $116\times$ in certain scenarios)} speedup compared to baseline. This performance gain stems from two factors: our detection and targeted mitigation strategy enables faster convergence, while the parallelized implementation optimizes computational efficiency.
\cref{tab:dcreg_efficiency} illustrates the relationship between sensor range, point cloud density, and computational requirements. As LiDAR range increases from $10$ to \SI{60}{m}, point cloud size grows substantially, leading to increased degeneracy frames and correspondingly higher computational costs. This analysis validates the practical importance of efficient degeneracy handling in resource-constrained scenarios.
Component-wise timing analysis in \cref{tab:dcreg_time_breakdown} reveals the computational costs across four key modules: degeneracy detection, mitigation, correspondence establishment, and Hessian matrix construction. Across multiple datasets, degeneracy detection and mitigation consume minimal computational resources while ensuring convergence. Correspondence establishment dominates processing time, followed by Hessian construction. Since correspondence quality directly depends on initial pose accuracy, these results underscore the critical role of reliable degeneracy handling in overall computational efficiency.

 {Frame-level analysis in \cref{tab:convergence_degenerate} compares ICP iteration counts and registration times for degenerate versus non-degenerate frames across different methods. Non-degenerate frames naturally require fewer iterations and less processing time due to the absence of degeneracy issues. For degenerate frames, \textit{DCReg} consistently achieves the lowest computational cost and competitive iteration counts. Notably, \textit{DCReg} also demonstrates superior performance on non-degenerate frames, indicating that robust degeneracy handling in sequential processing enhances both accuracy and efficiency for subsequent registration.}

\begin{table}[!t]
\centering
\caption{Computational Time Breakdown of DCReg}
\label{tab:dcreg_time_breakdown}
\footnotesize
\setlength{\tabcolsep}{3pt}
\renewcommand{\arraystretch}{1.1}
\begin{tabular}{@{}l||c||c|c|c|c||c@{}}
\toprule
\multirow{2}{*}{\textbf{Dataset}} & \multirow{2}{*}{\textbf{DR}} & \multicolumn{4}{c||}{\textbf{Component Time $\downarrow$}} & \textbf{Total Time $\downarrow$} \\
\cmidrule{3-6}
& & \textbf{Det.} & \textbf{Mit.} & \textbf{Corr.} & \textbf{Hess.} &  \\
& \scriptsize{(\%)} & & & & & \\
\midrule
Building & 15.09 & 0.043 & 0.012 & 3.950 & 0.530 & 4.608 \\
Cave01 & 46.53 & 0.043 & 0.011 & 3.460 & 0.380 & 3.910 \\
% Cave02 & 0 & 0.15 & 0.11 & 0.42 & 0.28 & 1.07 \\
% Cave04 & 0 & 0.13 & 0.09 & 0.38 & 0.25 & 0.96 \\
Stairs & 50.71 & 0.047 & 0.009 & 6.094 & 0.313 & 6.470 \\
Parking Lot & 86.74 & 0.064 & 0.071 & 1.662 & 0.212 & 2.105 \\
% Corridor & 72.45 & 0.24 & 0.28 & 0.75 & 0.52 & 1.93 \\
\bottomrule
\end{tabular}
\vspace{0.5pt}
\begin{flushleft}
\scriptsize{\textit{Note}: All times in \SI{}{ms}. \textbf{DR}: degeneracy ratio, \textbf{Det.}: degeneracy detection, \textbf{Mit.}: degeneration mitigation, \textbf{Corr.}: correspondence search, \textbf{Hess.}: Hessian construction.}
\end{flushleft}
\vspace{-1em}
\end{table}

\begin{table}[!t]
\centering
\caption{Convergence Performance on Degenerate vs. Non-degenerate Frames}
\label{tab:convergence_degenerate}
\footnotesize
\setlength{\tabcolsep}{5pt}
\renewcommand{\arraystretch}{1.1}
\begin{tabular}{@{}l||cc||cc@{}}
\toprule
& \multicolumn{2}{c||}{\textbf{Degenerate Frames}} & \multicolumn{2}{c}{\textbf{Non-degenerate Frames}} \\
\cmidrule{2-3} \cmidrule{4-5}
\textbf{Method} & \textbf{Iter. $\downarrow$} & \textbf{Time $\downarrow$} & \textbf{Iter. $\downarrow$} & \textbf{Time $\downarrow$} \\
& & \scriptsize{(ms)} & & \scriptsize{(ms)} \\
\midrule
\multicolumn{5}{c}{\textbf{Parking Lot}} \\
\midrule
ME-SR & 5.79 & 13.97 & 7.41 & 39.15 \\
ME-TSVD & 5.37 & 12.13 & 6.22 & 32.84 \\
ME-TReg & \cellcolor{blue!30}4.63 & \cellcolor{blue!10}10.05 & \cellcolor{blue!10}4.17 & \cellcolor{blue!10}22.13 \\
FCN-SR & 6.23 & 17.72 & $\textit{--}$ & $\textit{--}$ \\
\textbf{Ours} & \cellcolor{blue!10}\textbf{5.04} & \cellcolor{blue!30}\textbf{2.11} & \cellcolor{blue!30}\textbf{3.30} & \cellcolor{blue!30}\textbf{2.58} \\
\midrule
\multicolumn{5}{c}{\textbf{Corridor}} \\
\midrule
ME-SR & 5.70 & 13.80 & 7.34 & 38.84 \\
ME-TSVD & 5.36 & 11.99 & 6.22 & 32.76 \\
ME-TReg & \cellcolor{blue!10}4.63 & \cellcolor{blue!10}10.45 & \cellcolor{blue!10}4.22 & \cellcolor{blue!10}22.21 \\
\textbf{Ours} & \cellcolor{blue!30}\textbf{4.21} & \cellcolor{blue!30}\textbf{2.11} & \cellcolor{blue!30}\textbf{3.28} & \cellcolor{blue!30}\textbf{1.67} \\
\bottomrule
\end{tabular}
\vspace{0.5pt}
\begin{flushleft}
\scriptsize{\textit{Note}: \textbf{DR}: degeneracy ratio. Best and second-best results are highlighted in \colorbox{blue!20}{blue} and \colorbox{blue!10}{light blue}. FCN-SR fails on non-degenerate frames "$\textit{--}$".}
\end{flushleft}
\vspace{-2.5em}
\end{table}

\subsection{Degeneracy Analysis}
\subsubsection{Detection Reliability Analysis}
\begin{figure}
    \centering
    \includegraphics[width=0.45\textwidth]{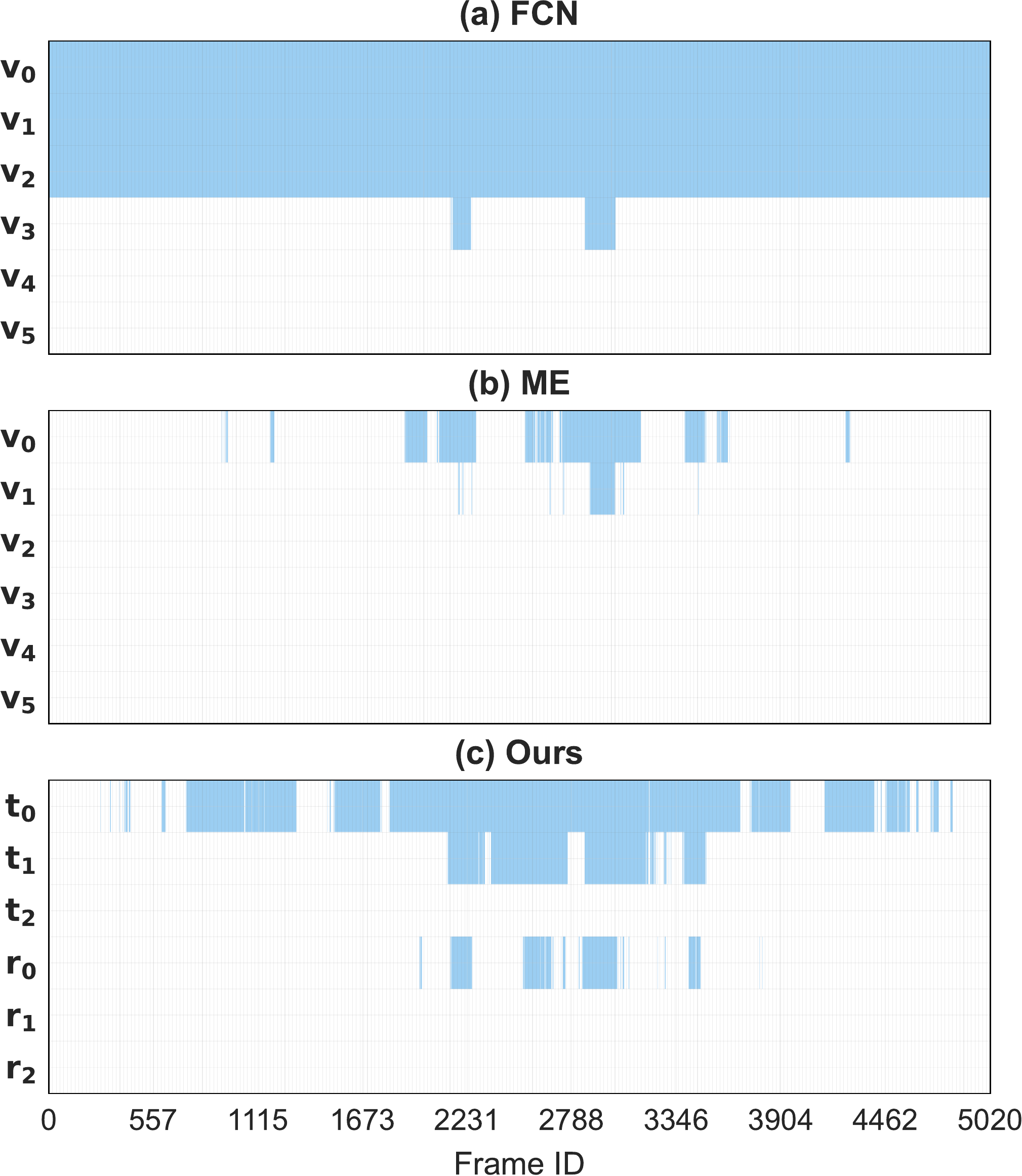}
\caption{Temporal evolution of degeneracy detection across different directions in eigenspace on the parking lot scenario. The light blue indicating detected degeneracy. The vertical axes represent eigenvectors ordered by ascending eigenvalues.
}
\label{fig:iterative_dete_evo}
    \vspace{-2em}
\end{figure}

\begin{figure}
    \centering
    \includegraphics[width=0.45\textwidth]{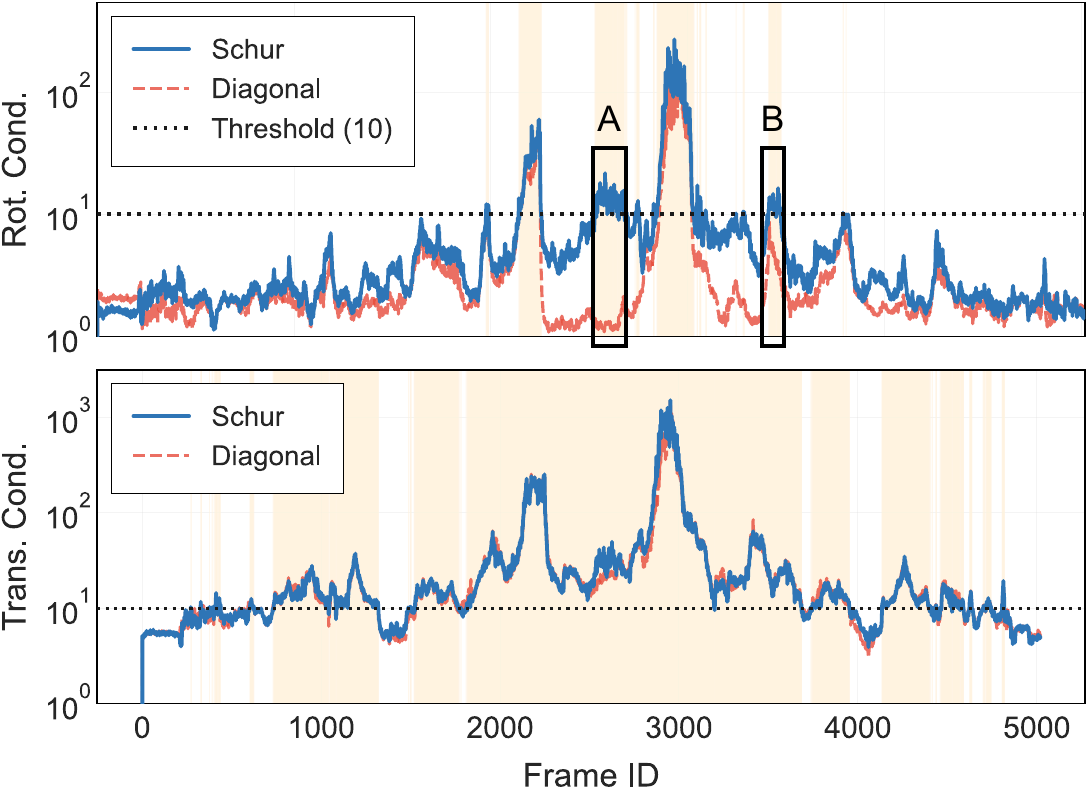}
\caption{Condition number comparison in translational (\textbf{Bottom}) and rotational (\textbf{Top}) subspaces over time on the parking lot dataset. Light yellow regions indicate detected degeneracy areas with detection thresholds set to $10$. Regions A and B demonstrate cases where rotational degeneracy is masked by diagonal methods.}
\label{fig:iterative_cond_evo}
    \vspace{-1em}
\end{figure}

\begin{figure}
    \centering
    \includegraphics[width=0.4\textwidth]{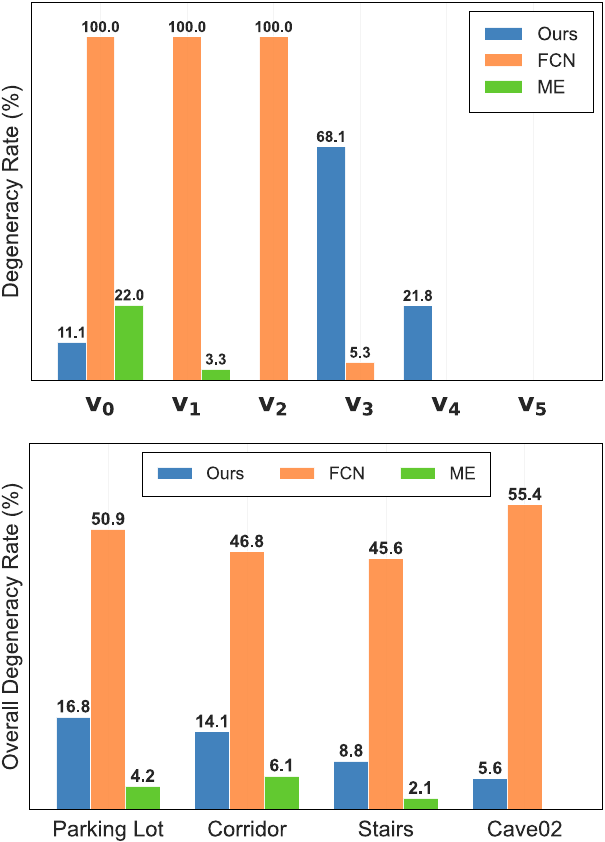}
\caption{\textbf{Top}: degeneracy ratio comparison across different eigen directions ($\bm{v_0}$-$\bm{v_5}$) for various algorithms on the parkinglot dataset. 
% The directions $\bm{v_0}$-$\bm{v_5}$ represent eigenvectors ordered by increasing eigenvalues in the eigenspace. 
 {\textbf{Bottom}}: overall degeneracy ratio comparison across different algorithms on multiple datasets.}
\label{fig:degeneracy_ratio_comparison_multi}
    \vspace{-2em}
\end{figure}

We designed comprehensive experiments to validate the effectiveness of the degeneracy detection module through three complementary analyses: dimensional degeneracy detection across different eigen dimensions over time (\Cref{fig:iterative_dete_evo}), comparative analysis of \textit{Schur complement} versus traditional diagonal condition numbers (\Cref{fig:iterative_cond_evo}), and statistical comparison of specific dimensions and overall degeneracy detection rates across methods (\Cref{fig:degeneracy_ratio_comparison_multi}).

\Cref{fig:iterative_dete_evo} demonstrates the dimensional detection capabilities on the parking lot dataset, revealing fundamental differences in detection strategies. The FCN consistently identifies all three translational dimensions ($\bm{v_0}$-$\bm{v_2}$) as degenerate throughout the process, while ME focuses primarily on the two smallest eigenvalue dimensions ($\bm{v_0}$, $\bm{v_1}$). In contrast, \textit{DCReg} simultaneously detects both translational and rotational degeneracy while correctly identifying the largest translational eigenvalue direction as non-degenerate. This behavior stems from the scale disparity problem inherent in ME and FCN approaches: translational eigenvalues consistently maintain lower magnitudes than rotational counterparts, resulting in persistently large eigenvalue ratios. Consequently, ME exhibit insufficient detection by overlooking larger eigenvalue directions, while FCN demonstrate over-detection by treating all translational dimensions as universally degenerate. Our subspace decoupling strategy effectively addresses this fundamental limitation, achieving more balanced degeneracy detection across dimensional spaces.

\begin{figure*}[!t]
    \centering
        \includegraphics[width=0.9\textwidth]{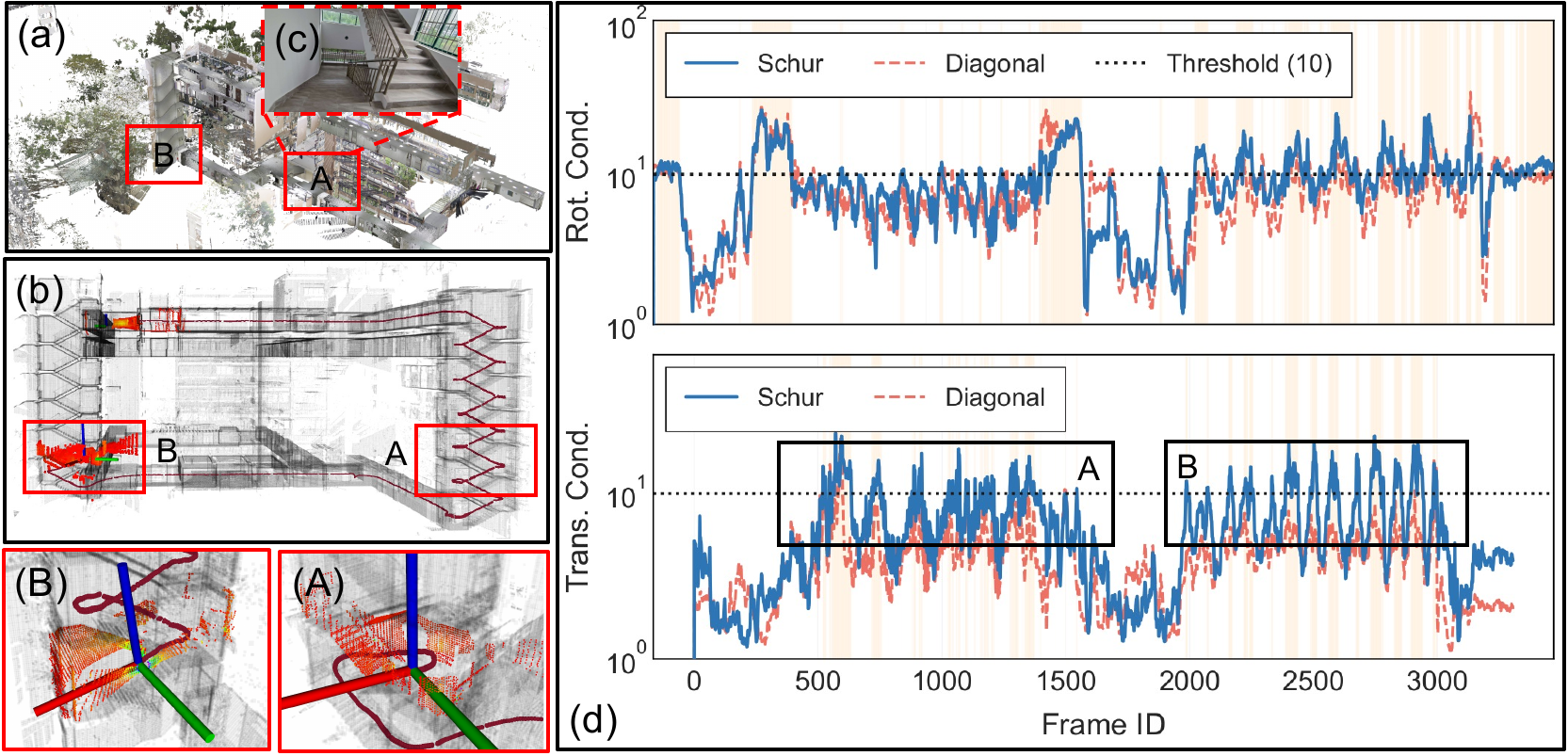}
\caption{The figure compares two different degeneracy case caused by narrow scenarios during (\textbf{A}) downstairs ($Z$ degenerate) and (\textbf{B}) upstairs (Roll-Pitch degenerate), with \textbf{(c)} showing the real-world image. \textbf{(a)}: the ground truth point cloud map. \textbf{(d)}: the evolution of translational and rotational condition numbers computed using our \textit{Schur complement} versus traditional diagonal condition numbers over time. Regions A and B clearly demonstrate condition number variations corresponding to scenarios (A) and (B). 
% This illustrates that degeneracy is not solely caused by repetitive features and textures, but also results from insufficient geometric constraints across different dimensions in the point cloud.
}
\label{fig:stairs_cond}
    \vspace{-1.5em}
\end{figure*}

\cref{fig:iterative_cond_evo} presents the condition number evolution during iterative optimization, revealing critical insights into how coupling effects between translational and rotational components impact degeneracy detection reliability.
For the translational subspace, the Schur condition numbers closely align with their diagonal counterparts throughout the optimization process. This consistency can be attributed to our pose parameterization scheme ($\mathbb{R}^3 \times SO(3)$), where translation parameters maintain relative independence from rotational components in the Hessian structure.
In contrast, the rotational subspace exhibits significant disparities between Schur and diagonal condition numbers, with values that can be either substantially higher or lower than diagonal estimates. This phenomenon provides empirical validation of \cref{thm:condition_bounds}, demonstrating that coupling effects fundamentally alter the conditioning characteristics of rotational components.
Particularly noteworthy are regions A and B, which correspond to the planar degeneracy scenario illustrated in \cref{fig:charictization_pk}. Under identical degeneracy thresholds, the Schur condition number successfully identifies the ill-conditioning, whereas the diagonal condition number incorrectly classifies the configuration as well-constrained. This discrepancy demonstrates that coupling effects can mask degeneracies when using diagonal analysis, while our Schur complement approach reliably exposes these hidden singularities.

Statistical analysis across multiple datasets in \Cref{fig:degeneracy_ratio_comparison_multi} quantifies the detection characteristics of three methods. FCN exhibits systematic over-detection, ME demonstrates insufficient detection, while \textit{DCReg} achieves balanced detection rates that better reflect actual degeneracy conditions. Dimension-specific analysis reveals that all methods correctly detect the largest eigenvalue dimension ($\bm{v_5}$) as non-degenerate, while only \textit{DCReg} successfully detects degeneracy in the $\bm{v_4}$ dimension, further demonstrating the enhanced reliability of our detection method.

\cref{fig:stairs_cond} examines staircase scenarios, with regions A and B corresponding to downward and upward transitions, respectively. Subplot (d) tracks the temporal evolution of condition numbers for both translational and rotational dimensions throughout the traversal.
Staircase environments present unique challenges due to reduced ground plane visibility and constrained passage geometry, leading to insufficient geometric constraints and consequent degeneracy manifestations. These regions frequently exhibit coupled translational-rotational degeneracies rather than isolated single-axis singularities, indicating that geometric structure, not repetitive textures-dominates the degeneracy patterns in such confined spaces.
Specifically, region A predominantly exhibits $Z$-axis degeneracy. Conversely, region B manifests primarily roll-pitch rotational degeneracy during ascent, where the inclined geometry provides insufficient constraints. Notably, diagonal methods fail to detect these critical degeneracies, misclassify them as well-constrained cases. This systematic under-detection further validates the superiority of our \textit{Schur complement} approach in revealing masked degeneracies that compromise registration reliability in structured indoor environments.

\subsubsection{Degeneracy Characterization}

Environmental constraints constitute the primary source of degeneracy in point cloud registration, representing inherent limitations that cannot be modified. However, initial pose serves as a critical secondary factor: when initial estimates approximate ground truth poses, even severely degenerate scenarios can achieve convergence through appropriate mitigation strategies that guide optimization directions during iterative refinement.
This analysis examines degeneracy quantification and motion source across three representative scenarios: open parking lot and corridor passages, providing comprehensive characterization of common degeneracy patterns.

\begin{figure*}
    \centering
    \includegraphics[width=0.8\textwidth]{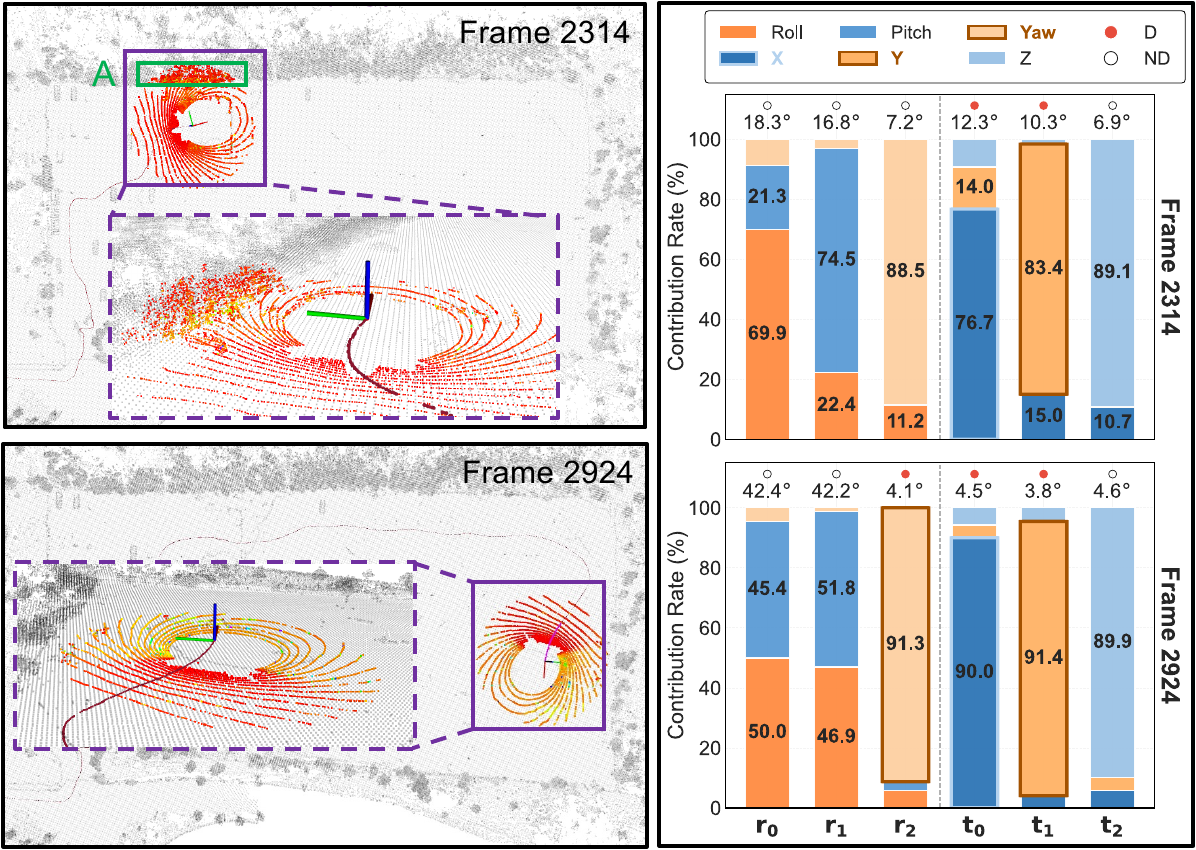}
\caption{Planar degeneracy analysis and characterization in the parkinglot sequence. The figure illustrates degeneracy scenarios comparison between frames $2314$ and $2924$. In frame $2314$, with the curb point clouds (region \textcolor{green}{A}), only $X$ and $Y$ degenerate. Frame $2924$ contains predominantly planar point clouds, resulting in additional yaw degenerate. The right panels show the precise contribution (\%) and strength (angles \si{\degree}) of each dimension in eigenspace ($\bm{r}_0-\bm{r}_2$, $\bm{t}_0-\bm{t}_2$) in terms of physical motion directions ($X$-yaw), enabling analysis of degeneracy motion sources.
}
\label{fig:charictization_pk}
    \vspace{-1.5em}
\end{figure*}

\textbf{Parking Lot Analysis:} \cref{fig:charictization_pk} presents detailed analysis of two characteristic frames from open parking areas, with left panels showing point cloud registration results against black prior map points, and right panels providing corresponding degeneracy decomposition. Frame $2314$ exhibits translational degeneracy in two dimensions ($\bm{t_0}$, $\bm{t_1}$). Through decoupled subspace analysis, we quantify that $\bm{t_0}$-direction degeneracy stems from 77\% $X$-axis motion, 13.6\% $Y$-axis contribution, and 9.4\% $Z$-axis influence, with $\bm{t_0}$ oriented \SI{12.1}{\degree} relative to the $X$-axis. This indicates that $X$-axis motion primarily drives $\bm{t_0}$-dimensional degeneracy. Frame $2924$ demonstrates additional rotational degeneracy ($\bm{r_0}$) alongside translational components ($\bm{t_0}$, $\bm{t_1}$), with $\bm{r_0}$-direction degeneracy primarily  {attributable to yaw motion}. The presence of curb point clouds in green Region A provides effective rotational constraints, explaining the absence of rotational degeneracy in Frame $2314$. These results align with theoretical expectations in \cref{subsec:decoupling_theory}: parking environments predominantly contain ground plane points, making yaw motion constraints inherently challenging.

% \subsubsection{Eigenvector Ambiguity Resolution}

\textbf{Corridor Analysis:} 
\Cref{fig:corridor_ambiguty} analyzes two corridor frames under \SI{5}{m} LiDAR range limitation, with Regions C and D showcasing wall point clouds at corridor termini. Frame analysis reveals $\bm{t_0}$ and $\bm{r_0}$ degeneracies primarily driven by $X$-axis and roll motion components, consistent with theoretical predictions for corridor environments. In Frame $1142$, increased wall points reduces degeneracy to $\bm{r_0}$ only, validating our theoretical framework. Comparative Hessian eigenvalue analysis reveals that the smallest rotational eigenvalue actually falls below the largest translational eigenvalue, indicating that the three smallest eigenvalues in complete Hessian analysis do not necessarily correspond to translational dimensions. This phenomenon demonstrates the practical significance of eigenvalue ordering ambiguity during SVD/EVD and its impact on spectral degeneracy analysis, thereby validating the critical importance of addressing vector ambiguity issues in our \textit{DCReg} framework.

\subsection{ Ablation Studies and Hybrid Analysis}\label{sub_sec:ably_hy}
\subsubsection{Component Contribution Analysis}

The proposed \textit{DCReg} algorithm comprises two core components: degeneracy detection and targeted mitigation module based on a structure-aware preconditioner. 
 {The detection module can operate independently and be combined with alternative mitigation strategies. 
Conversely, the spectral preconditioner in the mitigation module relies on reliable degeneracy analysis, but can be integrated with different solvers (e.g., CG or QR).}

 {To evaluate the contribution of each component, we conduct ablation studies on the corridor dataset with LiDAR range limited to \SI{10}{m}, and compare four configurations summarized in \Cref{tab:ablation_study}. 
Without any degeneracy handling (\textbf{w/o all}), the system suffers from noticeable drift in this highly degenerate corridor, despite its minimal per-frame runtime. 
Simply enabling detection without any degeneracy mitigation (\textbf{w/o PCG}) further degrades performance, as the detected degeneracies are not acted upon and the solver continues to update along ill-conditioned directions. 
Changing the linear solver alone has only a modest effect: using CG without preconditioning (\textbf{w/o Precond.}) improves ATE slightly over the QR solver, but still leaves substantial drift. 
In contrast, enabling the proposed spectral preconditioner while solving with QR (\textbf{w/o CG}) already yields a large accuracy gain  with only a small runtime increase, indicating that the structure-aware preconditioner is the dominant factor for improving robustness under degeneracy. 
Finally, \textbf{DCReg (Full)} combines the preconditioner with CG and achieves the best accuracy–efficiency trade-off: it further reduces ATE to \SI{7.44}{cm} while keeping the runtime comparable to the baseline.
}

\begin{figure*}
    \centering
        \includegraphics[width=0.95\textwidth]{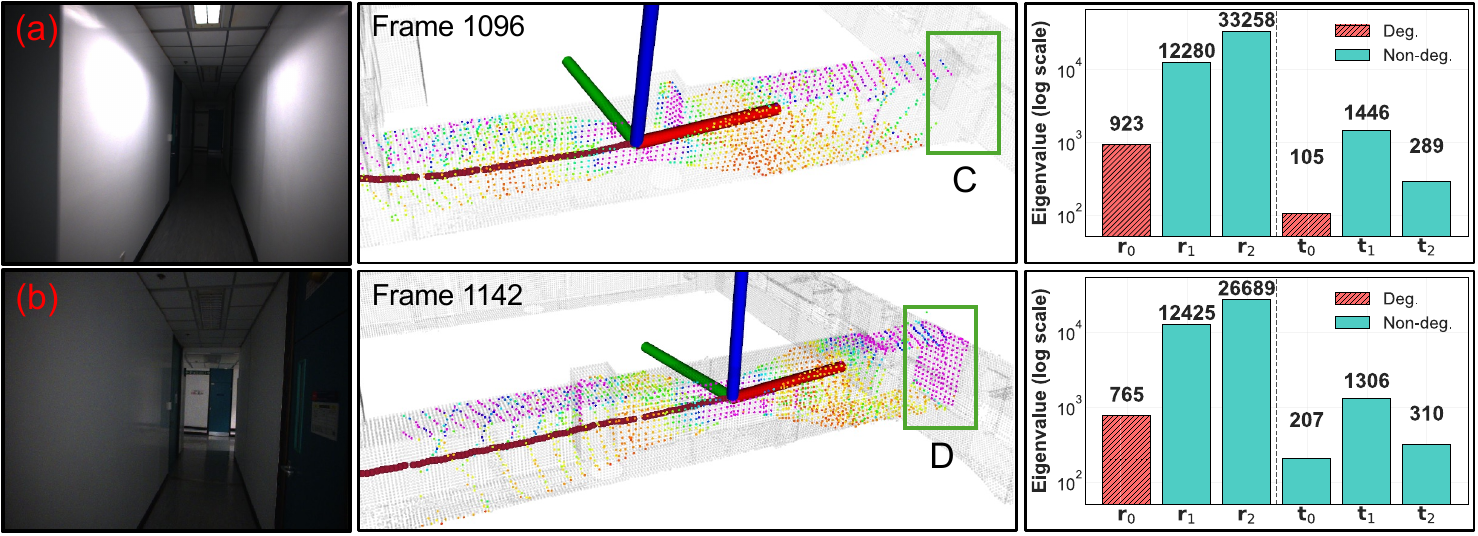}
\caption{Eigenvector ordering ambiguity in corridor scenarios. \textbf{Left}: Real-world images at frames $1096$ and $1142$. \textbf{Middle}: Localization (\textcolor{red}{red} trajectory) and mapping results. \textbf{Right}: Eigenvalues at first ICP iteration. Frame $1096$ shows $X$ and roll degeneracy, while frame $1142$ shows only roll degeneracy due to wall points in \textbf{region D}. Rotational eigenvalues are not consistently larger than translational ones, demonstrating ordering ambiguity that impacts degeneracy detection.}
\label{fig:corridor_ambiguty}
    \vspace{-1.5em}
\end{figure*}

% how to explain the results wo pcg
% 是否需要保留 Times
% \begin{table}[!t]
% \centering
% \caption{Ablation Study and Hybrid Analysis}
% \label{tab:ablation_study}
% \footnotesize
% \setlength{\tabcolsep}{8pt}
% \renewcommand{\arraystretch}{1.2}
% \begin{tabular}{@{}l||cc||c@{}}
% \toprule
% \multicolumn{4}{c}{\textbf{Corridor (DR: 58.04\%)}} \\
% \midrule
% \textbf{Configuration} & \textbf{ATE $\downarrow$} & \textbf{AC $\downarrow$} & \textbf{Time $\downarrow$} \\
% & \scriptsize{(cm)} & \scriptsize{(cm)} & \scriptsize{(ms)} \\
% \midrule
% \multicolumn{4}{c}{\textit{(a) Component Ablation}} \\

% \midrule
%  {w/o P} &  {23.95} &  {3.57} &  {1.82} \\
%  {w/o CG} & \cellcolor{blue!10}9.22 & \cellcolor{blue!10}{3.49} &  {1.27} \\
% w/o PCG & 164.14 & 4.42 & 2.03 \\
% w/o Both & 28.54 & 3.67 & \cellcolor{blue!30}\textbf{1.21} \\

% \midrule
% \multicolumn{4}{c}{\textit{(b) Hybrid Approaches (Our Detection + Traditional Mitigation)}} \\
% \midrule
% DCReg + SR & 37.75 & 4.61 & 1.57 \\
% DCReg + TSVD & 17.81 & 3.60 & 1.40 \\
% DCReg + TReg & 22.79 & 3.51 & 1.89 \\
% \midrule
% \textbf{DCReg (Full)} & \cellcolor{blue!30}\textbf{7.44} & \cellcolor{blue!30}\textbf{3.45} & \cellcolor{blue!10}\textbf{1.24} \\
% \bottomrule
% \end{tabular}
% \vspace{1pt}
% \begin{flushleft}
% \scriptsize{\textit{Note}: "w/o Both" represents \textit{DCReg} without degeneration awareness. "w/o PCG" means no mitigation. Best and second-best results are highlighted in \colorbox{blue!30}{blue} and \colorbox{blue!10}{light blue}.}
% \end{flushleft}
%     \vspace{-3em}
% \end{table}

\begin{table}[!t]
\centering
\caption{Ablation Study and Hybrid Analysis}
\label{tab:ablation_study}
\footnotesize
\setlength{\tabcolsep}{8pt}
\renewcommand{\arraystretch}{1.2}
\begin{tabular}{@{}l||cc||c@{}}
\toprule
\multicolumn{4}{c}{\textbf{Corridor (DR: 58.04\%)}} \\
\midrule
\textbf{Configuration} & \textbf{ATE $\downarrow$} & \textbf{AC $\downarrow$} & \textbf{Time $\downarrow$} \\
& \scriptsize{(cm)} & \scriptsize{(cm)} & \scriptsize{(ms)} \\
\midrule
\multicolumn{4}{c}{\textit{Component Ablation}} \\
\midrule
% Base (QR)    & 28.54  & 3.67 & \cellcolor{blue!30}\textbf{1.21} \\
% Det. only             & 164.14 & 4.42 & 2.03 \\
%  Det. + CG ($\bm{P}{=}\bm{I}$)      & 23.95  & 3.57 & 1.82 \\
%  Det. + Precond. + QR       & \cellcolor{blue!10}9.22 & \cellcolor{blue!10}3.49 & 1.27 \\
w/o all    & 28.54  & 3.67 & \cellcolor{blue!30}\textbf{1.21} \\
w/o PCG            & 164.14 & 4.42 & 2.03 \\
% w/o all    & 9.87  & 4.47 & 2.33 \\
% w/o PCG            & 10.26 & 3.57 & 1.30 \\
w/o Precond.     & 23.95  & 3.57 & 1.82 \\
w/o CG       & \cellcolor{blue!10}9.22 & \cellcolor{blue!10}3.49 & 1.27 \\
\midrule
\multicolumn{4}{c}{\textit{Hybrid (Our Detection + Classical Mitigation)}} \\
\midrule
DCReg + SR   & 37.75 & 4.61 & 1.57 \\
DCReg + TSVD & 17.81 & 3.60 & 1.40 \\
DCReg + TReg & 22.79 & 3.51 & 1.89 \\
\midrule
\textbf{DCReg (Full)} & \cellcolor{blue!30}\textbf{7.44} & \cellcolor{blue!30}\textbf{3.45} & \cellcolor{blue!10}1.24 \\
\bottomrule
\end{tabular}
\vspace{1pt}
\begin{flushleft}
\scriptsize{\textit{Note}: \textbf{w/o all}: QR solver without degeneracy awareness. \textbf{w/o PCG}: Schur-based detection without mitigation. \textbf{w/o Precond.}: CG solver without preconditioning. \textbf{w/o CG}: spectral preconditioning solved via QR. Hybrid approaches use QR solver with different mitigation method. Best/second-best: \colorbox{blue!30}{blue}/\colorbox{blue!10}{light blue}.}
\end{flushleft}
\vspace{-3em}
\end{table}

\subsubsection{Hybrid Method Evaluation} \label{sub_sec:hybrid_eva}

For hybrid method evaluations, coupling our detection with SR-based mitigation yielded counterintuitive negative optimization for both trajectory and mapping accuracy. This degradation stems from SR's approach of discarding updates along degenerate dimensions, potentially compromising initial pose estimates for subsequent iterations and ultimately reducing registration precision. However, our detection method combined with SR significantly outperformed the baseline ME-SR approach (\cref{tab:trajectory_accuracy_comparison}), validating the effectiveness of our balanced detection strategy for both translational and rotational dimensional degeneracies.
Similarly, both Ours+TSVD and DCReg+TReg consistently outperformed their respective ME-TSVD and ME-TReg counterparts, further confirming the critical importance of our degeneracy detection module.
Notably, DCReg+TSVD achieved superior results compared to DCReg+TReg and Ours+SR. This performance advantage likely stems from the spectral analysis characteristics in corridor scenarios, where eigenvector ordering ambiguity (\Cref{fig:corridor_ambiguty}) causes eigenvalues of degenerate rotational dimensions to approximate those of multiple translational dimensions. Under such conditions, directly discarding degenerate dimensional information may paradoxically yield better rank approximation effects. Conversely, TReg requires frequent parameter adjustments based on eigenvalue magnitudes to achieve optimal performance, which proves impractical in real-world applications.
The complete \textit{DCReg} algorithm demonstrates significant improvements in both trajectory and mapping accuracy while maintaining computational efficiency comparable to baseline methods. These results underscore that precise degeneracy detection combined with targeted mitigation substantially enhances overall registration accuracy in degenerate scenarios.

\begin{figure}
    \centering
    \includegraphics[width=0.45\textwidth]{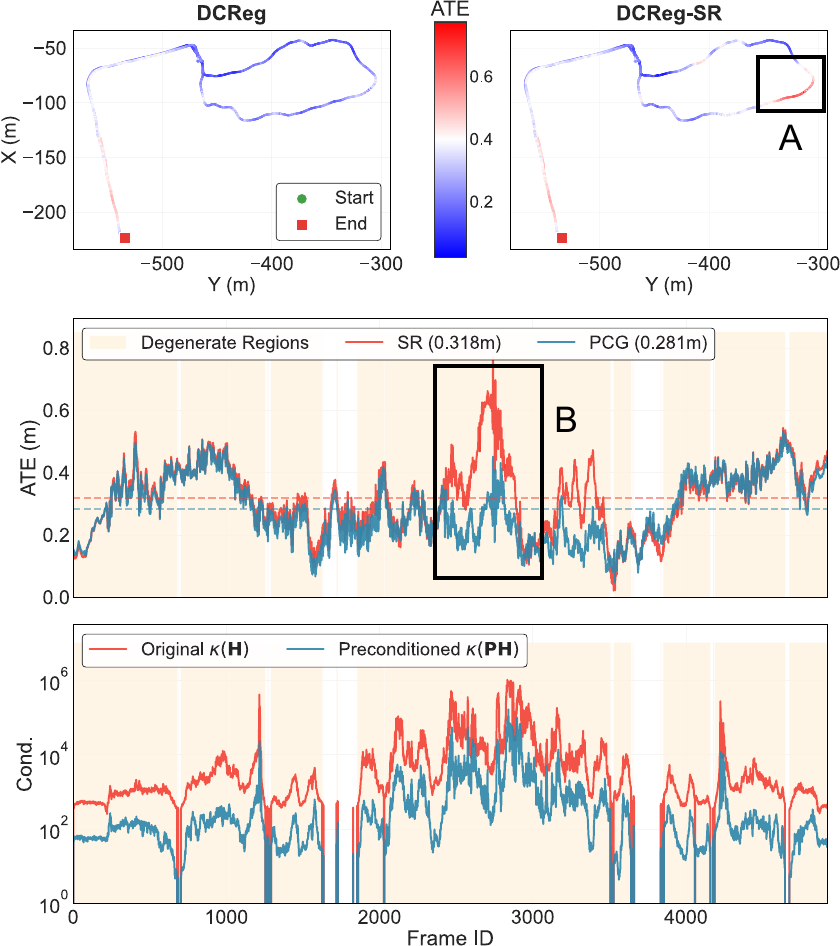}
\caption{Trajectory error and ATE comparison over time between DCReg and DCReg-SR on the parkinglot dataset. DCReg-SR exhibits significantly increased errors (Resion A and B) during trajectory segments where $XY$ and Yaw degenerate simultaneously, demonstrating the advantages of our targeted PCG method over SR.} 
\label{fig:dcReg_comprehensive_analysis_marked}
    \vspace{-2em}
\end{figure}

\Cref{fig:dcReg_comprehensive_analysis_marked} illustrates the ATE accuracy evolution over time for both \textit{DCReg} and \textit{DCReg}-SR algorithms on the parkinglot dataset. A pronounced accuracy degradation occurs in Region A, where simultaneous degeneracies in XY translation and yaw rotation manifest, resulting in substantial ATE increases. This phenomenon demonstrates the superiority of our targeted mitigation approach over conventional SR methods, as \textit{DCReg} specifically addresses dimensional degeneracies rather than applying uniform corrections across all dimensions.
The condition number analysis presented in subfigure (c) reveals two critical insights into the algorithmic behavior. First, during periods of severe degeneracy, the original problem exhibits dramatic condition number spikes, indicating numerical instability that directly correlates with the observed ATE degradation in Region A. Second, the preconditioned problem consistently maintains lower condition numbers compared to the original formulation throughout the entire trajectory, providing quantitative evidence that our PCG algorithm effectively mitigates degeneracy through systematic condition number control. This numerical conditioning improvement transform directly to the enhanced trajectory accuracy observed in the experimental results, validating the theoretical foundation of our targeted degeneracy mitigation strategy.

\subsubsection{Parameter Sensitivity}

This section examines the sensitivity of \textit{DCReg} to parameter variations and demonstrates the robustness of our approach compared to existing methods. The parameter analysis encompasses both the degeneracy detection and targeted mitigation modules.

For the detection module, \textit{DCReg} requires only a single threshold parameter: the degeneracy condition number ratio. This parameter design offers significant advantages over alternative approaches. Unlike ME methods that require environment-specific parameter tuning, or FCN approaches that necessitate frequent adjustments due to scale discrepancies between translational and rotational eigenvalues, our subspace-based detection threshold possesses clear physical interpretation. The threshold is determined solely by eigenvalue ratios across different dimensions, eliminating the need for case-by-case parameter optimization.
The targeted PCG module involves three parameters: the target subspace condition number $\kappa$, PCG iteration tolerance, and maximum iteration count. The latter two parameters demonstrate remarkable stability across applications. Setting the tolerance to $1 \times 10^{-6}$ consistently achieves convergence within \rt{$10$} iterations, rendering these parameters practically invariant across different scenarios.

% \Cref{fig:dcreg_parameter_analysis} presents a comprehensive analysis of target condition number $\kappa_{tg}$ variations on the parkinglot dataset, examining trajectory accuracy, mapping precision, and average PCG iteration counts. When $\kappa$ ranges from $1$ to $10$, both ATE and mapping accuracy remain remarkably stable while average PCG iterations increase gradually. This behavior validates our theoretical analysis in \cref{subsec:complementary_perspectives}, confirming that smaller condition numbers accelerate convergence and that subspace condition number control effectively influences overall optimization performance.
% \rt{Theoretical analysis in \cref{subsec:complementary_perspectives} establishes that the target condition number should satisfy $1 < \kappa < 10$.}
% Values exceeding this range can induce information loss in mildly degenerate scenarios. This prediction is also empirically confirmed in \Cref{fig:dcreg_parameter_analysis}, where configurations with $\kappa \in \{20, 50, 100\}$ exhibit significant degradation in both trajectory and mapping accuracy, validating our theoretical framework.
% Based on these findings, we recommend setting the $\kappa_{tg}$ equal to the threshold $\kappa_{th}$ in practical implementations, ensuring consistency between detection and mitigation strategies while maintaining optimal performance across diverse scenarios.

\begin{figure}[!t]
    \centering
    \includegraphics[width=0.48\textwidth]{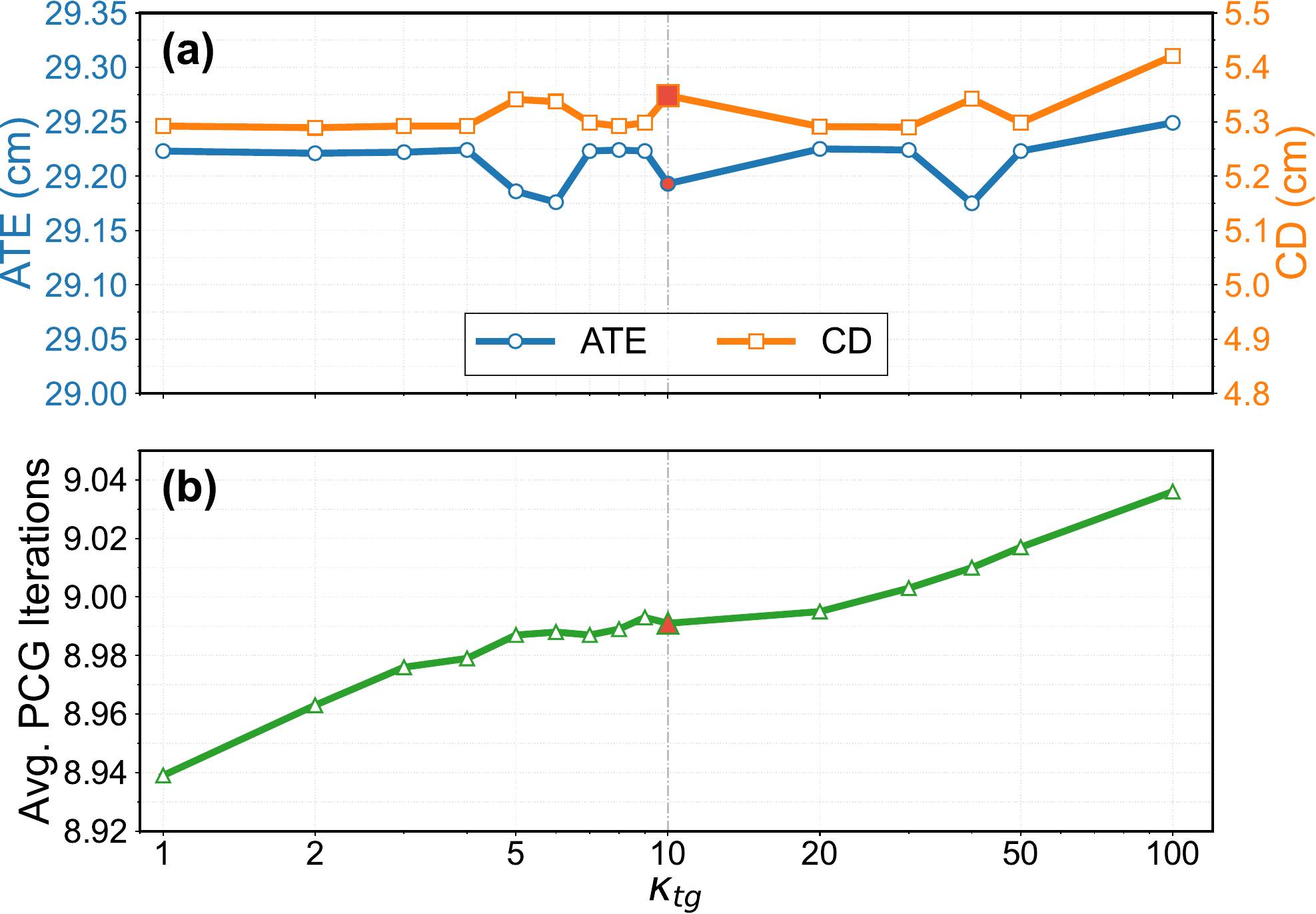}
\caption{Parameter analysis on the parking lot dataset. (a) ATE and CD. (c) Average PCG iterations of each frame. {While PCG iterations increase monotonically with $\kappa$, both ATE and CD remain stable, demonstrating DCReg's robustness to parameter tuning.}} 
\label{fig:dcreg_parameter_analysis}
    \vspace{-2em}
\end{figure}

{
\Cref{fig:dcreg_parameter_analysis} presents a sensitivity analysis of the target condition number $\kappa_{\mathrm{tg}}$ on the parking-lot dataset, examining trajectory accuracy, mapping precision, and average PCG iteration counts. 
When $\kappa_{\mathrm{tg}}$ ranges from $1$ to $10$, both ATE and CD remain almost stable, while the average number of PCG iterations increases gradually. 
This behavior is consistent with our analysis in \cref{subsec:complementary_perspectives}: a smaller target condition number enforces a tighter upper bound on the subspace condition number, yielding a better-conditioned preconditioned system and thus faster convergence, without affecting the final solution.
For larger values $\kappa_{\mathrm{tg}} \in \{20,50,100\}$, the spectral shaping becomes weaker and the number of PCG iterations continues to grow. 
Nevertheless, both ATE and CD vary only mildly even in this extreme regime, indicating that \textit{DCReg} is robust to mis-tuning of $\kappa_{\mathrm{tg}}$. 
This robustness illustrates a key advantage of our preconditioner design over explicit regularization: mis-specified $\kappa_{\mathrm{tg}}$ primarily impacts computational cost (through the condition number of $\bm{P}\bm{H}$) rather than biasing the estimator, since the original normal equations remain unchanged.
In practice, we set $\kappa_{\mathrm{tg}}$ equal to the detection threshold $\kappa_{\mathrm{th}}$ ( $\kappa_{\mathrm{th}}{=}10$), which lies in the flat region of the curve and provides a good trade-off between robustness and runtime across datasets.
}

\section{{Discussions and Limitations}}\label{sec:discussion}

This section analyzes key insights and limitations encountered in addressing degenerate registration problems, providing guidance for practical applications.

While our framework provides comprehensive analysis and solution for degenerate registration, the fundamental cause of degeneracy remains rooted in environmental geometry.  {In theoretically absolute degenerate scenarios, such as perfect planar surfaces or infinite corridors, certain dimensional information approaches zero, rendering any degeneracy mitigation method ineffective. However, such idealized conditions rarely exist in real-world environments.
As in a near-planar surfaces, even when eigenspace dimensions exhibit insufficient information, the values remain non-zero given reasonable initial poses. Under these conditions, degeneracy mitigation methods can meaningfully improve numerical stability and convergence characteristics. This observation underscores the practical value of our approach in realistic scenarios where environmental geometry introduces conditioning challenges without complete information loss.}
The convergence basin of point cloud registration algorithms imposes a critical constraint on degeneracy mitigation effectiveness. When initial pose estimates fall outside this basin, optimization becomes trapped in local minima, rendering degeneracy analysis ineffective regardless of the mitigation strategy employed. Conversely, given reasonable initial poses within the convergence domain, degeneracy mitigation can strategically adjust pose estimation during optimization iterations, guiding the solution toward correct global optima and thereby enhancing both robustness and efficiency of registration algorithms.

This analysis reveals that reliable degenerate registration depends on the interplay between three factors: environmental geometry, initial pose, and mitigation strategy. Our framework demonstrates particular value in the intermediate regime where partial information exists and initial estimates provide reasonable starting points, conditions that characterize the majority of practical robotic applications. Understanding these boundaries helps practitioners assess when degeneracy mitigation will provide meaningful benefits and when alternative strategies may be necessary.

\section{Future Directions}\label{sec:future}

{
Beyond the specific design of \textit{DCReg}, we believe this work suggests a broader direction for the community.} 

{First, degeneracy analysis should not stop at identifying weak directions after aggregating all measurements. A natural next step is to move toward measurement-level or factor-level information analysis~\cite{tuna2024informed}, namely, separating the pure constraining power of different measurements from varying noise. Such a perspective could support selective or active sensing and extend degeneracy-aware analysis to a wider class of estimation problems, including calibration, radar or visual SLAM.}
{Second, the pose decoupling principle developed here is not limited to LiDAR degeneracy~\cite{xu2023ring}. Extending it to heterogeneous parameter groups in multimodal fusion may help reveal the observability~\cite{xu2026rotation} of each parameter block and avoid spurious apparent constraints caused by error compensation.} 
{Third, while our experiments already evaluate \textit{DCReg} within a system-level localization pipeline, an important open direction is to better understand how local geometric degeneracy interacts with temporal priors in the kalman-filter~\cite{chen2024relead} and factor-graph~\cite{hu2024paloc} optimization systems. Strong priors from IMU or other sensors can also stabilize weakly constrained local estimation, but they can also redistribute or mask degeneracy patterns.} 
{Finally, dynamic or sparse environments lead to rapidly changing measurement availability and dimensionally imbalanced constraints. Extending the present framework to such time-varying observation regimes is therefore another promising direction for future research.}

\section{Conclusion}
\label{sec:conclusion}

We present the decoupled and quantitatively characterized approach for degenerate point cloud registration with exceptional computational efficiency. Our method leverages insights from linear algebra (SVD, Schur complement, Schmidt orthogonalization), geometry (subspace analysis), estimation theory, and optimization (PCG) to design two fundamental modules: subspace decoupling and targeted-PCG solver.
The subspace decoupling module employs \textit{Schur complement} to eliminate rotation-translation coupling effect while quantifying relationships between eigenspace directions and physical motions. This enables reliable degeneracy detection by explaining motion induced degeneracy sources in specific environments. The targeted-PCG solver provides directional condition number enhancement for identified degenerate directions, improving convergence and optimization stability. We provide comprehensive theoretical analysis for both modules, representing the rigorous treatment of their kind for degenerate registration problems.

Extensive validation across simulated and real-world datasets demonstrates four key advantages: (1) significant improvements in trajectory and mapping accuracy during long-duration localization tasks, (2) enhanced convergence reliability with  {typical $5-30\times$ (up to $116\times$)} speedup in average localization performance, (3) quantitative characterization of degeneracy motions across diverse scenarios, and (4) seamless integration capability with existing degeneracy mitigation methods while achieving accuracy improvements. Our efficient open-source C++ implementation facilitates broader adoption.

Beyond core contributions, \texttt{DCReg} provides a fundamental analytical framework revealing that pose estimation depends entirely on correspondence pairs that evolve with initial poses during iteration. Consequently, degeneracy stems not only from repetitive geometric features (e.g., tunnels) but also from sparse environments and poor initial poses causing insufficient constraints. This insight opens numerous research opportunities for advancing LiDAR-based safe and reliable perception systems.
% Future work includes extending degeneracy analysis and mitigation to long-duration localization scenarios and investigating applications in multi-modal sensor fusion and dynamic environment handling. The theoretical framework established here provides a foundation for addressing ill-conditioned estimation problems across diverse robotic perception tasks.
{
    More broadly, we view \texttt{DCReg} as a step toward a general framework for degeneracy-aware state estimation: from pose-level weak-direction diagnosis to factor-level information analysis, from local geometric analysis to prior-coupled estimation systems, and from pose decoupling to heterogeneous-parameter sensor fusion.
}

\begin{acks}
 {
The authors would like to thank Dr. Binqian Jiang and Dr. Jianhao Jiao for insightful discussions that enriched the theoretical framework of this work. We are grateful to Mr. Turcan Tuna for his technical assistance with the baseline implementation. We also thank Dr. Sile Li and Miss Tianyu Ma for their unwavering support during challenging phases of this research.
}
\end{acks}

\bibliographystyle{SageH}
\bibliography{refs.bib}

\appendix
\section{Proofs for Schur-Complement Conditioning}
\label{appendix:schur_conditioning}

% This appendix provides rigorous proofs for Theorem~\ref{thm:effective_curvature}, Proposition~\ref{prop:projection}, and Proposition~\ref{prop:scale_invariance}.

\subsection{Proof of Theorem 1}\label{appendix:effective_curvature}

Consider the quadratic objective from the linearized system:
\begin{equation}
Q(\bm{\xi}) = \tfrac{1}{2}\bm{\xi}^\top\bm{H}\bm{\xi} + \bm{g}^\top\bm{\xi}
\end{equation}
where $\bm{\xi} = [\bm{\phi}^\top, \delta\bm{t}^\top]^\top \in \mathbb{R}^6$, $\bm{g} = [\bm{g}_R^\top, \bm{g}_t^\top]^\top = \bm{J}^\top\bm{r}_0$.
Expanding in block form:
% \begin{equation}
% Q(\bm{\phi}, \delta\bm{t}) = \tfrac{1}{2}
% \begin{bmatrix}
% \bm{\phi} \\ \delta\bm{t}
% \end{bmatrix}^\top
% \begin{bmatrix}
% \bm{H}_{RR} & \bm{H}_{Rt} \\
% \bm{H}_{tR} & \bm{H}_{tt}
% \end{bmatrix}
% \begin{bmatrix}
% \bm{\phi} \\ \delta\bm{t}
% \end{bmatrix}
% + 
% \begin{bmatrix}
% \bm{g}_R \\ \bm{g}_t
% \end{bmatrix}^\top
% \begin{bmatrix}
% \bm{\phi} \\ \delta\bm{t}
% \end{bmatrix}
% \end{equation}
\begin{equation}
\begin{aligned}
Q(\bm{\phi},\delta\bm{t})
&= \tfrac{1}{2}
\begin{bmatrix}\bm{\phi}\\ \delta\bm{t}\end{bmatrix}^{\!\top}
\begin{bmatrix}
\bm{H}_{RR} & \bm{H}_{Rt}\\[1pt]
\bm{H}_{tR} & \bm{H}_{tt}
\end{bmatrix}
\begin{bmatrix}\bm{\phi}\\ \delta\bm{t}\end{bmatrix} \\
&\quad
+ \begin{bmatrix}\bm{g}_R\\ \bm{g}_t\end{bmatrix}^{\!\top}
  \begin{bmatrix}\bm{\phi}\\ \delta\bm{t}\end{bmatrix}.
\end{aligned}
\end{equation}

\textbf{Step 1: Partial minimization.} Setting $\nabla_{\delta\bm{t}} Q = \bm{0}$:
\begin{equation}
\bm{H}_{tR}\bm{\phi} + \bm{H}_{tt}\delta\bm{t} + \bm{g}_t = \bm{0}
\end{equation}

Solving for optimal $\delta\bm{t}$ given $\bm{\phi}$:
\begin{equation}
\delta\bm{t}^*(\bm{\phi}) = -\bm{H}_{tt}^{-1}(\bm{H}_{tR}\bm{\phi} + \bm{g}_t)
\end{equation}

\textbf{Step 2: Reduced objective.} 
Substituting $\delta\bm{t}^*(\bm{\phi})$ into $Q$:
% \begin{align}
% \footnotesize
% Q_{\text{red}}(\bm{\phi}) &= Q(\bm{\phi}, \delta\bm{t}^*(\bm{\phi})) \notag \\
% &= \tfrac{1}{2}\bm{\phi}^\top\bm{H}_{RR}\bm{\phi} + \bm{\phi}^\top\bm{H}_{Rt}\delta\bm{t}^*(\bm{\phi}) + \tfrac{1}{2}\delta\bm{t}^{*\top}(\bm{\phi})\bm{H}_{tt}\delta\bm{t}^*(\bm{\phi}) \notag \\
% &\quad + \bm{g}_R^\top\bm{\phi} + \bm{g}_t^\top\delta\bm{t}^*(\bm{\phi})
% \end{align}
\begin{equation}
% \footnotesize
\begin{aligned}
Q_{\mathrm{red}}(\bm{\phi})
&= Q\bigl(\bm{\phi}, \dtstar\bigr) \\
&= \tfrac{1}{2}\bm{\phi}^{\top}\bm{H}_{RR}\bm{\phi}
  + \bm{\phi}^{\top}\bm{H}_{Rt}\dtstar \\
&\quad
  + \tfrac{1}{2}\dtstar^{\top}\bm{H}_{tt}\dtstar
  + \bm{g}_{R}^{\top}\bm{\phi}
  + \bm{g}_{t}^{\top}\dtstar .
\end{aligned}
\end{equation}

After algebraic manipulation:
% \begin{equation}
% \footnotesize
% Q_{\text{red}}(\bm{\phi}) = \tfrac{1}{2}\bm{\phi}^\top\underbrace{(\bm{H}_{RR} - \bm{H}_{Rt}\bm{H}_{tt}^{-1}\bm{H}_{tR})}_{\bm{S}_R}\bm{\phi} + \underbrace{(\bm{g}_R - \bm{H}_{Rt}\bm{H}_{tt}^{-1}\bm{g}_t)^\top}_{\tilde{\bm{g}}_R^\top}\bm{\phi} + \text{const}
% \end{equation}
\begin{equation}
\begin{aligned}
Q_{\mathrm{red}}(\bm{\phi})
&= \tfrac{1}{2}\bm{\phi}^{\top}
   \underbrace{\bigl(\bm{H}_{RR}-\bm{H}_{Rt}\bm{H}_{tt}^{-1}\bm{H}_{tR}\bigr)}_{\bm{S}_{R}}
   \bm{\phi} \\
&\quad + 
   \underbrace{\bigl(\bm{g}_{R}-\bm{H}_{Rt}\bm{H}_{tt}^{-1}\bm{g}_{t}\bigr)^{\top}}_{\tilde{\bm{g}}_{R}^{\top}}
   \bm{\phi}
   + \text{const.}
\end{aligned}
\end{equation}

This establishes part (i).

\textbf{Step 3: Sensitivity analysis.} 
The reduced normal equation is:
\begin{equation}
\bm{S}_R\bm{\phi}^* = -\tilde{\bm{g}}_R
\end{equation}

Under perturbation $\tilde{\bm{g}}_R \mapsto \tilde{\bm{g}}_R + \Delta\tilde{\bm{g}}_R$:
\begin{equation}
\bm{S}_R(\bm{\phi}^* + \Delta\bm{\phi}^*) = -(\tilde{\bm{g}}_R + \Delta\tilde{\bm{g}}_R)
\end{equation}

Therefore $\bm{S}_R\Delta\bm{\phi}^* = -\Delta\tilde{\bm{g}}_R$, yielding:
% \begin{equation}
% \footnotesize
% \|\Delta\bm{\phi}^*\|_2 = \|\bm{S}_R^{-1}\Delta\tilde{\bm{g}}_R\|_2 \leq \|\bm{S}_R^{-1}\|_2 \|\Delta\tilde{\bm{g}}_R\|_2 = \frac{1}{\lambda_{\min}(\bm{S}_R)} \|\Delta\tilde{\bm{g}}_R\|_2
% \end{equation}
\begin{equation}
\begin{aligned}
\|\Delta\bm{\phi}^{\star}\|_2
&= \|\bm{S}_R^{-1}\,\Delta\tilde{\bm{g}}_R\|_2
\ \leq\ \|\bm{S}_R^{-1}\|_2\,\|\Delta\tilde{\bm{g}}_R\|_2 \\
&= \lambda_{\min}^{-1}(\bm{S}_R)\,\|\Delta\tilde{\bm{g}}_R\|_2 .
\end{aligned}
\end{equation}

Similarly, 
$\|\bm{\phi}^*\|_2 = \|\bm{S}_R^{-1}\tilde{\bm{g}}_R\|_2 \geq \frac{\|\tilde{\bm{g}}_R\|_2}{\|\bm{S}_R\|_2} = \frac{\|\tilde{\bm{g}}_R\|_2}{\lambda_{\max}(\bm{S}_R)}$.
Combining:
% \begin{equation}
% \footnotesize
% \frac{\|\Delta\bm{\phi}^*\|_2}{\|\bm{\phi}^*\|_2} \leq \frac{\lambda_{\max}(\bm{S}_R)}{\lambda_{\min}(\bm{S}_R)} \cdot \frac{\|\Delta\tilde{\bm{g}}_R\|_2}{\|\tilde{\bm{g}}_R\|_2} = \kappa(\bm{S}_R) \cdot \frac{\|\Delta\tilde{\bm{g}}_R\|_2}{\|\tilde{\bm{g}}_R\|_2}
% \end{equation}
\begin{equation}
% \footnotesize
\begin{aligned}
\frac{\|\Delta\bm{\phi}^{\star}\|_2}{\|\bm{\phi}^{\star}\|_2}
&\le
\frac{\lambda_{\max}(\bm{S}_R)}{\lambda_{\min}(\bm{S}_R)}
\cdot
\frac{\|\Delta\tilde{\bm{g}}_R\|_2}{\|\tilde{\bm{g}}_R\|_2} \\
&=
\kappa(\bm{S}_R)\,
\frac{\|\Delta\tilde{\bm{g}}_R\|_2}{\|\tilde{\bm{g}}_R\|_2}\, .
\end{aligned}
\end{equation}
This proves part (ii). The proof for $\bm{S}_t$ follows by symmetry.\qed

\subsection{Proof of Proposition~\ref{prop:projection}}\label{appendix:projection}

\paragraph{Unweighted Case.}
Consider the Jacobian decomposition $\bm{J} = [\bm{J}_R \mid \bm{J}_t] \in \mathbb{R}^{m \times 6}$, where $\bm{J}_R \in \mathbb{R}^{m \times 3}$ and $\bm{J}_t \in \mathbb{R}^{m \times 3}$ represent the rotational and translational components, respectively. The associated Hessian matrix admits the block structure:
\begin{equation}
\bm{H} = \bm{J}^\top\bm{J} = 
\begin{bmatrix}
\bm{J}_R^\top\bm{J}_R & \bm{J}_R^\top\bm{J}_t \\
\bm{J}_t^\top\bm{J}_R & \bm{J}_t^\top\bm{J}_t
\end{bmatrix} = 
\begin{bmatrix}
\bm{H}_{RR} & \bm{H}_{Rt} \\
\bm{H}_{tR} & \bm{H}_{tt}
\end{bmatrix},
\end{equation}
where $\bm{H}_{RR} = \bm{J}_R^\top\bm{J}_R \in \mathbb{R}^{3 \times 3}$, $\bm{H}_{tt} = \bm{J}_t^\top\bm{J}_t \in \mathbb{R}^{3 \times 3}$, and $\bm{H}_{Rt} = \bm{H}_{tR}^\top = \bm{J}_R^\top\bm{J}_t \in \mathbb{R}^{3 \times 3}$.

The orthogonal projector onto $\mathrm{range}(\bm{J}_t)$ using the Moore-Penrose pseudoinverse is:
\begin{equation}
\bm{P}_t = \bm{J}_t\bm{J}_t^{+} \in \mathbb{R}^{m \times m},
\end{equation}
where $\bm{J}_t^{+} = (\bm{J}_t^\top\bm{J}_t)^{-1}\bm{J}_t^\top$ when $\bm{J}_t$ has full column rank.

The Schur complement of $\bm{H}_{tt}$ in $\bm{H}$ yields {the rotational components:}
\begin{align}
\bm{S}_R &= \bm{H}_{RR} - \bm{H}_{Rt}\bm{H}_{tt}^{+}\bm{H}_{tR} \\
&= \bm{J}_R^\top\bm{J}_R - \bm{J}_R^\top\bm{J}_t(\bm{J}_t^\top\bm{J}_t)^{+}\bm{J}_t^\top\bm{J}_R \\
&= \bm{J}_R^\top\left(\bm{I}_m - \bm{J}_t(\bm{J}_t^\top\bm{J}_t)^{+}\bm{J}_t^\top\right)\bm{J}_R \\
&= \bm{J}_R^\top(\bm{I}_m - \bm{P}_t)\bm{J}_R.
\end{align}

Since $\bm{P}_t$ is an orthogonal projector (satisfying $\bm{P}_t^2 = \bm{P}_t$ and $\bm{P}_t^\top = \bm{P}_t$), the operator $(\bm{I}_m - \bm{P}_t)$ is the orthogonal projector onto $\mathrm{range}(\bm{J}_t)^\perp = \mathrm{null}(\bm{J}_t^\top)$. This projection removes components of $\mathrm{range}(\bm{J}_R)$ that can be explained by $\bm{J}_t$, retaining only the rotational information that cannot be compensated by translation.

\paragraph{Weighted Case.}
Let $\bm{W} \in \mathbb{R}^{m \times m}$ be a positive definite information matrix ($\bm{W} \succ 0$). Define the whitened Jacobians as:
\begin{align}
\tilde{\bm{J}} &\triangleq \bm{W}^{1/2}\bm{J} \in \mathbb{R}^{m \times 6}, \\
\tilde{\bm{J}}_R &\triangleq \bm{W}^{1/2}\bm{J}_R \in \mathbb{R}^{m \times 3}, \\
\tilde{\bm{J}}_t &\triangleq \bm{W}^{1/2}\bm{J}_t \in \mathbb{R}^{m \times 3}.
\end{align}

The weighted Hessian matrix becomes:
\begin{equation}
\bm{H} = \bm{J}^\top\bm{W}\bm{J} = \tilde{\bm{J}}^\top\tilde{\bm{J}}.
\end{equation}

Following the same decomposition, the weighted Schur complement is:
\begin{align}
\bm{S}_R &= \bm{J}_R^\top\bm{W}\bm{J}_R - \bm{J}_R^\top\bm{W}\bm{J}_t\left(\bm{J}_t^\top\bm{W}\bm{J}_t\right)^{+}\bm{J}_t^\top\bm{W}\bm{J}_R.
\end{align}

Let $\tilde{\bm{P}}_t = \tilde{\bm{J}}_t\tilde{\bm{J}}_t^{+}$ be the orthogonal projector onto $\mathrm{range}(\tilde{\bm{J}}_t)$ in the standard inner product. Using the identity:
\begin{equation}
\bm{W}^{1/2}\tilde{\bm{P}}_t\bm{W}^{1/2} = \bm{W}\bm{J}_t\left(\bm{J}_t^\top\bm{W}\bm{J}_t\right)^{+}\bm{J}_t^\top\bm{W},
\end{equation}
we obtain two equivalent expressions for the weighted Schur complement:
\begin{align}
\bm{S}_R &= \bm{J}_R^\top\bm{W}^{1/2}\left(\bm{I}_m - \tilde{\bm{P}}_t\right)\bm{W}^{1/2}\bm{J}_R \\
&= \bm{J}_R^\top\left[\bm{W} - \bm{W}\bm{J}_t\left(\bm{J}_t^\top\bm{W}\bm{J}_t\right)^{+}\bm{J}_t^\top\bm{W}\right]\bm{J}_R.
\end{align}

Define the $\bm{W}$-orthogonal projector onto $\mathrm{range}(\bm{J}_t)$ as:
\begin{equation}
\bm{P}_t^{(\bm{W})} \triangleq \bm{J}_t\left(\bm{J}_t^\top\bm{W}\bm{J}_t\right)^{+}\bm{J}_t^\top\bm{W}.
\end{equation}

This operator satisfies the properties of a $\bm{W}$-orthogonal projector:
\begin{align}
\left(\bm{P}_t^{(\bm{W})}\right)^2 &= \bm{P}_t^{(\bm{W})}, \\
\left(\bm{P}_t^{(\bm{W})}\right)^\top\bm{W} &= \bm{W}\bm{P}_t^{(\bm{W})}.
\end{align}

Therefore,  {the weighted rotational Schur complement} can be expressed compactly as:
\begin{equation}
\bm{S}_R = \bm{J}_R^\top\bm{W}\left(\bm{I}_m - \bm{P}_t^{(\bm{W})}\right)\bm{J}_R,
\end{equation}
where $(\bm{I}_m - \bm{P}_t^{(\bm{W})})$ is the $\bm{W}$-orthogonal projector onto $\mathrm{range}(\bm{J}_t)^\perp$ with respect to the $\bm{W}$-inner product $\langle \bm{u}, \bm{v} \rangle_{\bm{W}} = \bm{u}^\top \bm{W} \bm{v}$.

In both cases, the eigenvalues of $\bm{S}_R$ in the projected subspace accurately reflect the rotational information that cannot be compensated by translational degrees of freedom, providing a geometrically meaningful decomposition of  {the system's conditioning properties.}
This completes the proof. \qed

\subsection{Proof of Proposition~\ref{prop:scale_invariance}}
\label{appendix:scale_invariance}

% \textbf{Part (i):} For any $\bm{u} \in \mathbb{R}^3$, the quadratic form:
% \begin{align}
% \bm{u}^\top\bm{S}_R\bm{u} &= \min_{\bm{v} \in \mathbb{R}^3} \left\{ 
% \begin{bmatrix}
% \bm{u} \\ \bm{v}
% \end{bmatrix}^\top
% \bm{H}
% \begin{bmatrix}
% \bm{u} \\ \bm{v}
% \end{bmatrix}
% \right\} \notag \\
% &\leq 
% \begin{bmatrix}
% \bm{u} \\ \bm{0}
% \end{bmatrix}^\top
% \bm{H}
% \begin{bmatrix}
% \bm{u} \\ \bm{0}
% \end{bmatrix}
% = \bm{u}^\top\bm{H}_{RR}\bm{u}
% \end{align}

% Therefore $\bm{S}_R \preceq \bm{H}_{RR}$ in the Loewner order.

% \textbf{Part (ii):} 
Under scaling $\delta\bm{t}' = s\delta\bm{t}$, the Hessian blocks transform as:
\begin{equation}
\bm{H}_{tt}' = s^2\bm{H}_{tt}, \quad \bm{H}_{Rt}' = s\bm{H}_{Rt}, \quad \bm{H}_{tR}' = s\bm{H}_{tR}
\end{equation}

The \textit{Schur complement} are:
\begin{align}
\bm{S}_R' &= \bm{H}_{RR} - \bm{H}_{Rt}'(\bm{H}_{tt}')^{-1}\bm{H}_{tR}' \notag \\
&= \bm{H}_{RR} - (s\bm{H}_{Rt})(s^2\bm{H}_{tt})^{-1}(s\bm{H}_{tR}) \notag \\
&= \bm{H}_{RR} - \bm{H}_{Rt}\bm{H}_{tt}^{-1}\bm{H}_{tR} = \bm{S}_R
\end{align}

Hence $\kappa(\bm{S}_R)$ is invariant to translation scaling. \qed

\subsection{Proof of Theorem~\ref{thm:condition_bounds}}\label{appendix:condition_bounds}

% \section{Proof of Theorem~\ref{thm:condition_bounds}}
% \label{appendix:condition_bounds}

% \subsection{Spectral Decomposition}
Consider the \textit{Schur complement} $\bm{S}_R$ in \cref{eq:schur_R}. Since $\bm{H}_{tt} \succ 0$, we have $\bm{M}_R \succeq 0$.

% \subsection{Eigenvalue Inequalities}
\textbf{Part (i):} For any $\bm{u} \in \mathbb{R}^3$, the quadratic form:
\begin{align}
\bm{u}^\top\bm{S}_R\bm{u} &= \argmin_{\bm{v} \in \mathbb{R}^3} \left\{ 
\begin{bmatrix}
\bm{u} \\ \bm{v}
\end{bmatrix}^\top
\bm{H}
\begin{bmatrix}
\bm{u} \\ \bm{v}
\end{bmatrix}
\right\} \notag \\
&\leq 
\begin{bmatrix}
\bm{u} \\ \bm{0}
\end{bmatrix}^\top
\bm{H}
\begin{bmatrix}
\bm{u} \\ \bm{0}
\end{bmatrix}
= \bm{u}^\top\bm{H}_{RR}\bm{u}
\end{align}

Therefore $\bm{S}_R \preceq \bm{H}_{RR}$ in the Loewner order.

\textbf{Part (ii):} According to Rayleigh-Ritz Principle~\cite{horn2013matrix}, 
% \begin{lemma}[Rayleigh-Ritz Principle~\cite{horn2013matrix}]
for symmetric matrices $\bm{A}, \bm{B} \in \mathbb{R}^{n \times n}$ with $\bm{B} \succeq 0$:
\begin{equation}
\lambda_i(\bm{A} - \bm{B}) \leq \lambda_i(\bm{A}), \quad i = 1,\ldots,n.
\end{equation}
% \end{lemma}

Applying this to $\bm{S}_R = \bm{H}_{RR} - \bm{M}_R$ yields:
\begin{equation}
\lambda_i(\bm{S}_R) \leq \lambda_i(\bm{H}_{RR}), \quad i = 1,2,3.
\end{equation}

\textbf{Part (iii):} According to Weyl's Perturbation Theorem~\cite{Weyl1912}
% \begin{lemma}[Weyl's Perturbation Theorem~\cite{Weyl1912}]
For Hermitian matrices $\bm{A}, \bm{B} \in \mathbb{C}^{n \times n}$:
\begin{equation}
\lambda_i(\bm{A} + \bm{B}) \geq \lambda_i(\bm{A}) + \lambda_{\min}(\bm{B}), \quad i = 1,\ldots,n.
\end{equation}
% \end{lemma}

Let $\bm{A} = \bm{H}_{RR}$ and $\bm{B} = -\bm{M}_R$:
\begin{equation}
\lambda_i(\bm{S}_R) \geq \lambda_i(\bm{H}_{RR}) - \lambda_{\max}(\bm{M}_R).
\end{equation}

% \subsection{Condition Number Bound}
From the spectral radius definition and the established bounds:
\begin{align}
\kappa(\bm{S}_R) &= \frac{\lambda_{\max}(\bm{S}_R)}{\lambda_{\min}(\bm{S}_R)} \\
&\leq \frac{\lambda_{\max}(\bm{H}_{RR})}{\lambda_{\min}(\bm{H}_{RR}) - \lambda_{\max}(\bm{M}_R)},
\end{align}
where the inequality holds when $\lambda_{\min}(\bm{H}_{RR}) > \lambda_{\max}(\bm{M}_R)$.
The bound for $\bm{S}_t$ follows by symmetry. \qed

\section{Proofs for Targeted Stabilization}
\label{appendix:tg_stabilization}

\subsection{Proof of Theorem~\ref{thm:solution_equivalence}}
\label{appendix:solution_equivalence}

% \begin{proof}
\textbf{Part (i):} From Proposition~\ref{prop:projection}, we established that $\bm{S}_R = \bm{J}_R^\top(\bm{I} - \bm{P}_t)\bm{J}_R$. Following the derivation in \cref{eq:schur_R}, the reduced gradient after eliminating translation becomes:
\begin{equation}
\tilde{\bm{g}}_R = \bm{g}_R - \bm{H}_{Rt}\bm{H}_{tt}^{-1}\bm{g}_t = \bm{J}_R^\top(\bm{I} - \bm{P}_t)\bm{r}_0
\end{equation}

Since $(\bm{I} - \bm{P}_t)$ projects onto the orthogonal complement of $\mathrm{range}(\bm{J}_t)$, we have $\tilde{\bm{g}}_R \in \mathrm{range}(\bm{S}_R)$, ensuring the existence of a minimizer for $Q(\bm{\phi})$.

\textbf{Part (ii):} With $\tilde{\bm{g}}_R \in \mathrm{range}(\bm{S}_R)$ established, the optimality condition $\bm{S}_R\bm{\phi} = \tilde{\bm{g}}_R$ in the eigenbasis yields:
\begin{equation}
[\bm{V}_R^\top\bm{\phi}]_i = 
\begin{cases}
[\bm{V}_R^\top\tilde{\bm{g}}_R]_i/\lambda_{R,i} & \text{if } \lambda_{R,i} > 0\\
0 & \text{if } \lambda_{R,i} = 0
% 0 & \text{if } \lambda_{R,i} = 0 \quad  {\text{(minimum-norm selection)}}
\end{cases}
\end{equation}

This is precisely the Moore-Penrose pseudoinverse: $\bm{\phi}^* = \bm{S}_R^{+}\tilde{\bm{g}}_R$.

\textbf{Part (iii):} For the regularized system $(\bm{S}_R + \epsilon\bm{I})\bm{\phi}_{\epsilon} = \tilde{\bm{g}}_R$, following the eigenvalue analysis from \cref{eq:update_eigen}:
\begin{equation}
[\bm{V}_R^\top\bm{\phi}_{\epsilon}]_i = \frac{[\bm{V}_R^\top\tilde{\bm{g}}_R]_i}{\lambda_{R,i} + \epsilon} \xrightarrow{\epsilon \to 0^+} [\bm{V}_R^\top\bm{\phi}^*]_i \quad \text{for } \lambda_{R,i} > 0
\end{equation}
establishing invariance on observable components. \qed
% \end{proof}

\subsection{Proof of Theorem~\ref{thm:map_interpretation}}
\label{appendix:map_interpretation}

% \begin{proof}
\textbf{Part (i):} The MAP objective with Gaussian prior $\bm{\phi} \sim \mathcal{N}(\bm{0}, \bm{\Gamma}_R^{-1})$ yields:
\begin{equation}
\bm{\phi}_{\text{MAP}} = \argmin_{\bm{\phi}} \left[\frac{1}{2}\bm{\phi}^\top\bm{S}_R\bm{\phi} - \tilde{\bm{g}}_R^\top\bm{\phi} + \frac{1}{2}\bm{\phi}^\top\bm{\Gamma}_R\bm{\phi}\right]
\end{equation}
which is equivalent to the stated regularized problem with posterior Hessian $\bm{S}_R + \bm{\Gamma}_R$.

\textbf{Part (ii):} Using the eigenvalue clamping from \cref{eq:eigenvalue_clamping}, define:
\begin{equation}
\bm{\Gamma}_R = \bm{V}_R\text{diag}(\tilde{\lambda}_{R,i} - \lambda_{R,i})\bm{V}_R^\top
\end{equation}
Since both $\bm{S}_R$ and $\bm{\Gamma}_R$ share the same eigenvector basis $\bm{V}_R$:
\begin{equation}
\bm{S}_R + \bm{\Gamma}_R = \bm{V}_R\text{diag}(\lambda_{R,i} + \tilde{\lambda}_{R,i} - \lambda_{R,i})\bm{V}_R^\top = \bm{V}_R\tilde{\bm{\Lambda}}_R\bm{V}_R^\top
\end{equation}

\textbf{Part (iii):} From the clamping definition $\tilde{\lambda}_{R,i} = \max(\lambda_{R,i}, \lambda_{R,3}/\kappa_{\text{tg}})$:
\begin{itemize}
\item $\tilde{\lambda}_{R,3} = \lambda_{R,3}$ (maximum eigenvalue unchanged)
\item $\tilde{\lambda}_{R,1} \geq \lambda_{R,3}/\kappa_{\text{tg}}$ (clamping lower bound)
\end{itemize}
Therefore: $\kappa(\bm{S}_R + \bm{\Gamma}_R) = \tilde{\lambda}_{R,3}/\tilde{\lambda}_{R,1} \leq \kappa_{\text{tg}}$. \qed
% \end{proof}

\end{document}